\documentclass[sigconf]{acmart}
\usepackage{graphicx}
\usepackage{subcaption}
\usepackage[linesnumbered,ruled,vlined]{algorithm2e}
\usepackage{multirow}
\usepackage{enumerate}
\usepackage{adjustbox}
\graphicspath{ {images/} {images-suppl/} }

\AtBeginDocument{%
  \providecommand\BibTeX{{%
    \normalfont B\kern-0.5em{\scshape i\kern-0.25em b}\kern-0.8em\TeX}}}

\copyrightyear{2020}
\acmYear{2020}
\setcopyright{acmcopyright}
\acmConference[MM '20]{Proceedings of the 28th ACM International Conference on Multimedia}{October 12--16, 2020}{Seattle, WA, USA}
\acmBooktitle{Proceedings of the 28th ACM International Conference on Multimedia (MM '20), October 12--16, 2020, Seattle, WA, USA}
\acmPrice{15.00}
\acmDOI{10.1145/3394171.3413814}
\acmISBN{978-1-4503-7988-5/20/10}

\acmSubmissionID{mmfp0668}

\begin{document}
\fancyhead{}
\title{Performance Optimization for Federated Person Re-identification via Benchmark Analysis}










\author{Weiming Zhuang$^{1, 2}$ \quad Yonggang Wen$^{1}$ \quad Xuesen Zhang$^{2}$ \quad Xin Gan$^{2}$ \quad Daiying Yin$^{2}$}
\author{Dongzhan Zhou$^{2}$ \quad Shuai Zhang$^{2}$ \quad Shuai Yi$^{2}$}
\affiliation{%
 $^{1}$Nanyang Technological University, Singapore \\ 
 $^{2}$SenseTime Research, China
}
\email{weiming001@e.ntu.edu.sg, ygwen@ntu.edu.sg}
\email{{zhangxuesen, ganxin, yindaiying, zhoudongzhan, zhangshuai, yishuai}@sensetime.com}

\renewcommand{\authors}{Weiming Zhuang, Yonggang Wen, Xuesen Zhang, Xin Gan, Daiying Yin, Dongzhan Zhou, Shuai Zhang, and Shuai Yi}

\renewcommand{\shortauthors}{W. Zhuang et al.}

\begin{abstract}

Federated learning is a privacy-preserving machine learning technique that learns a shared model across decentralized clients. It can alleviate privacy concerns of personal re-identification, an important computer vision task. In this work, we implement federated learning to person re-identification (\textit{FedReID}) and optimize its performance affected by statistical heterogeneity in the real-world scenario. We first construct a new benchmark to investigate the performance of FedReID. This benchmark consists of (1) nine datasets with different volumes sourced from different domains to simulate the heterogeneous situation in reality, (2) two federated scenarios, and (3) an enhanced federated algorithm for FedReID. The benchmark analysis shows that the client-edge-cloud architecture, represented by the federated-by-dataset scenario, has better performance than client-server architecture in FedReID. It also reveals the bottlenecks of FedReID under the real-world scenario, including poor performance of large datasets caused by unbalanced weights in model aggregation and challenges in convergence. Then we propose two optimization methods: (1) To address the unbalanced weight problem, we propose a new method to dynamically change the weights according to the scale of model changes in clients in each training round; (2) To facilitate convergence, we adopt knowledge distillation to refine the server model with knowledge generated from client models on a public dataset. Experiment results demonstrate that our strategies can achieve much better convergence with superior performance on all datasets. We believe that our work will inspire the community to further explore the implementation of federated learning on more computer vision tasks in real-world scenarios.

\end{abstract}



\maketitle

\begin{table*}[]
\caption{The characteristics of 9 datasets of FedReID benchmark.}
\begin{tabular}{lccccccc}
\hline
\multicolumn{1}{l}{\multirow{3}{*}{Datasets}} &
  \multicolumn{1}{c}{\multirow{3}{*}{\# Cameras}} &
  \multicolumn{2}{c}{Train} &
  \multicolumn{1}{c}{} &
  \multicolumn{3}{c}{Test} \\ \cline{3-4}  \cline{6-8}  
\multicolumn{1}{c}{} &
  \multicolumn{1}{c}{} &
  \multirow{2}{*}{\# IDs} &
  \multirow{2}{*}{\# Images} &
  \multirow{2}{*}{} &
  \multicolumn{2}{c}{Query} &
  Gallery \\ \cline{6-7} 
\multicolumn{1}{c}{} &
  \multicolumn{1}{c}{} & & & & \# IDs & \# Images & \# Images \\ \hline \hline
MSMT17 \cite{Wei2017Msmt} & 15 & 1,041 & 32,621 &  & 3,060 & 11,659 & 82,161 \\ 
DukeMTMC-reID \cite{zheng2017dukemtmc-reid} & 8  & 702   & 16,522 &  & 702   & 2,228  & 17,611 \\
Market-1501 \cite{Zheng2015Market1501} & 6  & 751   & 12,936 & & 750   & 3,368  & 19,732 \\ 
CUHK03-NP \cite{Li2014CUHK03} & 2  & 767   & 7,365 &  & 700   & 1,400  & 5,332  \\ 
PRID2011 \cite{prid2011}  & 2  & 285   & 3,744 &   & 100   & 100    & 649    \\ 
CUHK01 \cite{li2012cuhk01} & 2  & 485   & 1,940  &  & 486   & 972    & 972    \\ 
VIPeR \cite{Gray2008ViewpointIP}  & 2  & 316   & 632 & & 316   & 316    & 316    \\ 
3DPeS \cite{3dpes} & 2  & 93    & 450   &  & 86    & 246    & 316    \\ 
iLIDS-VID \cite{iLIDS-VID}  & 2  & 59    & 248  &   & 60    & 98     & 130    \\ 
\hline
\end{tabular}
\label{tab:dataset}
\end{table*}

\section{Introduction}

\begin{figure}[h]
    \centering
    \includegraphics[width=\linewidth]{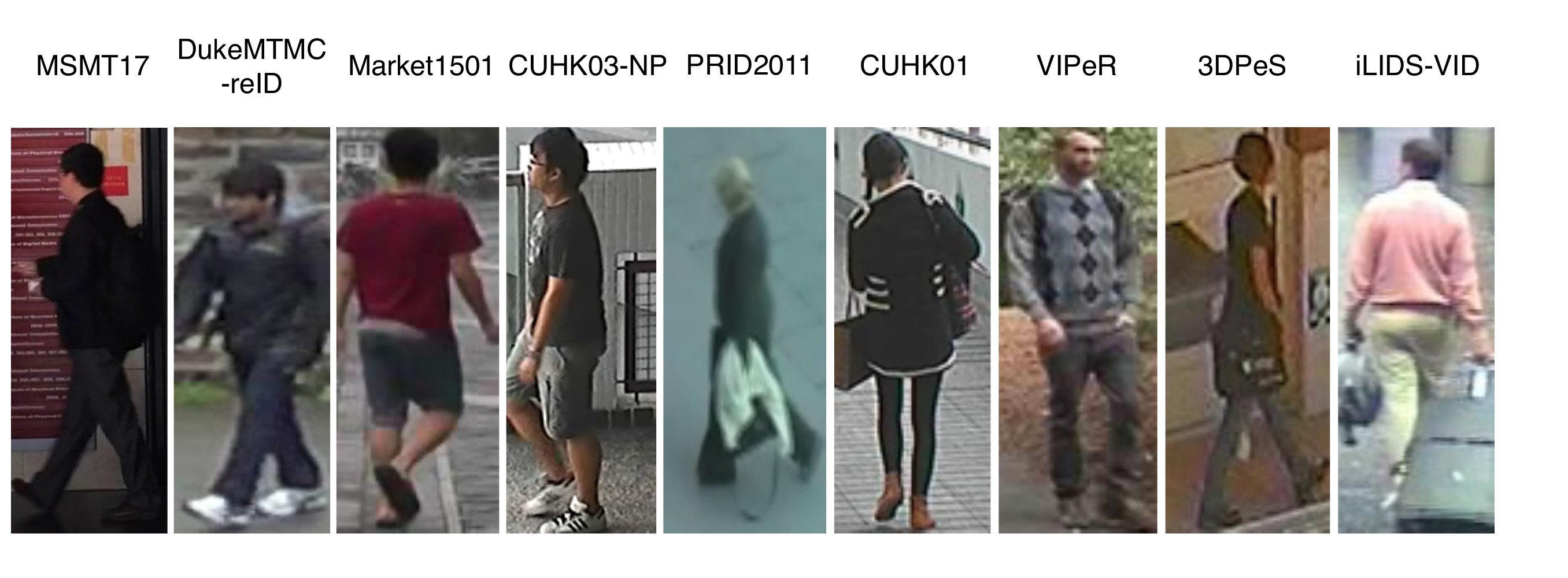}
    \caption{Sample images of the 9 selected datasets.}
    \label{fig:dataset}
\end{figure}

The increasing awareness of personal data protection \cite{gdpr} has limited the development of person re-identification (ReID). Person ReID is an important computer vision task that matches the same individual in a gallery of images \cite{zheng2016person-reid-survey}. The training of person ReID relies on centralizing a huge amount of personal image data, imposing potential privacy risks on personal information, and even causing the suspension of person ReID research projects in some countries. Therefore, it is necessary to navigate its development under the premise of privacy preservation.

Federated learning is a privacy-preserving machine learning framework that can train a person ReID model using decentralized data from the cameras. Since edges share model updates instead of training data with the server \cite{McMahanMRHA17}, federated learning can effectively mitigate potential privacy leakage risks. Multimedia researchers and practitioners can also leverage this benefit to multimedia content analysis tasks \cite{deepQoE, Chen-content-analysis}. In addition to privacy protection, the implementation of federated learning to person ReID (FedReID) also possesses other advantages: reducing communication overhead by avoiding massive data uploads \cite{McMahanMRHA17}; enabling a holistic model that is applicable to different scenarios; obtaining local models at edges that can adapt local scenes. Video surveillance for communities is a good use case for FedReID \cite{webankWhitepaper}. Different communities collaborate to train a centralized model without video data leaving communities. 
 
Despite the advantages of federated learning, little work studies its implementation to person ReID. Hao et al. \cite{Hao2018EdgeAIBench} only mentioned the possibility of this implementation. Statistical heterogeneity---data with non-identical and independent distribution (non-IID) and unbalanced volumes---is one of the key challenges for FedReID in the real-world scenario \cite{Li2020FedChallenges}. Zhao et al. \cite{zhao2018non-iid} showed non-IID data harms the performance of federated learning significantly and Li et al. \cite{fedprox} stated that it causes the challenge of convergence, but little work studies statistical heterogeneity in FedReID. 

This work aims to optimize the performance of FedReID by performing benchmark analysis. Comprehensive experimental results and analysis of the newly constructed benchmark and the proposed optimization methods demonstrate their usefulness and effectiveness. To the best of our knowledge, this is the first implementation of federated learning to person ReID. We summarize the contributions of this paper as follows:

\begin{enumerate}[(I)]
    \item We construct a new benchmark for FedReID and conduct benchmark analysis to investigate its bottlenecks and insights. Our benchmark, \textit{FedReIDBench}, has following features: (1) using 9 representative ReID datasets---samples shown in Figure \ref{fig:dataset}---with large variances to simulate the real-world situation of non-IID and unbalanced data, (2) defining representative federated scenarios for person ReID, (3) proposing a suitable algorithm for FedReID, (4) standardizing model structure and performance evaluation metrics, and (5) creating reference implementation to define the training procedures. The benchmark analysis results set a good baseline for future research on this topic. 
    \item We propose two methods to optimize performance: knowledge distillation and dynamic weight adjustment. Knowledge distillation \cite{kd} addresses the convergence problem caused by non-IID data. Dynamic weight adjustment in model aggregation solves the performance decay issue caused by imbalance datasets. 
\end{enumerate}

The rest of the paper is organized as follows. In Section \ref{sec:related-work}, we review related work about person ReID and federated learning. Section \ref{sec:benchmark} presents the benchmark for FedReID. We analyze the benchmark results and provide insights in Section \ref{sec:benchmark-analysis}. In Section \ref{sec:performance-optimization}, we propose optimization methods to improve the performance of FedReID. Section \ref{sec:conclusion} summarizes this paper and provides future directions.

\section{Related Work}
\label{sec:related-work}

\subsection{Person Re-identification}

Given a query image, the person ReID system aims to retrieve images with the same identity from a large gallery, based on their similarities. It has wide applications such as video surveillance and content-based video retrieval \cite{zheng2016person-reid-survey}. Compared with traditional hand-crafted feature operators, deep neural networks enable better extracting representative features and hence greatly improve the performance of ReID \cite{2014deepReID, Liu-reid-multi-scale, local-cnn-reid-acmmm, multi-gra-acmmm}. The person ReID datasets contain images from different camera views. Training person ReID models requires centralizing a large amount of these data, which raises potential privacy risks because these images contain personal information and identification. Thus, federated learning is beneficial for person ReID to preserve privacy.

\subsection{Federated Learning}

\textbf{Federated learning benchmarks} Caldas et al. in \cite{Leaf} presented LEAF, a benchmark framework focusing on image classification and some natural language process tasks. Luo et. al in \cite{luo2019realworld} proposed real-world image datasets for object detection. Both works adopt Federated Averaging (FedAvg) algorithm proposed by McMahan et al.\cite{McMahanMRHA17} as the baseline implementation. In this work, we introduce a new benchmark in the combination of federated learning and person ReID, where we report a comprehensive analysis to reveal the problems and provide insights for the simulated real-world scenario.

\noindent\textbf{Non-IID data in federated learning} Federated learning faces the challenge of non-IID data \cite{zhao2018non-iid}, which is different from distributed deep learning that trains a large scale deep network with parallel computation using IID data in clusters \cite{large-scale-dnn-NIPS2012, sunpeng}. Zhao et al.~\cite{zhao2018non-iid} proposed sharing data that represents global distribution to clients to improve non-IID performance. Yao et al. \cite{xin2019meta} proposed FedMeta, a method to fine-tune the server model after aggregation using metadata acquired from voluntary clients. Li et al. in \cite{fedprox} offer FedProx, an algorithm to improve the convergence of FedAvg by adding a proximal term to restrict the local update to be closer to the global model. We also provide two solutions for problems caused by non-IID data in the ReID task. Inspired by data sharing strategy \cite{zhao2018non-iid} and FedMeta \cite{xin2019meta}, one of the solution adopts knowledge distillation with an additional unlabelled dataset to facilitate convergence. 

\section{Federated Person ReID Benchmark}
\label{sec:benchmark}

In this section, we introduce FedReIDBench, a new benchmark for implementing federated learning to person ReID. It includes 9 datasets (Section \ref{sec:datasets}), choices of federated scenarios (Section \ref{sec:fed-scenarios}), the model structure (Section \ref{sec:network-model}), the federated training algorithms (Section \ref{sec:fed-algo}), the performance metrics (Section \ref{sec:performance-metrics}), and reference implementation (Section \ref{sec:referenced-implementation}).

\subsection{Datasets}
\label{sec:datasets}

To simulate the real-world scenario of FedReID, we select 9 different datasets whose properties are shown in Table~\ref{tab:dataset}. These datasets have significant variances in image amounts, identity numbers, scenes (indoor or outdoor), and the number of camera views, which lead to huge domain gaps among each other \cite{Liu_2019_CVPR}. These variances simulate the statistical heterogeneity in reality. The disparity in image amounts simulates the imbalance of data points across edges and the domain gaps results in the non-IID problem. The simulated statistical heterogeneity makes the FedReID scenario more challenging and closer to the real-world situation. 

\subsection{Federated Scenarios}
\label{sec:fed-scenarios}

\begin{figure}[h!]
  \centering
  \begin{subfigure}[b]{0.46\linewidth}
    \includegraphics[width=\linewidth]{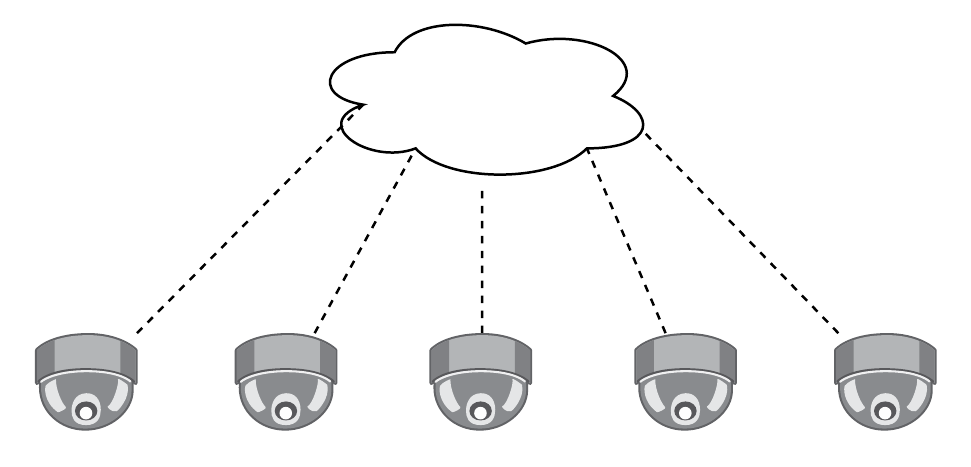}
    \caption{}
  \end{subfigure}
  \hspace{1em}%
  \begin{subfigure}[b]{0.46\linewidth}
    \includegraphics[width=\linewidth]{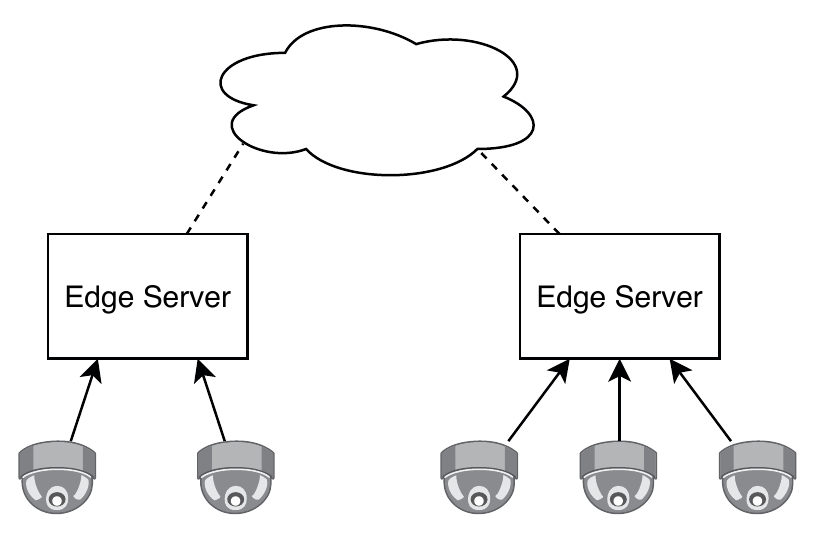}
    \caption{}
  \end{subfigure}
  \caption{Federated-by-camera scenario vs. Federated-by-dataset scenario. (a) represents federated-by-camera-scenario: cameras collaboratively perform federated learning with the server. (b) represents federated-by-dataset scenario: edge servers collect data from multiple cameras before performing federated learning.}
  \label{fig:federated-scenarios}
\end{figure}

We design two different approaches that representing two real-world scenarios for applying federated learning to person ReID (Figure \ref{fig:federated-scenarios}).

\textbf{Federated-by-camera scenario} represents a standard client-server architecture. Each camera is defined as an individual client to directly communicate with the server to conduct the federated learning process. Under this scenario, keeping images in clients significantly reduces the risk of privacy leakage. However, this scenario exerts high requirements on the computation ability of cameras to train deep models, which makes practical deployment harder. A good illustration in the real world would be a community that deploys multiple cameras to train one person ReID model.

\textbf{Federated-by-dataset scenario} represents a client-edge-cloud architecture, where clients are defined as the edge servers. The edge servers construct dataset from multiple cameras and then collaboratively conduct federated learning with the central server. A real-world scenario could be several communities collaboratively train ReID models with an edge server connecting to multiple cameras in each community.

\subsection{Model Structure}
\label{sec:network-model}

A common baseline for deep person ReID is the ID-discriminative embedding (IDE) model \cite{zheng2016person-reid-survey}. We use the IDE model with backbone ResNet-50 \cite{resnet} as our model structure to perform federated learning. However, the model structure is not identical in all clients---their identity classifiers may be different. Clients have a different number of identities in both federated scenarios introduced in Section~\ref{sec:fed-scenarios} and the dimension of the identity classifier in the model depends on the number of identities, so they may have different model structures. This difference affects the federated algorithm we discussed in the following section (Section \ref{sec:fed-algo}).

\subsection{Federated Learning Algorithms}
\label{sec:fed-algo}

In this section, we introduce FedAvg, the key algorithm of federated learning, and outline our proposed method Federated Partial Averaging (FedPav) for FedReID.

\textbf{Federated Averaging} (FedAvg) \cite{McMahanMRHA17} is a standard federated learning algorithm which includes operations on both the server and clients: clients train models with their local dataset and upload model updates to the server; the server is responsible for initializing the network model and aggregating model updates from clients by weighted average. FedAvg requires the models in the server and clients having the same network architecture, while as discussed in Section \ref{sec:network-model}, the identity classifiers of clients could be different. Hence, we introduce an enhanced federated learning algorithm for FedReID:  Federated Partial Averaging.

\begin{figure}[t]
    \centering
    \includegraphics[width=0.47\textwidth]{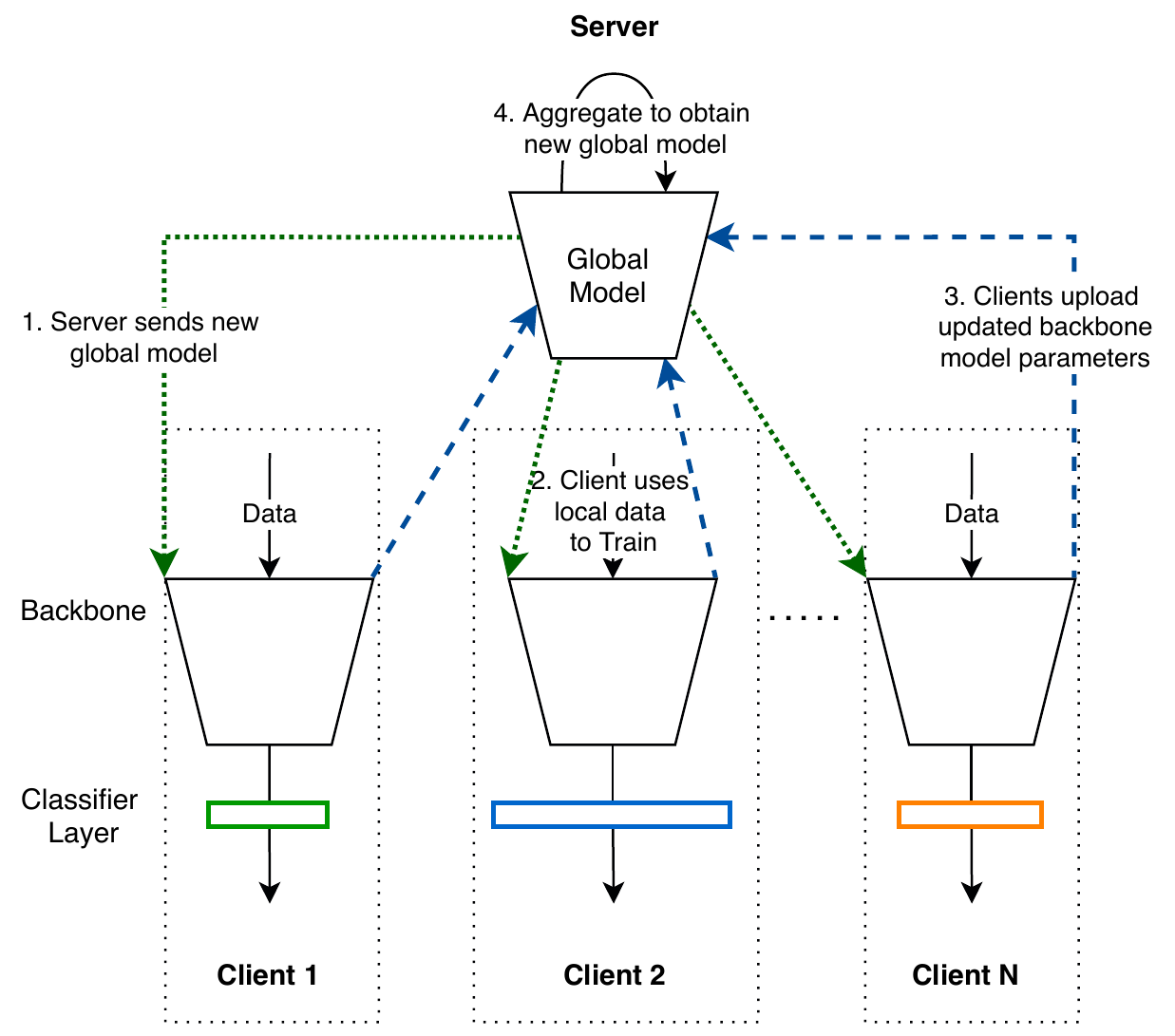}
    \caption{Illustration of Federated Partial Averaging (FedPav). The global model is the backbone. Each round of training includes the following steps: (1) A server sends the global model to clients. (2) Clients use local data to train the models with their classifiers. (3) Clients upload the backbone parameters to the server. (4) The server aggregates model updates from clients by weighted average to obtain a new global model.}
    \label{fig:fedpav}
\end{figure}

\textbf{Federated Partial Averaging} (FedPav) enables federated training with clients that have partially different models. It is similar to FedAvg in the whole training process except that each client sends only part of the updated model to the server. Figure \ref{fig:fedpav} depicts the implementation of FedPav to FedReID. Models in clients share an identical backbone, varying the identity classifiers, so the clients only send model parameters of the backbone to the server for aggregation. 

We describe the training process as follows: (1) At the beginning of a new training round, the server selects $K$ out of $N$ total clients to participate in the training and sends the global model to clients. (2) Each client concatenates the global model with the identity classifier from the previous training round to form a new model. It then trains the model on local data with stochastic gradient descent for $E$ number of local epochs with batch size $B$ and learning rate $\eta$. (3) Each client preserves the classifier layer and uploads the updated model parameters of the backbone. (4) The server aggregates these model updates, obtaining a new global model. We summarize FedPav in Algorithm \ref{algo:fedpav}. 

FedPav aims to obtain models outperforming \textit{local training}, which represents models trained on individual datasets. FedPav outputs a high-quality global model $w^T$ and a \textit{local model} $w_k^T$ for each client. These models are evaluated and compared with \textit{local training} in Section \ref{sec:federated-by-datasets}. Since ReID evaluation uses an image as a query to search for similar images in a gallery, we can omit the identity classifier in evaluation.  

\SetKwInput{KwInput}{Input}                
\SetKwInput{KwOutput}{Output}              
\SetKwFunction{FnClient}{}
\SetKwFunction{FnServer}{}

\begin{algorithm}[]
    \caption{Federated Partial Averaging (FedPav)}
    \label{algo:fedpav}
    \SetAlgoLined
    \KwInput{$E, B, K, \eta, T, N, n_k, n$}
    \KwOutput{$w^T, w^T_k$}
    
    \SetKwProg{Fn}{Server}{:}{}
    \Fn{\FnServer}{
        initialize $w^0$ \;
        \For{each round t = 0 to T-1}{
            $C_t \leftarrow$ (randomly select K out of N clients)\;
            \For{each client $k \in C_t$ concurrently}{
                $w^{t+1}_k \leftarrow $ \textbf{ClientExecution}($w^t$, $k$, $t$)\;
            }
            // $n$: total size of dataset; $n_k$: size of client $k$'s dataset\;
            $w^{t+1} \leftarrow \sum_{k \in C_t} \frac{n_k}{n} w^{t+1}_k$\;
        }
        \KwRet $w^{T}$\;
    }
    
    \SetKwProg{Fn}{ClientExecution}{:}{\KwRet}
    \Fn{\FnClient{w, k, t}}{
        $v \leftarrow$ (retrieve additional layers $v$ if $t > 0$ else initialize)\;
        $\mathcal{B} \leftarrow $ (divide local data into batches of size $B$)\;
        \For{each local epoch e = 0 to E-1}{
            \For{$b \in \mathcal{B}$}{
                // $(w, v)$ concatenation of two vectors\;
                $(w, v) \leftarrow (w, v) - \eta \triangledown \mathcal{L}((w, v); b)$\;
            }
        }
        store $v$\;
        \KwRet $w$\;
    }
\end{algorithm}

\subsection{Performance Metrics}
\label{sec:performance-metrics}

To evaluate the performance of FedReID, we need to measure not only the accuracy of algorithms but also the communication cost because the federated learning setting limits the communication bandwidth.

\noindent\textbf{ReID Evaluation Metrics} We use the standard person ReID evaluation metrics to evaluate the accuracy of our algorithms: Cummulative Matching Characteristics (CMC) curve and mean Average Precision (mAP) \cite{zheng2016person-reid-survey}. CMC ranks the similarity of a query identity to all the gallery images; Rank-k presents the probability that the top-k ranked images in the gallery contain the query identity. We measure CMC at rank-1, rank-5, and rank-10. mAP calculates the mean of average precision in all queries.

\noindent\textbf{Communication Cost} We measure communication cost by the number of communication rounds times twice the model size (uploading and downloading). Larger communication rounds lead to higher communication costs if the model size is constant. 

\subsection{Reference Implementation}
\label{sec:referenced-implementation}

To facilitate the reproducibility, FedReIDBench provides a set of referenced implementations, including FedPav and optimization methods. It also includes scripts to preprocess the ReID datasets.

\section{Benchmark Analysis}
\label{sec:benchmark-analysis}

Using the benchmark defined in Section \ref{sec:benchmark}, we conducted extensive experiments on different federated settings and gained meaningful insights by analyzing these results. For all the experiments, we initialize the ResNet-50 model using pre-trained ImageNet model \cite{zheng2016person-reid-survey}. We present rank-1 accuracy in most experiments and provide mAP results in the supplementary material.

\subsection{Federated-by-camera Scenario}

Since the existing ReID datasets contain images from multiple cameras, we consider implementing federated learning regarding each camera as a client. We assume that the cameras have sufficient computation power to train neural network models. Some cameras in the industry already have such capability.

We measure the performance in the federated-by-camera scenario by two datasets: Market-1501 \cite{Zheng2015Market1501} dataset that contains training data from 6 camera views and CUHK03-NP \cite{Li2014CUHK03} dataset that contains images from 2 camera views. We split the Market1501 dataset to 6 clients and CUHK03-NP dataset to 2 clients with each client containing data from one camera view. To compare with the performance of the federated-by-camera scenario, we define a federated-by-identity scenario by splitting a dataset into several clients, each of them has the same number of identities from different camera views. The number of clients in the federated-by-dataset scenario is equal to the number of camera views. For example, we split Market-1501 to 6 clients by identities, thus each client contains 125 non-overlap identities. We also add \textit{local training} to the comparison. We implement FedPav on both federated-by-camera and federated-by-identity scenarios under the same setting and summarize the results in Table \ref{tab:federated-by-cameras}. 

The federated-by-camera scenario has poor performance compared with local training and federated-by-identity scenario on both datasets. Learning from cross-camera knowledge is essential to train a ReID model. Since each client only learns from one-camera-view data in the federated-by-camera scenario, the model is incapable to generalize to the multi-camera evaluation. Furthermore, these results suggest that, instead of standard client-server architecture, the federated-by-dataset scenario that represents client-edge-cloud architecture could be more suitable. We conduct the following experiments on the federated-by-dataset scenario.

\begin{table}[t]
\caption{Performance comparison of federated-by-camera scenario, federated-by-identity scenario, and local training on Market-1501 dataset and CUHK03-NP dataset. The federated-by-camera scenario has the lowest accuracy.}
\begin{tabular}{llrlll}
\hline
\multicolumn{2}{c}{Setting}       & Rank-1            & Rank-5            & Rank-10  &mAP         \\ \hline
\multicolumn{2}{c}{\textit{Market-1501} \cite{Zheng2015Market1501}} & \multicolumn{1}{l}{} &  \\
\multicolumn{2}{l}{Local Training} & 88.93 & 95.34 & 96.88 & 72.62 \\
\multicolumn{2}{l}{Federated-by-identity}           & 85.69    & 93.44          & 95.81          & 66.36          \\
\multicolumn{2}{l}{Federated-by-camera}        & 61.13                               & 74.88          & 80.55          & 36.57          \\  \hline
\multicolumn{2}{c}{\textit{CUHK03-NP} \cite{Li2014CUHK03}}      & \multicolumn{1}{l}{} &   \\  
\multicolumn{2}{l}{Local Training} & 49.29 & 68.86 & 76.57 & 44.52 \\  
\multicolumn{2}{l}{Federated-by-identity}                &51.71                            & 69.50        & 76.79        & 47.39       \\
\multicolumn{2}{l}{Federated-by-camera}            & 11.21                              & 19.14          & 25.71          & 11.11          \\ \hline
\end{tabular}
\label{tab:federated-by-cameras}
\end{table}

\subsection{Federated-by-dataset Scenario}
\label{sec:federated-by-datasets}

In this section, we analyze the results of the federated-by-dataset scenario and investigate the impact of batch size $B$, the impact of local epochs $E$, performance comparison to \textit{local training}, and the convergence of FedPav. We conducted all the following experiments with 9 clients and each client trained on one of the 9 datasets. In each communication round, we selected all clients for aggregation.

\textbf{Impact of Batch Size} Batch size is an important hyperparameter in the FedPav, which affects the computation in clients. With the same number of local epochs and fixed size of a dataset, smaller batch size leads to higher computation in clients in each round of training. We compare the performance of different batch sizes with setting $E = 1$ and total 300 rounds of communication in Figure \ref{fig:batch-size}. The performance increases in most datasets as we add more computation by changing batch size from 128 to 32. Hence, we use $B = 32$ as the default batch size setting for other experiments.

\begin{figure}[t]
    \centering
    \includegraphics[width=0.47\textwidth]{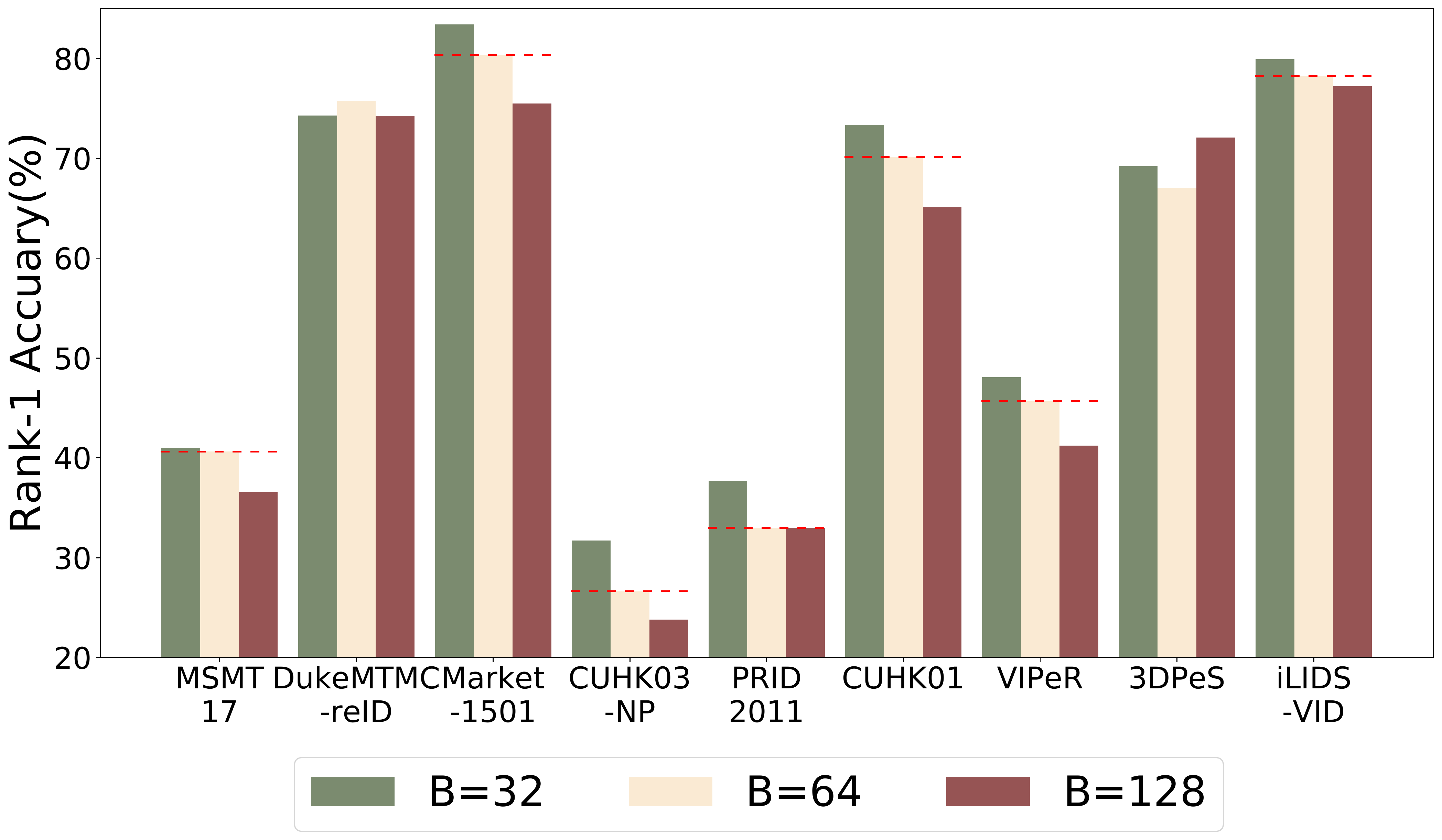}
    \caption{Performance (rank-1) comparison of different batch sizes, fixing local epochs $E = 1$. Batch size $B = 32$ has the best performance in most datasets.}
    \label{fig:batch-size}
\end{figure}

 \begin{figure}[t]
    \centering
    \includegraphics[width=0.47\textwidth]{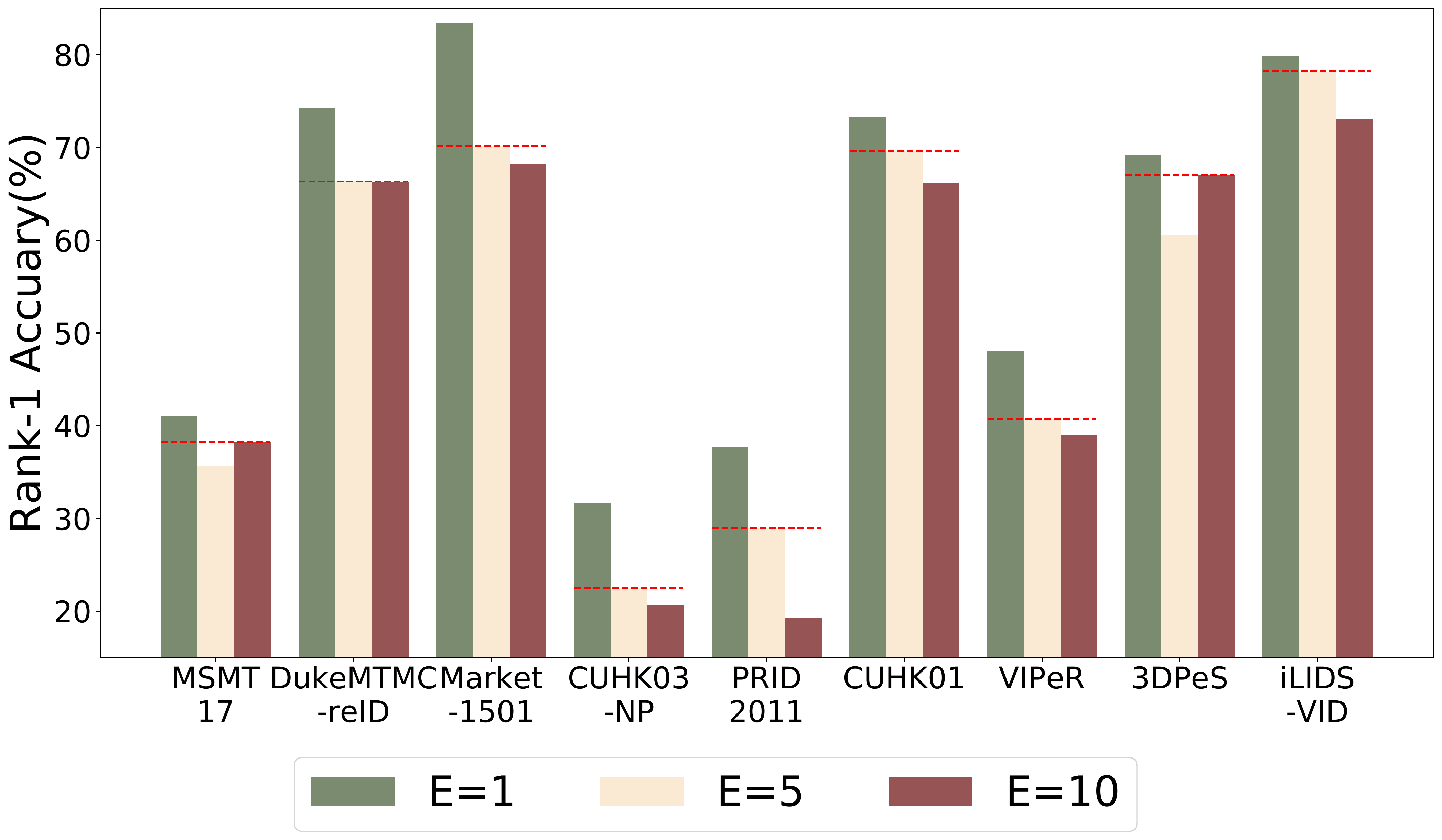}
    \caption{Performance comparison of different number of local epochs, fixing batch size $B = 32$ and total training rounds $ET = 300$. Local epoch $E = 1$ has the best performance in all datasets.}
    \label{fig:local-epoch}
\end{figure}

\textbf{Communication Cost} The number of local epochs in FedPav represents the trade-off between communication cost and performance. Figure \ref{fig:local-epoch} compares the rank-1 accuracy of number of local epochs $E = 1$, $E = 5$, and $E = 10$ with $B = 32$ and 300 total training rounds. Although $E = 10$ outperforms $E = 5$ in few datasets, decreasing $E$ generally improves performance and $E = 1$ greatly outperforms $E = 5$ and $E = 10$ in all datasets. It indicates the trade-off between performance and communication cost in FedReID. A smaller number of local epochs achieve better performance but result in higher communication cost.

\begin{figure}[t]
  \centering
  \begin{subfigure}[b]{0.9\linewidth}
    \includegraphics[width=\linewidth]{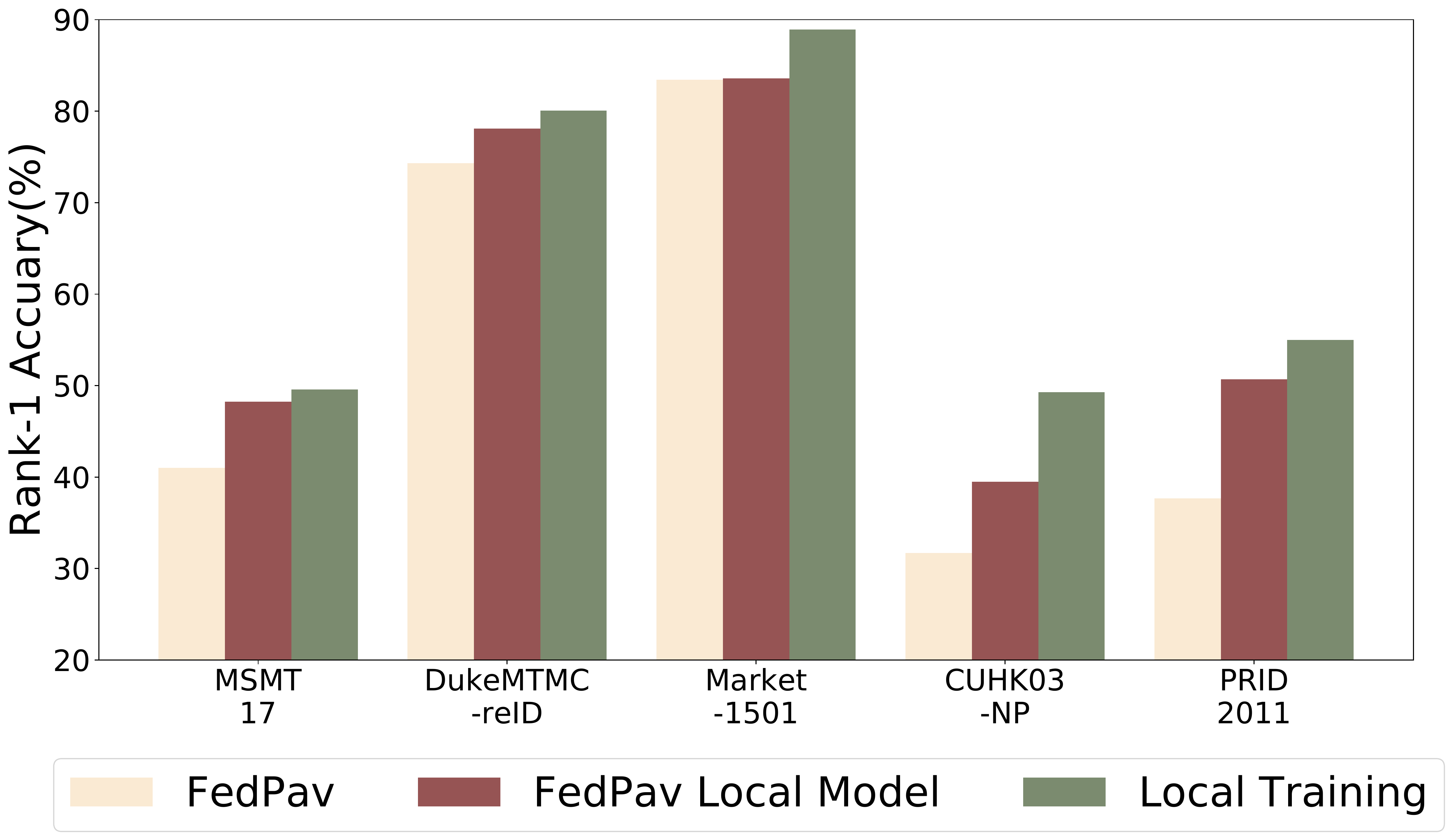}
    \caption{}
    \label{fig:fedpav-vs-local-a}
  \end{subfigure}
  \begin{subfigure}[b]{0.9\linewidth}
    \includegraphics[width=\linewidth]{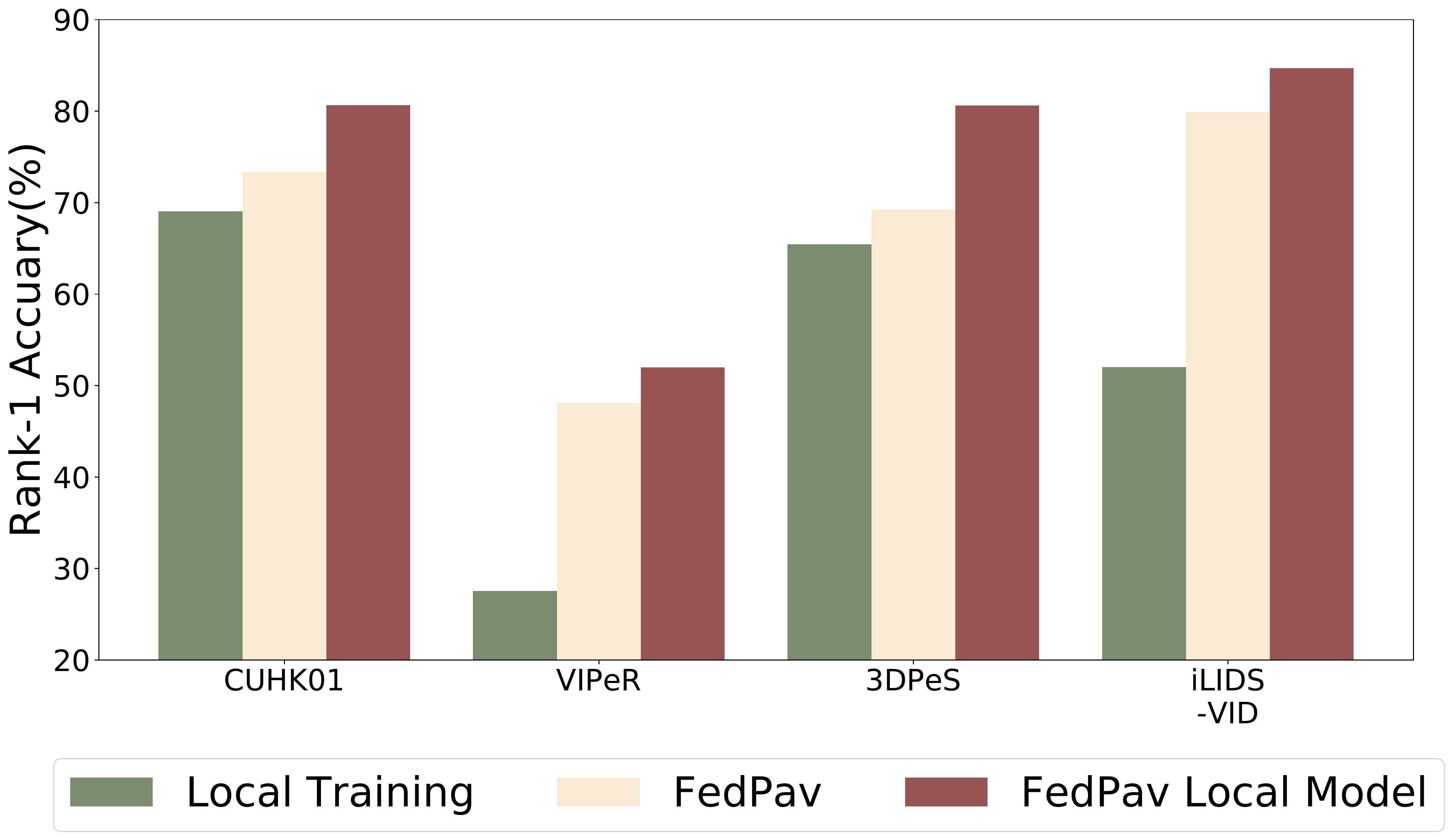}
    \caption{}
    \label{fig:fedpav-vs-local-b}
  \end{subfigure}
  \caption{Performance comparison of FedPav and local training (training on individual datasets). Although both the federated model and local models perform worse than local training on large datasets in (a), they outperform local training on small datasets in (b). The local models before aggregation outperform the federated model on all datasets.}
  \label{fig:fedpav-vs-lcoal}
\end{figure}

\textbf{Upper Bound of FedPav} We compare the performance of the models obtained from the FedPav algorithm with the \textit{local training}. According to previous discussion, $E = 1$ and $B = 32$ is the best setting of the FedPav algorithm. Thus, we use this setting for the FedPav algorithm. 

We summarize the results in Figure \ref{fig:fedpav-vs-lcoal}. Although the federated model performs worse than \textit{local training} on large datasets such as MSMT17 \cite{Wei2017Msmt} and Market-1501 \cite{Zheng2015Market1501} (Figure \ref{fig:fedpav-vs-local-a}), it outperforms \textit{local training} on smaller datasets such as CUHK01 \cite{li2012cuhk01} and VIPeR \cite{Gray2008ViewpointIP} (Figure \ref{fig:fedpav-vs-local-b}). These results suggest that the models trained on smaller datasets gain knowledge more effectively from other clients. Two reasons could explain these results: the models trained on larger datasets dominates in aggregation, so these clients absorb less knowledge from other clients; the models trained on small datasets have weaker generalization ability, so gaining more knowledge from larger datasets improves their ability.

The local models, models trained in clients before uploading to the server, is a proxy to measure the best performance of clients in FedReID. Server aggregation leads to performance decay for all datasets comparing the performance of local models and the federated model in Figure \ref{fig:fedpav-vs-lcoal}. It suggests that the server has the potential to better integrate the knowledge from the clients. Moreover, the \textit{local training} performs better than the local models in large datasets (Figure \ref{fig:fedpav-vs-local-a}), suggesting the bottlenecks of FedPav algorithm.

\begin{figure}[t]
  \centering
  \begin{subfigure}[b]{0.7\linewidth}
    \includegraphics[width=\linewidth]{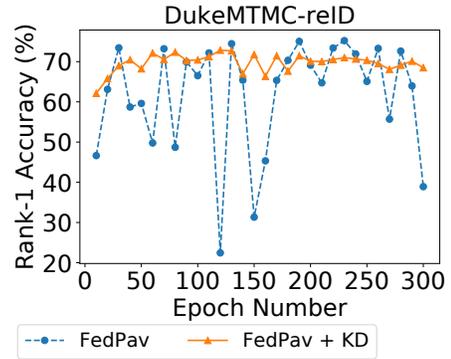}
    \caption{}
    \label{fig:convergence-kd-b}
  \end{subfigure}
  \begin{subfigure}[b]{0.7\linewidth}
    \includegraphics[width=\linewidth]{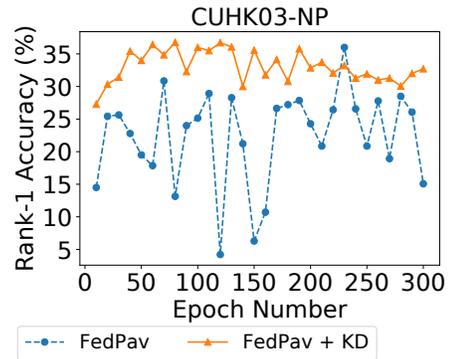}
    \caption{}
    \label{fig:convergence-kd-c}
  \end{subfigure}
  \caption{Convergence of FedPav with knowledge distillation (KD), with local epoch $E = 1$, and batch size $B = 32$, and evaluation computed every 10 rounds. (a) and (b) shows the convergence improvement on DukeMTMC-reID and CUHK03-NP.}
  \label{fig:convergence-kd}
\end{figure}

\textbf{Convergence of FedPav} The non-IID datasets affect the convergence of FedReID training. Figure \ref{fig:convergence-kd} shows rank-1 accuracy of the federated model trained by FedPav on DukeMTMC-reID \cite{zheng2017dukemtmc-reid} and CUHK03-NP \cite{Li2014CUHK03} in 300 rounds of communication, with evaluation computed every 10 rounds and fixing $E = 1$ and $B = 32$. The rank-1 accuracy of FedPav on both datasets fluctuates through the training process. The non-IID of 9 datasets in the benchmark causes the difficulty in convergence when aggregating the models from clients as Li et al. \cite{fedprox} stated the negative impact of non-IID data. To better measure the training performance, we average the performance of three best-federated models from different epochs in our experiments.

\section{Performance Optimization}
\label{sec:performance-optimization}

Based on the insights of benchmark analysis, we further investigate methods to optimize the performance of FedReID. We adopt knowledge distillation in Section \ref{sec:kd}, propose weight adjustment in Section \ref{sec:weight-adjustment}, and present the combination of these two methods in Section \ref{sec:weight-kd}.

\begin{algorithm}[t]
    \caption{FedPav with Knowledge Distillation}
    \label{algo:fedpav-kd}
    \SetAlgoLined
    \KwInput{$E, B, K, \eta, T, N, n_k, n, \mathcal{D}_{shared}$}
    \KwOutput{$w^T, w^T_k$}
    
    \SetKwProg{Fn}{Server}{:}{}
    \Fn{\FnServer}{
        initialize $w^0$\;
        distribute $\mathcal{D}_{shared}$ to clients\;
        \For{each round t = 0 to T-1}{
            $C_t \leftarrow$ (randomly select K out of N clients)\;
            \For{each client $k \in C_t$ concurrently}{
                $w^{t+1}_k, \ell^{t+1}_k \leftarrow $ \textbf{ClientExecution}($w^t$, $k$, $t$)\;
            }
            // $n$: total size of dataset; $n_k$: size of client $k$'s dataset\;
            $w^{t+1} \leftarrow \sum_{k \in C_t} \frac{n_k}{n} w^{t+1}_k$\;
            $\ell^{t+1} \leftarrow \frac{1}{K} \sum_{k \in C_t} \ell^{t+1}_k$\;
            $w^{t+1} \leftarrow $ (fine-tune $w^{t+1}$ with $\ell^{t+1}$ and $\mathcal{D}_{shared}$)
        }
        \KwRet $w^{T}$\;
    }
    
    \SetKwProg{Fn}{ClientExecution}{:}{\KwRet}
    \Fn{\FnClient{w, k, t}}{
        $v \leftarrow$ (retrieve additional layers $v$ if $t > 0$ else initialize)\;
        $\mathcal{B} \leftarrow $ (divide local data into batches of size $B$)\;
        \For{each local epoch e = 0 to E-1}{
            \For{$b \in \mathcal{B}$}{
                // $(w, v)$ concatenation two vectors\;
                $(w, v) \leftarrow (w, v) - \eta \triangledown \mathcal{L}((w, v); b)$\;
            }
        }
        store $v$\;
        $\ell \leftarrow$ (predict soft labels with $w$, $\mathcal{D}_{shared}$)\;
        \KwRet $w$, $\ell$\;
    }
\end{algorithm}

\subsection{Knowledge Distillation}
\label{sec:kd}

We apply knowledge distillation to the FedPav algorithm to improve its performance and convergence in this section. As discussed in Section \ref{sec:federated-by-datasets}, the FedPav algorithm has difficulty to converge and the local models perform better than the federated model. Knowledge distillation (KD) \cite{kd} is a method to transfer knowledge from one model (teacher model) to another model (student model). We adopt knowledge distillation to transfer the knowledge from clients to the server: each client is a teacher and the server is the student. 

To conduct knowledge distillation, we need a public dataset to generate soft labels from clients. We use an unlabelled CUHK02 \cite{cuhk02} dataset as an example to apply knowledge distillation to federated learning. CUHK02 \cite{cuhk02} dataset extends CUHK01 \cite{li2012cuhk01} dataset by four more pairs of camera views. It has 1816 identities in 7264 images. 
 
Algorithm \ref{algo:fedpav-kd} summarizes the training process with knowledge distillation: (1) At the beginning of the training, we distribute the CUHK02 \cite{cuhk02} dataset $\mathcal{D}_{shared}$ to all clients together with the initialized model $w^0$. (2) Each client uses the shared dataset $\mathcal{D}_{shared}$ to generate soft labels $\ell_k$ after training on their local datasets. These soft labels $\ell_k$ are features containing knowledge of the client model. (3) Each client uploads the model updates $w_k$ and the soft labels $\ell_k$ to the server. (4) The server averages these soft labels with $\ell = \frac{1}{K} \sum_{k \in C_t} \ell_k$. (5) The server trains the federated model $w$ using the shared dataset $\mathcal{D}_{shared}$ and the averaged soft labels $\ell$. The last step fine-tunes the federated model to mitigate the instability of aggregation and drives it for better convergence.

Figure \ref{fig:convergence-kd} compares the rank-1 accuracy performance of FedPav and FedPav with knowledge distillation on DukeMTMC-reID \cite{zheng2017dukemtmc-reid} dataset (Figure \ref{fig:convergence-kd-b}) and CUHK03-NP \cite{Li2014CUHK03} dataset (Figure \ref{fig:convergence-kd-c}). It shows that knowledge distillation reduces the volatility and helps the training to converge. However, knowledge distillation does not guarantee performance improvement: it improves the rank-1 accuracy on CUHK03-NP \cite{Li2014CUHK03}, while this advantage is unclear in DukeMTMC-reID \cite{zheng2017dukemtmc-reid} dataset. We hold the view that the domain distribution of the shared public dataset has a substantial impact on the final performance of the federated model on each dataset. The domain gap between CUHK02 \cite{cuhk02} dataset and CUHK03-NP \cite{Li2014CUHK03} dataset is smaller, so knowledge distillation elevates the performance on CUHK03-NP \cite{Li2014CUHK03} dataset significantly. We provide results of other datasets and mAP accuracy in the supplementary material.

\subsection{Weight Adjustment}
\label{sec:weight-adjustment}

In this section, we propose a method to adjust the weights of model aggregation to alleviate the unbalanced impact from huge differences in sizes of datasets. These weights in FedPav are proportional to the size of datasets: clients with large datasets like MSMT17 \cite{Wei2017Msmt} occupy around 40\% of total weights, while clients with small datasets like iLIDS-VID \cite{iLIDS-VID} have only 0.3\%, which is a negligible contribution to the federated model. Although clients with larger datasets are reasonable to have larger weights in aggregation, we anticipate that the huge discrepancy of weights between clients with small and large datasets hinders clients with large datasets from obtaining knowledge from other clients effectively. Therefore, we propose more appropriate weights for model aggregation. 

\begin{algorithm}[t]
    \caption{FedPav + Cosine Distance Weight}
    \label{algo:fedpav-weight}
    \SetAlgoLined
    \KwInput{$E, B, K, \eta, T, N$}
    \KwOutput{$w^T, w^T_k$}
    
    \SetKwProg{Fn}{Server}{:}{}
    \Fn{\FnServer}{
        initialize $w^0$\;
        \For{each round t = 0 to T-1}{
            $C_t \leftarrow$ (randomly select K out of N clients)\;
            \For{each client $k \in C_t$ concurrently}{
                $w^{t+1}_k, m_k \leftarrow $ \textbf{ClientExecution}($w^t$, $k$, $t$)\;
            }
            $m \leftarrow \sum_{k \in C_t} \frac{1}{m_k}$\;
            $w^{t+1} \leftarrow \sum_{k \in C_t} \frac{m_k}{m} w^{t+1}_k$\;
        }
        \KwRet $w^{T}$\;
    }
    
    \SetKwProg{Fn}{ClientExecution}{:}{}
    \Fn{\FnClient{$w$, k, t}}{
        $v \leftarrow$ (retrieve additional layers $v$ if $t > 0$ else initialize)\;
        $\mathcal{B} \leftarrow $ (divide local data into batches of size $B$)\;
        // $(w, v)$ concatenation of two vectors\;
        $(w^{t}, v^{t}) \leftarrow (w, v)$\;
        \For{each local epoch e = 0 to E-1}{
            \For{$b \in \mathcal{B}$}{
                $(w^{t}, v^{t}) \leftarrow (w^{t}, v^{t}) - \eta \triangledown \mathcal{L}((w^{t}, v^{t}); b)$\;
            }
        }
        
        $\mathcal{D}_{batch} \leftarrow$ one batch of $\mathcal{B}$\;
        \For{each data $d \in \mathcal{D}_{batch}$}{
            $f \leftarrow$ (generate logits with data and $(w, v)$)\;
            $f^{t} \leftarrow$ (generate logits with data and $(w^{t}, v^{t})$)\;
            $m_d \leftarrow 1 - cosine\_similarity(f, f^{t})$\;
        }
        $m^{t} \leftarrow \frac{1}{\lvert \mathcal{D}_{batch} \rvert} \sum_{d \in \mathcal{D}_{batch}} m_d$\;
        store $v^{t}$\;
        \KwRet $w^{t}, m^{t}$\;
    }
\end{algorithm}

\textbf{Cosine Distance Weight} We propose a method, \textit{Cosine Distance Weight} (CDW), to dynamically allocate weights depending on the changes of models: larger changes should contribute more (i.e. have larger weights) in model aggregation such that more newly learned knowledge can reflect in the federated model. We measure the model changes of each client $k$ by cosine distance in the following steps: (1) The client randomly selects a batch of training data $\mathcal{D}_{batch}$. (2) When the client receives model from the server in a new round of training $t$, it generates logits $f_k^t$ with $\mathcal{D}_{batch}$ and the local model $(w^t_k, v^t_k)$ formed by concatenation of the global model and local identity classifier. (3) The client conducts training to obtain a new model $(w^{t+1}_k, v^{t+1}_k)$. (4) It generates logits $f_k^{t+1}$ with $(w^{t+1}_k, v^{t+1}_k)$ and $\mathcal{D}_{batch}$. (5) The client computes the weight by averaging cosine distance of logits for each data point in the batch $m^{t+1}_k = mean(1 - cosine\_similarity(f_k^t, f_k^{t+1}))$. (6) The client sends $m_k^{t+1}$ to the server and the server replaces the weight in FedPav with it. We summarize this new algorithm in Algorithm \ref{algo:fedpav-weight}.

We experimented on FedPav with cosine distance weight under the same setting as Figure \ref{fig:fedpav-vs-lcoal}. Table \ref{tab:fedpav-cos-weight} shows that \textit{cosine distance weight} improves the performance significantly on all datasets. It demonstrates that we obtain a more holistic model that generalizes well in different domains. FedPav's \textit{local model}---the model before aggregation that has the best accuracy for local dataset---performs worse than \textit{local training} (the best accuracy of training individual datasets) on large datasets. However, \textit{local model} of FedPav with cosine similarity weight outperforms local training on all datasets. It indicates that all clients with different sizes of datasets are beneficial to participate in federated learning because they can obtain better-quality models comparing with the best models trained on their local datasets.

\begin{table}[h]
\caption{The increase in rank-1 accuracy comparing to local training. The local models of FedPav with cosine distance weight (CDW) outperform local training in all datasets.}
\begin{tabular}{ccccc}
\hline
\multicolumn{2}{l}{Datasets} & FedPav & CDW & CDW Local Model \\ \hline \hline
\multicolumn{2}{l}{MSMT17 \cite{Wei2017Msmt}} & -8.55 & -6.21 & \textbf{+4.01} \\
\multicolumn{2}{l}{DukeMTMC-reID \cite{zheng2017dukemtmc-reid}} & -5.77 & -2.90  & \textbf{+1.34}\\ 
\multicolumn{2}{l}{Market-1501 \cite{Zheng2015Market1501}} & -5.51 & -3.79  & \textbf{+1.43} \\
\multicolumn{2}{l}{CUHK03-NP \cite{Li2014CUHK03}}  & -17.58  & -15.05  & \textbf{+1.21}\\
\multicolumn{2}{l}{PRID2011 \cite{prid2011}} & -17.33 & -13.67 & \textbf{+7.33}\\
\multicolumn{2}{l}{CUHK01 \cite{li2012cuhk01}} & +4.32 & +9.30 & \textbf{+13.75}\\
\multicolumn{2}{l}{VIPeR \cite{Gray2008ViewpointIP}} & +20.57  & +20.36 & \textbf{+25.95}\\ 
\multicolumn{2}{l}{3DPeS \cite{3dpes}}  & +3.79 & +7.31 & \textbf{+16.26}\\
\multicolumn{2}{l}{iLIDS-VID \cite{iLIDS-VID}} & +27.89  & +28.57  & \textbf{+30.27}\\
\hline
\end{tabular}
\label{tab:fedpav-cos-weight}
\end{table}

\begin{figure}[h]
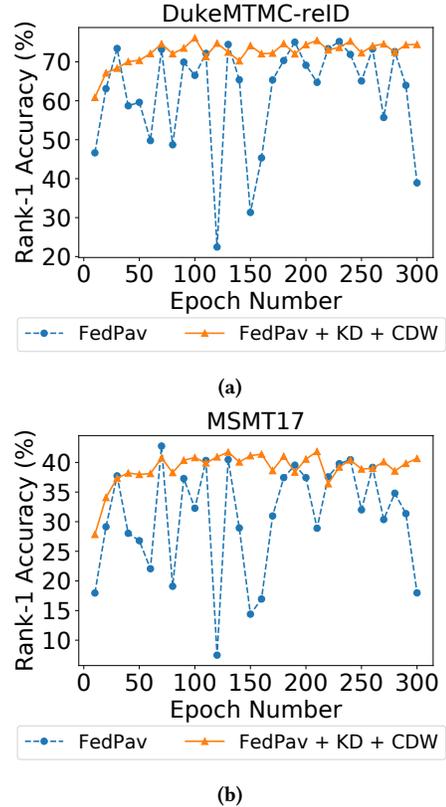

  \centering
  \begin{subfigure}[b]{0.7\linewidth}
    \includegraphics[width=\linewidth]{DukeMTMC-reID_CDW_Rank1.pdf}
    \caption{}
  \end{subfigure}
  \begin{subfigure}[b]{0.7\linewidth}
    \includegraphics[width=\linewidth]{MSMT17_CDW_Rank1.pdf}
    \caption{}
  \end{subfigure}
  \caption{Performance improvement on (a) DukeMTMC-reID and (b) MSMT17 by applying both knowledge distillation (KD) and cosine distance weight (CDW) to FedPav, with evaluation computed every 10 rounds.}
  \label{fig:weight-kd}
\end{figure}

\subsection{Knowledge Distillation and Weight Adjustment}
\label{sec:weight-kd}

In this section, we implement both dynamic weight adjustment and knowledge distillation to FedPav. We aim to achieve higher performance and better convergence with this combination by gaining advantages of both.

Figure \ref{fig:weight-kd} shows the performance of FedPav with knowledge distillation and cosine distance weight on two datasets. This combination improves the performance and the convergence of the training on these two datasets. We provide results of other datasets in the supplementary material.

\section{Conclusion}
\label{sec:conclusion}

In this paper, we investigated the statistical heterogeneity challenge of implementing federated learning to person re-identification, by performing benchmark analysis on a newly constructed benchmark that simulates the heterogeneous situation in the real-world scenario. This benchmark defines federated scenarios and introduces a federated learning algorithm---FedPav. The benchmark analysis presented bottlenecks and useful insights that are beneficial for future research and industrialization. Then we proposed two optimization methods to improve the performance of FedReID. To address the challenge of convergence, we adopted knowledge distillation to fine-tune the server model with knowledge generated from clients on an additional public dataset. To the elevate performance of large datasets, we dynamically adjusted the weights for model aggregation depending on the scale of model changes in clients. Numerical results indicate that these optimization methods can effectively facilitate convergence and achieve better performance. This paper focuses only on the statistical heterogeneity of FedReID in the real-world scenario. For future work, the system heterogeneity challenge will be taken into consideration.

\bibliographystyle{ACM-Reference-Format}
\bibliography{sample-base}

\newpage

\appendix

\section{Experiments}

\subsection{Experiment Settings}

We present the default experiment setting in this section. We used batch size $B = 32$, local epoch $E = 1$, and total rounds of training $T = 300$. In each client, the initialized learning rates were different for the identity classifier and the backbone: 0.05 for the identity classifier and 0.005 for the backbone, with step size 40 and gamma 0.1. The optimizer was set with weight decay 5e-4 and momentum 0.9. The learning rate for the server fine-tuning was 0.0005. If not specified, we conducted the experiments under this default setting.

\subsection{Impact of Batch Size}

Table \ref{tab:B=32}, \ref{tab:B=64}, and \ref{tab:B=128} show the performance of the federated model with batch size $B = 32$, $B = 64$, and $B = 128$ on rank-1 accuracy, rank-5 accuracy, rank-10 accuracy, and mAP accuracy. Figure \ref{fig:batch-size} shows the performance comparison of these three batch sizes measured by mAP accuracy. These experiments demonstrate the same conclusion as the paper.

\subsection{Impact of Local Epoch}

Figure \ref{fig:local-epoch} shows the mAP performance comparison of local epochs $E = 1$, $E = 5$, and $E = 10$.

\subsection{Local Training}

We report the performance of rank-1 accuracy, rank-5 accuracy, rank-10 accuracy, and mAP accuracy of \textit{local training} in Table \ref{tab:local-training}.

\subsection{FedPav with Cosine Distance Weight}

We report the performance of rank-1 accuracy, rank-5 accuracy, rank-10 accuracy, and mAP accuracy of FedPav with cosine distance weight (CDW) on all datasets in Table \ref{tab:fedpav-cdw}.

\subsection{FedPav with Knowledge Distillation}

\begin{figure}[h!]
  \centering
  \begin{subfigure}[b]{0.45\linewidth}
    \includegraphics[width=\linewidth]{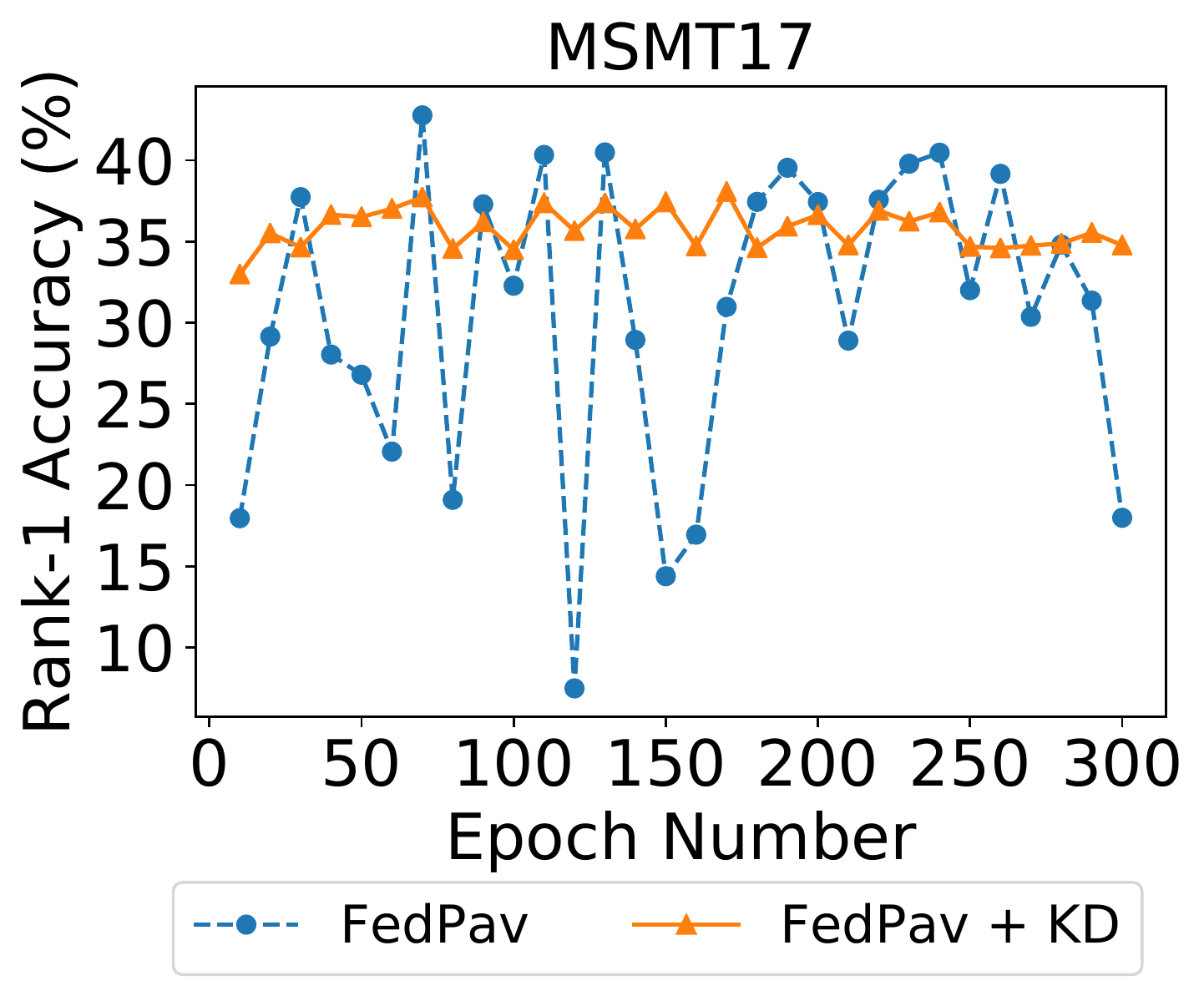}
    \caption{}
  \end{subfigure}
  \hspace{1em}%
  \begin{subfigure}[b]{0.45\linewidth}
    \includegraphics[width=\linewidth]{DukeMTMC-reID_Normal_Rank1.pdf}
    \caption{}
  \end{subfigure}
  
  \centering
  \begin{subfigure}[b]{0.45\linewidth}
    \includegraphics[width=\linewidth]{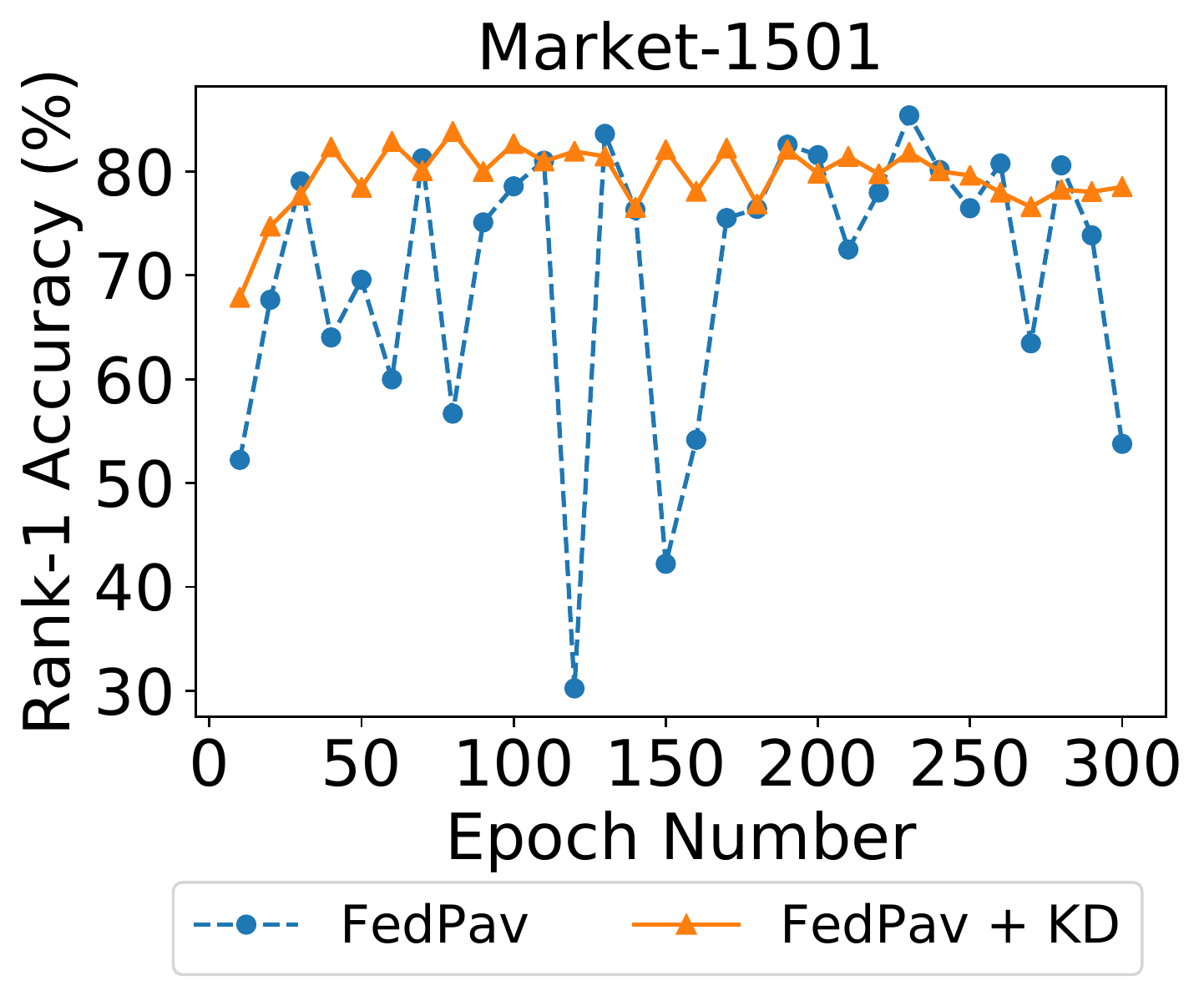}
    \caption{}
  \end{subfigure}
  \hspace{1em}%
  \begin{subfigure}[b]{0.45\linewidth}
    \includegraphics[width=\linewidth]{CUHK03-NP_Normal_Rank1.pdf}
    \caption{}
  \end{subfigure}
  
  \centering
  \begin{subfigure}[b]{0.45\linewidth}
    \includegraphics[width=\linewidth]{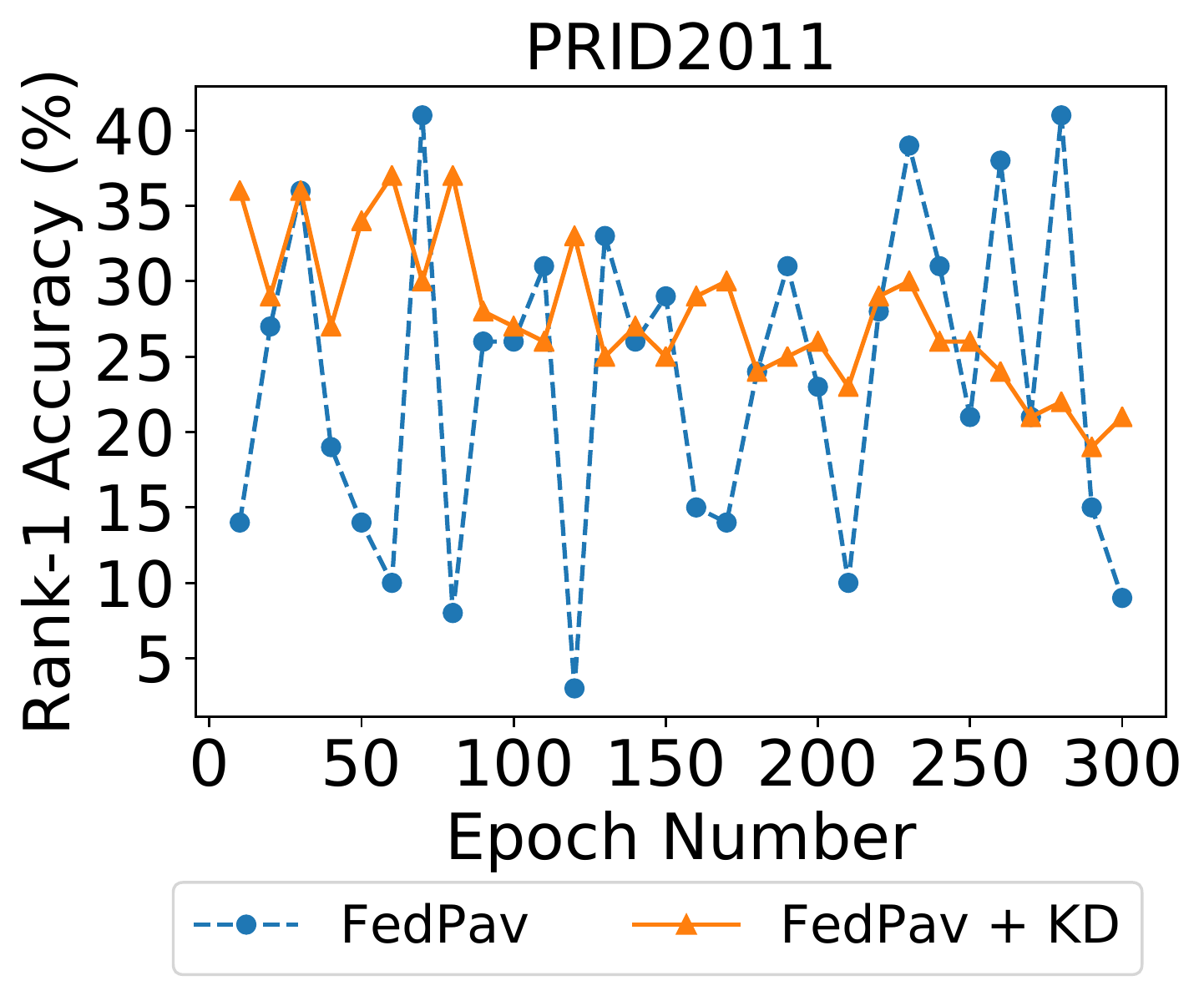}
    \caption{}
  \end{subfigure}
  \hspace{1em}%
  \begin{subfigure}[b]{0.45\linewidth}
    \includegraphics[width=\linewidth]{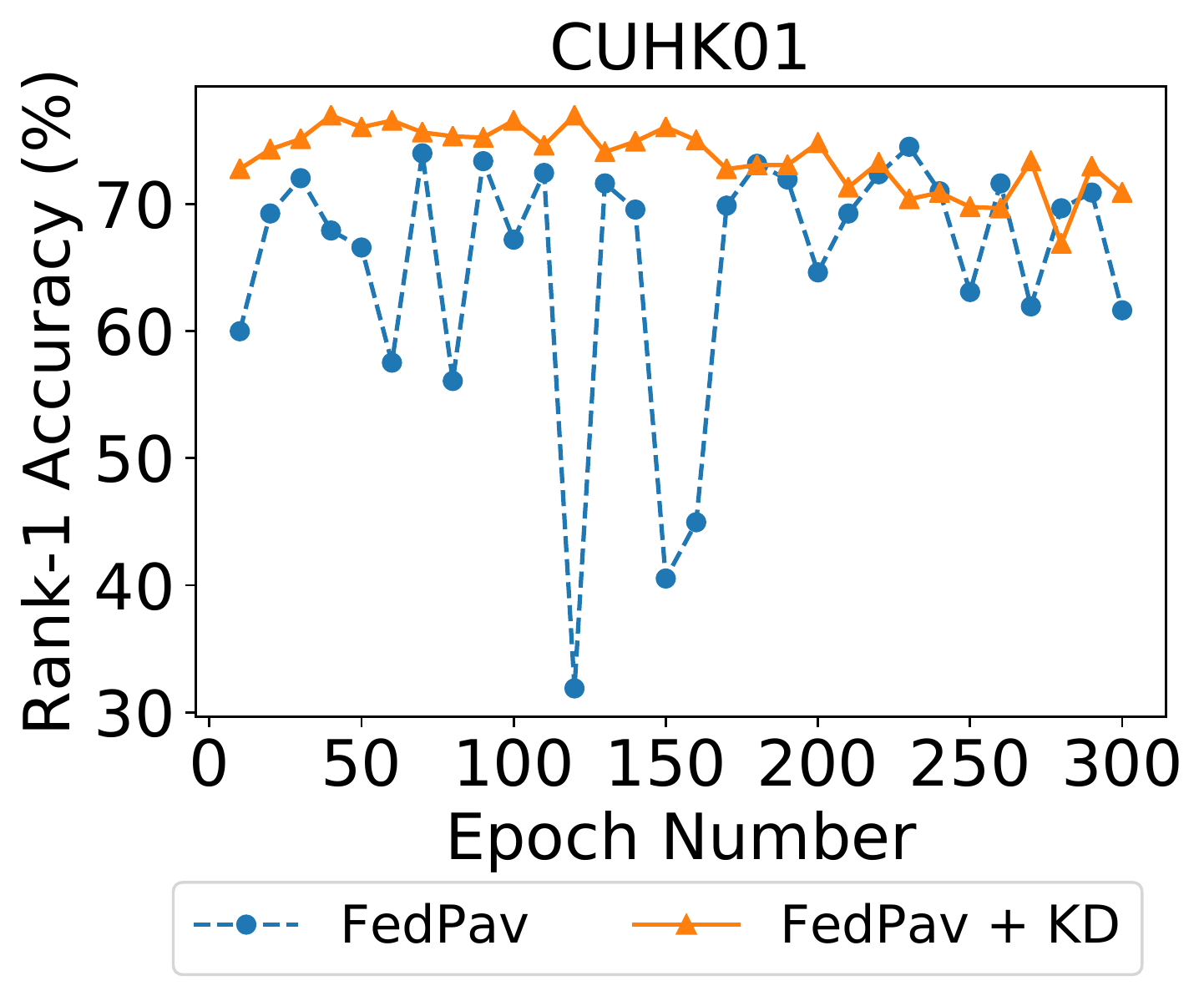}
    \caption{}
  \end{subfigure}
  
  \centering
  \begin{subfigure}[b]{0.45\linewidth}
    \includegraphics[width=\linewidth]{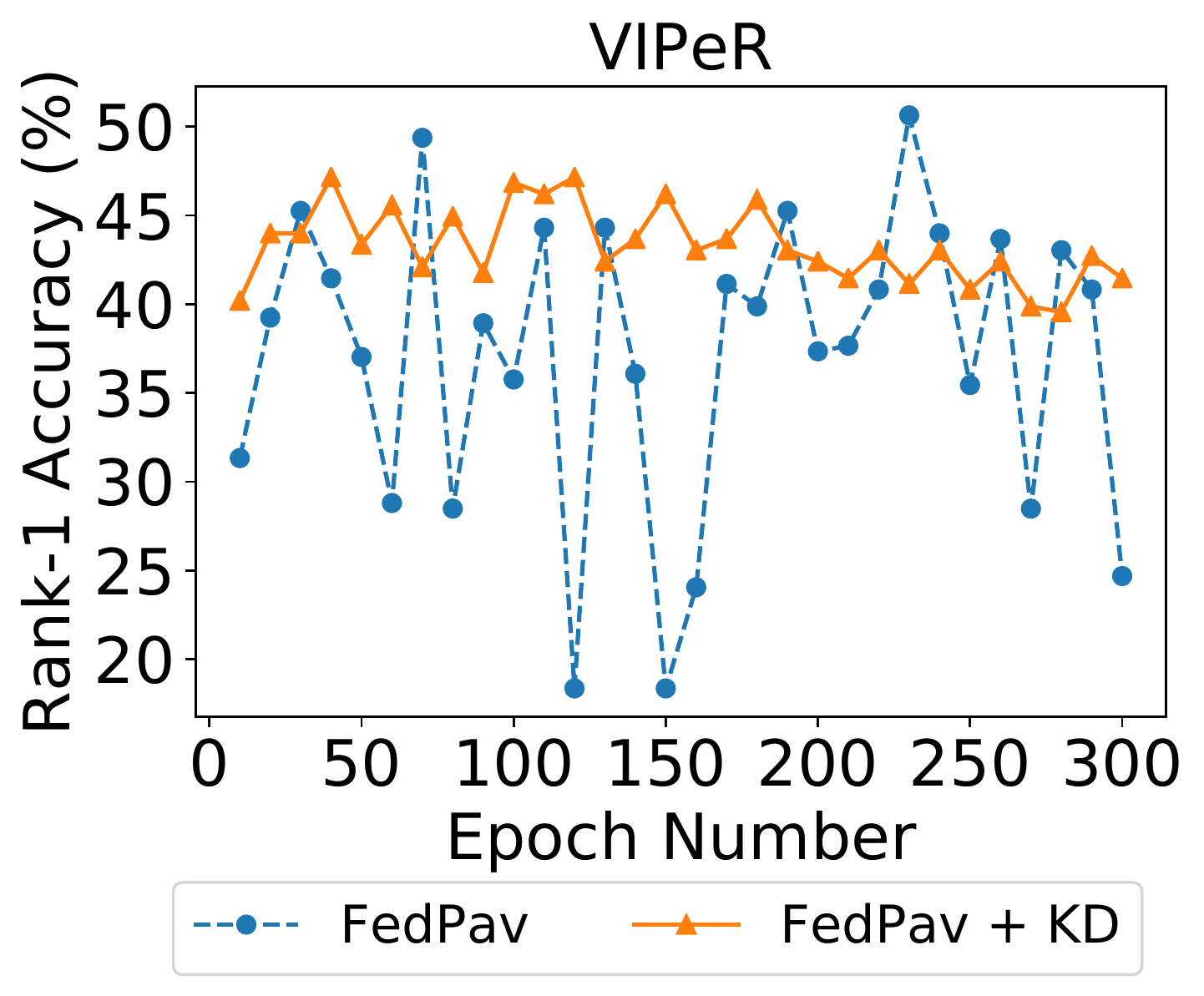}
    \caption{}
  \end{subfigure}
  \hspace{1em}%
  \begin{subfigure}[b]{0.45\linewidth}
    \includegraphics[width=\linewidth]{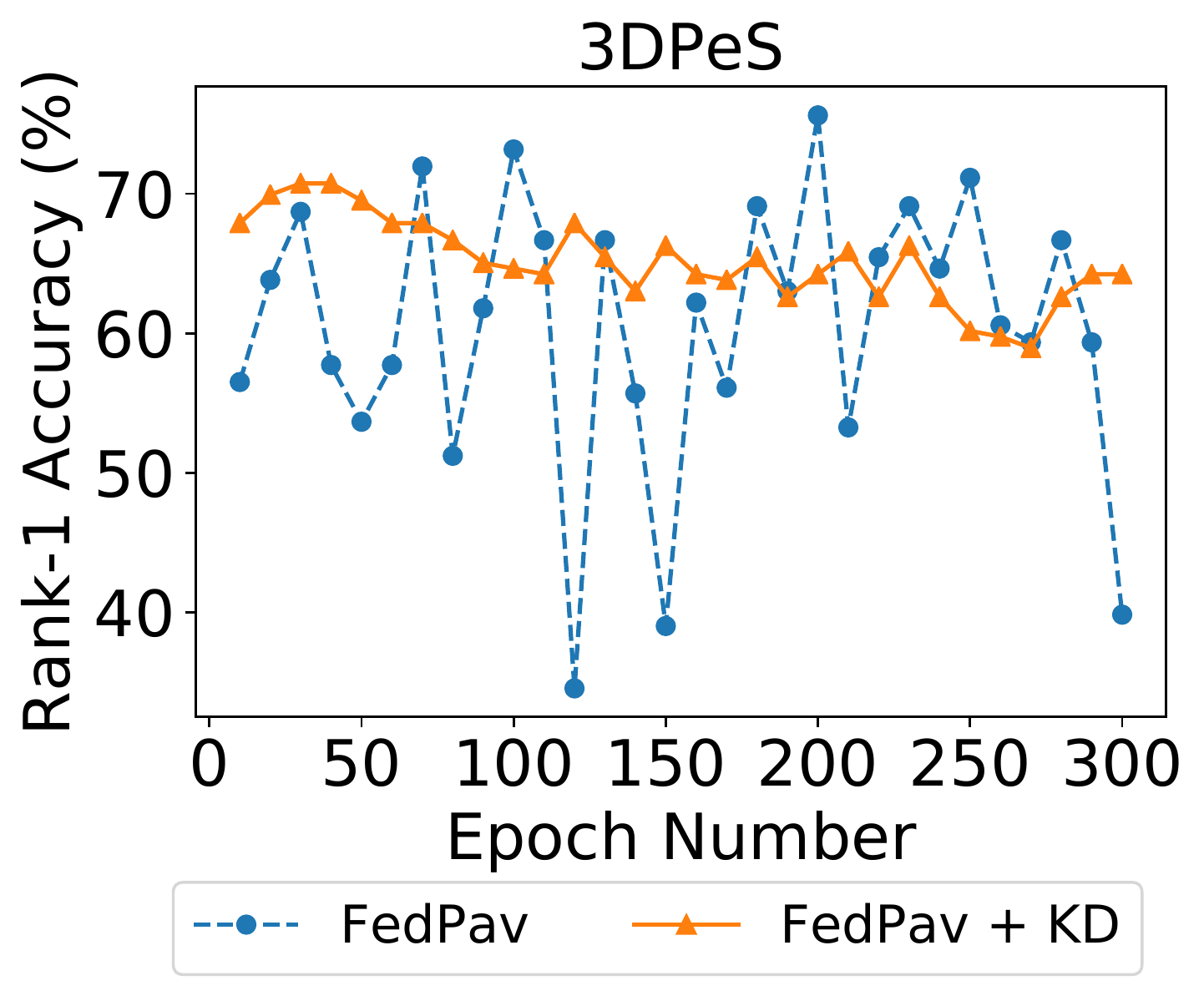}
    \caption{}
  \end{subfigure}
  
  \centering
  \begin{subfigure}[b]{0.45\linewidth}
    \includegraphics[width=\linewidth]{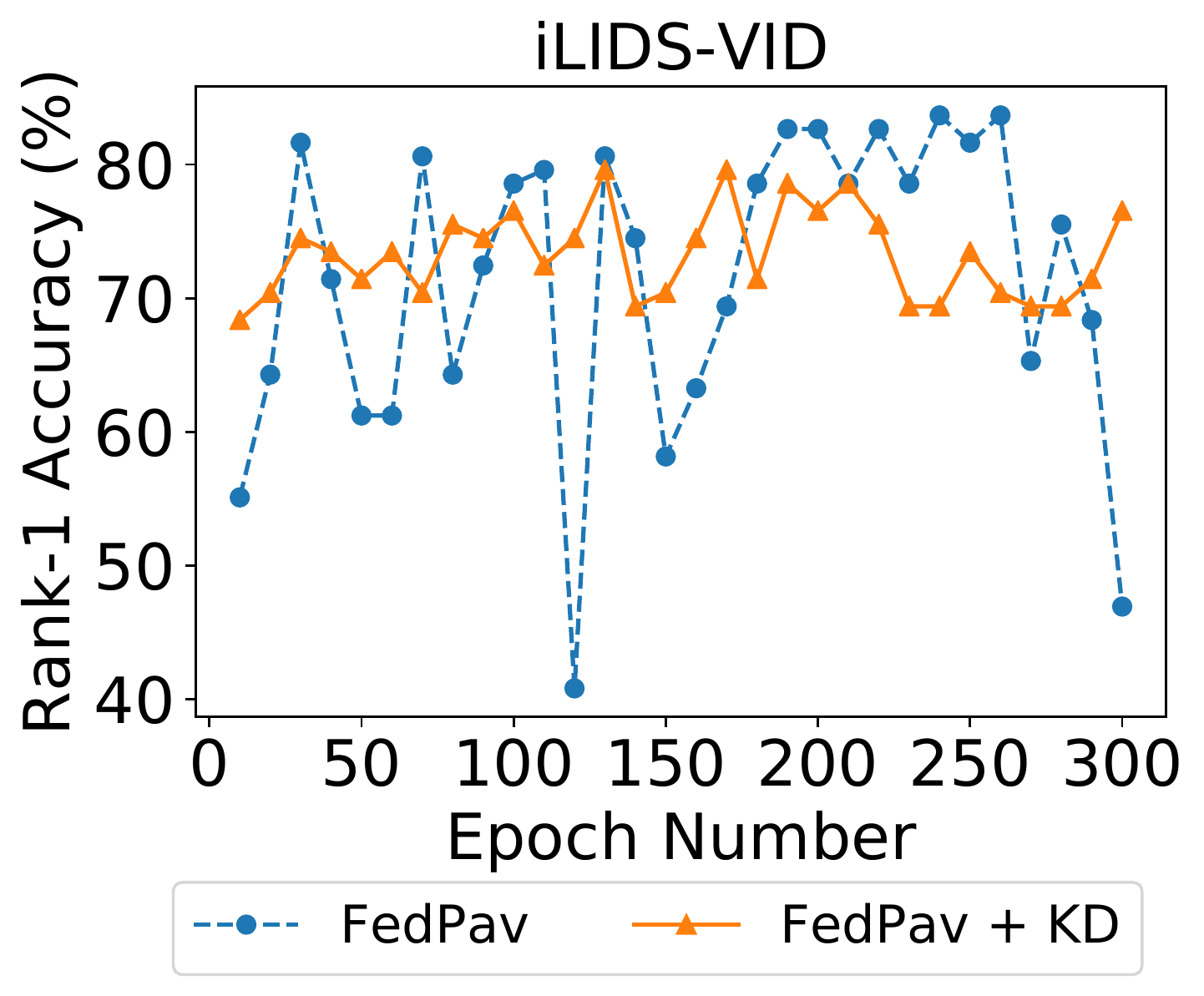}
    \caption{}
  \end{subfigure}
  \caption{Performance and convergence comparison of FedPav and FedPav with knowledge distillation (KD) in all datasets, measured by rank-1 accuracy.}
  \label{fig:fedpav-kd-rank1}
  \end{figure}

\begin{figure}[h!]
  \centering
  \begin{subfigure}[b]{0.45\linewidth}
    \includegraphics[width=\linewidth]{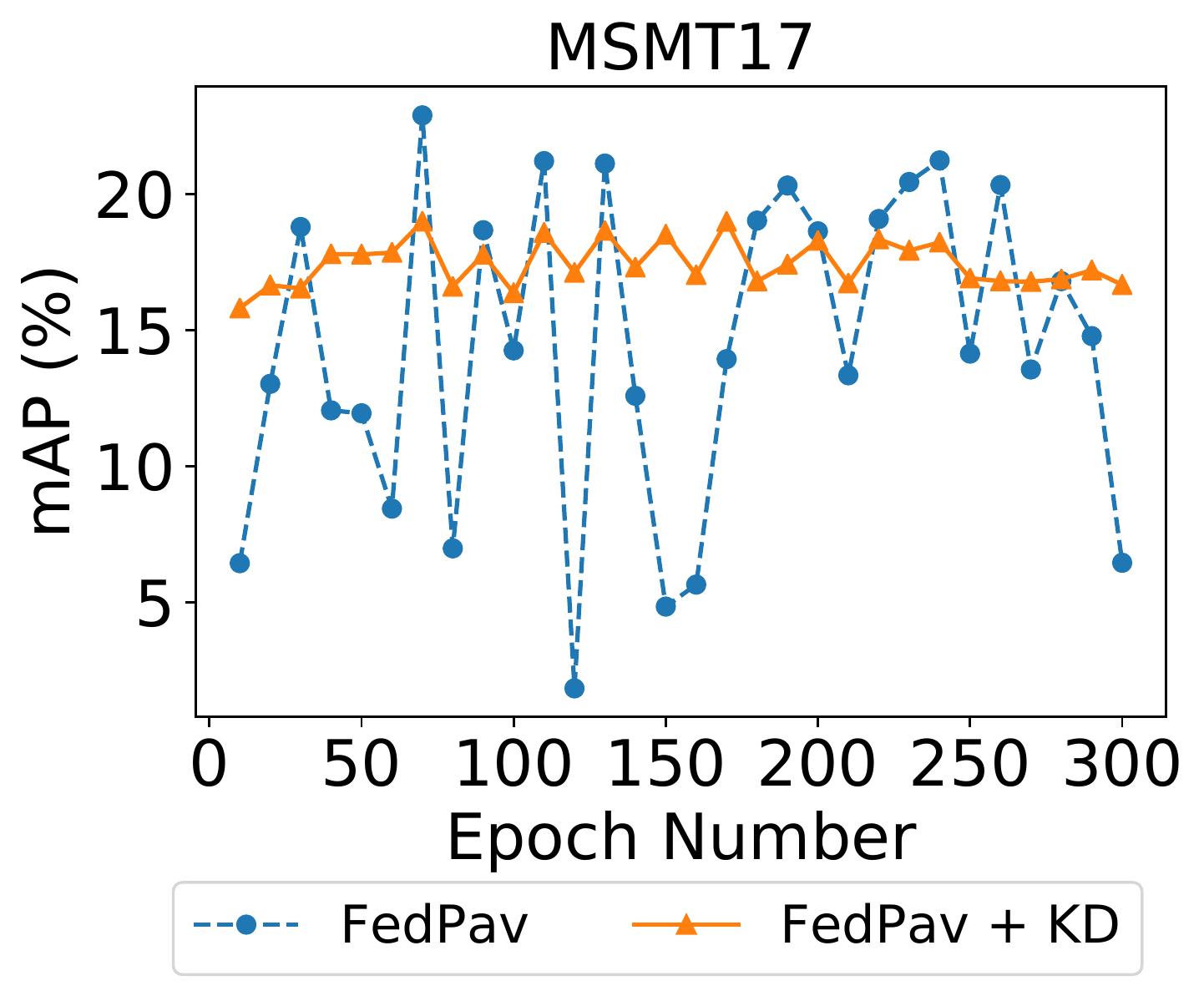}
    \caption{}
  \end{subfigure}
  \hspace{1em}%
  \begin{subfigure}[b]{0.45\linewidth}
    \includegraphics[width=\linewidth]{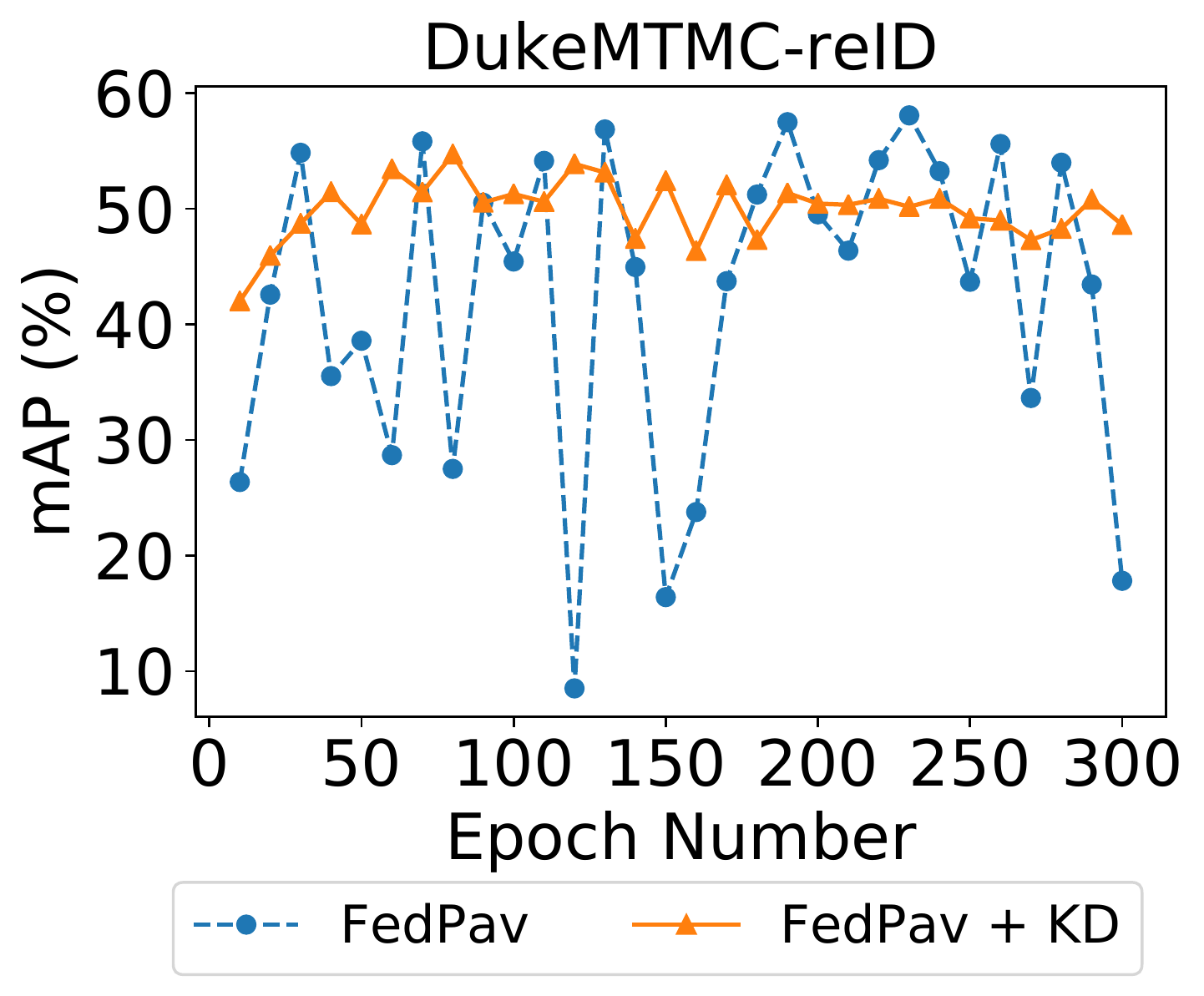}
    \caption{}
  \end{subfigure}
 
  \centering
  \begin{subfigure}[b]{0.45\linewidth}
    \includegraphics[width=\linewidth]{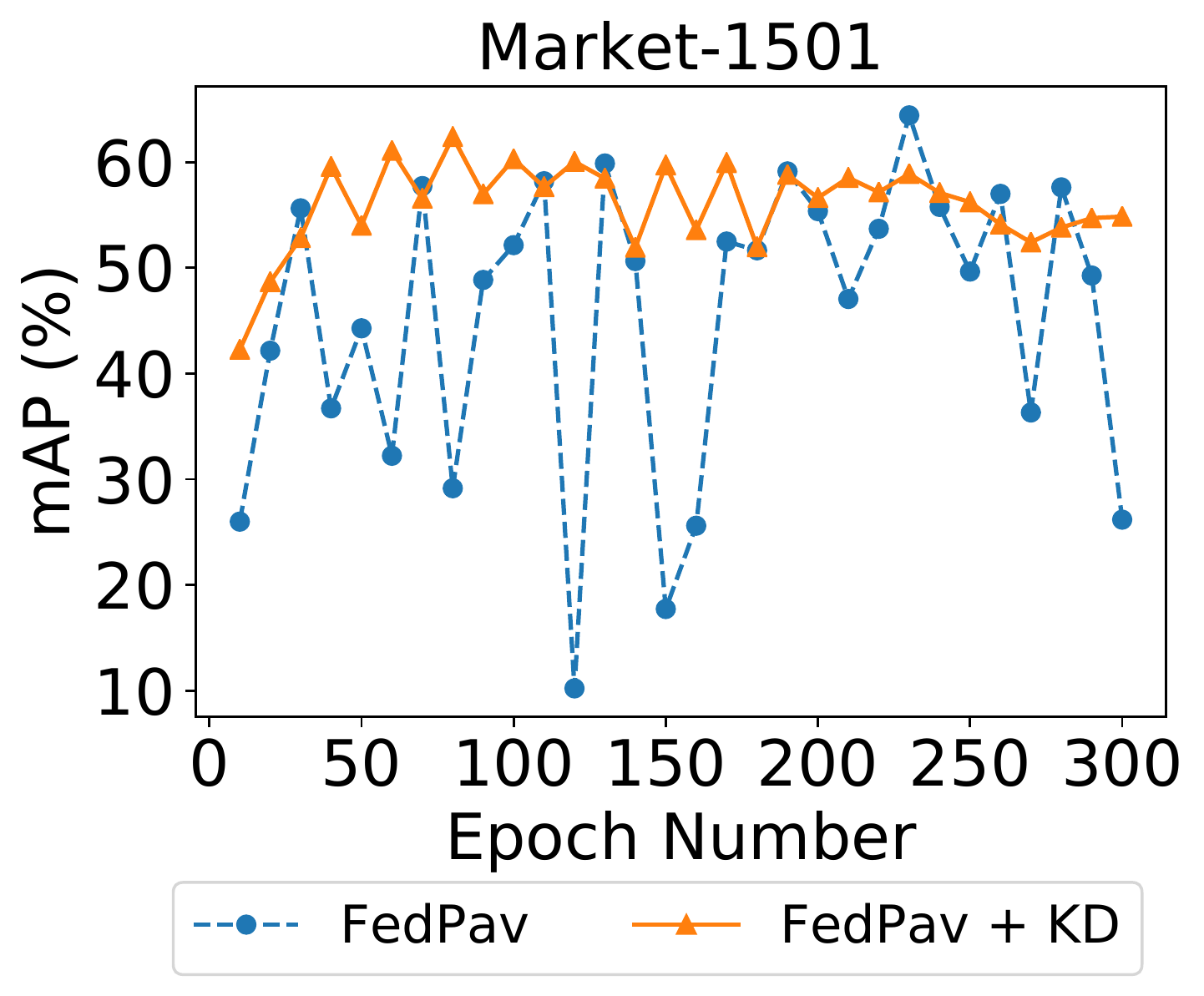}
    \caption{}
  \end{subfigure}
  \hspace{1em}%
  \begin{subfigure}[b]{0.45\linewidth}
    \includegraphics[width=\linewidth]{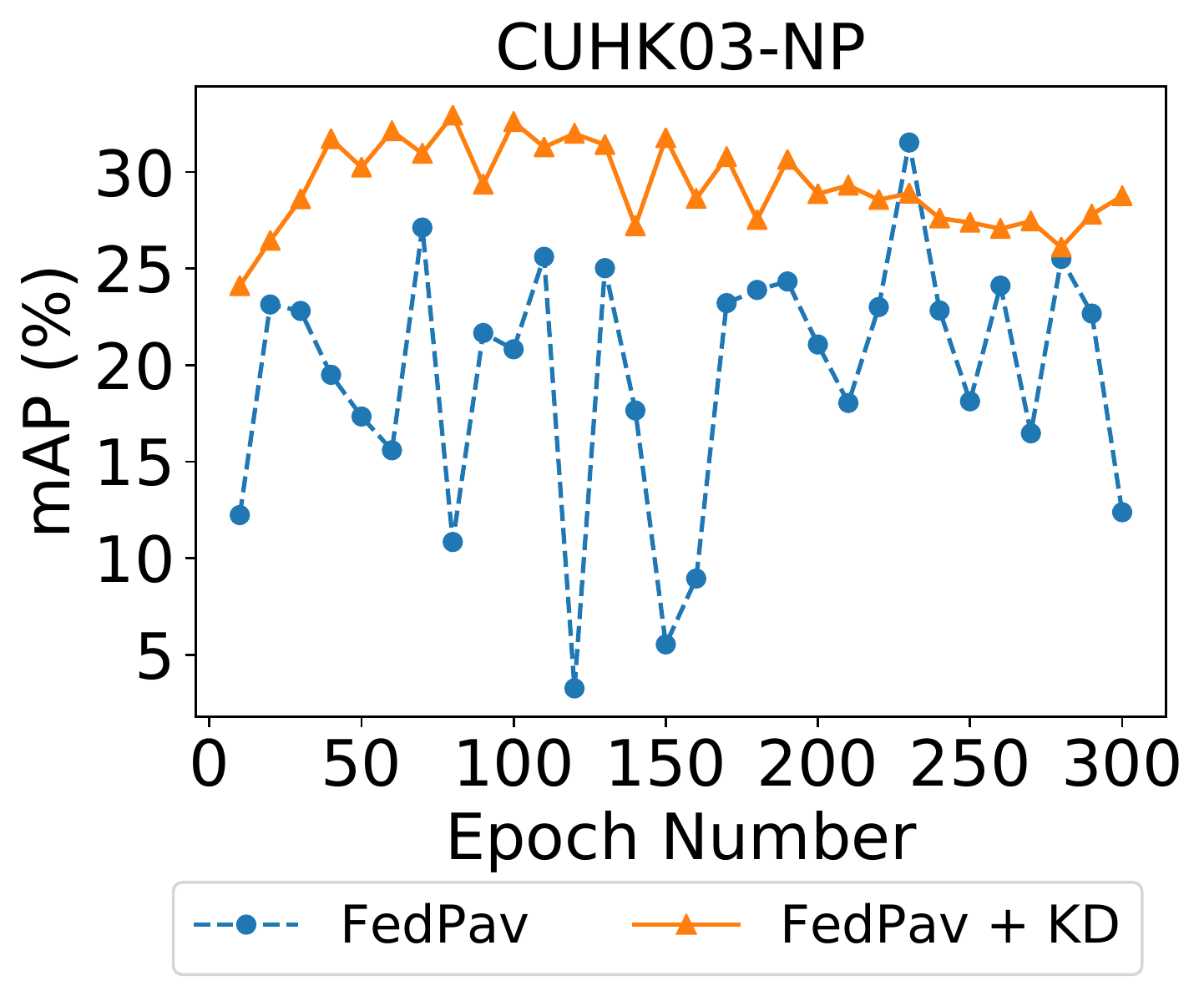}
    \caption{}
  \end{subfigure}
  
  \centering
  \begin{subfigure}[b]{0.45\linewidth}
    \includegraphics[width=\linewidth]{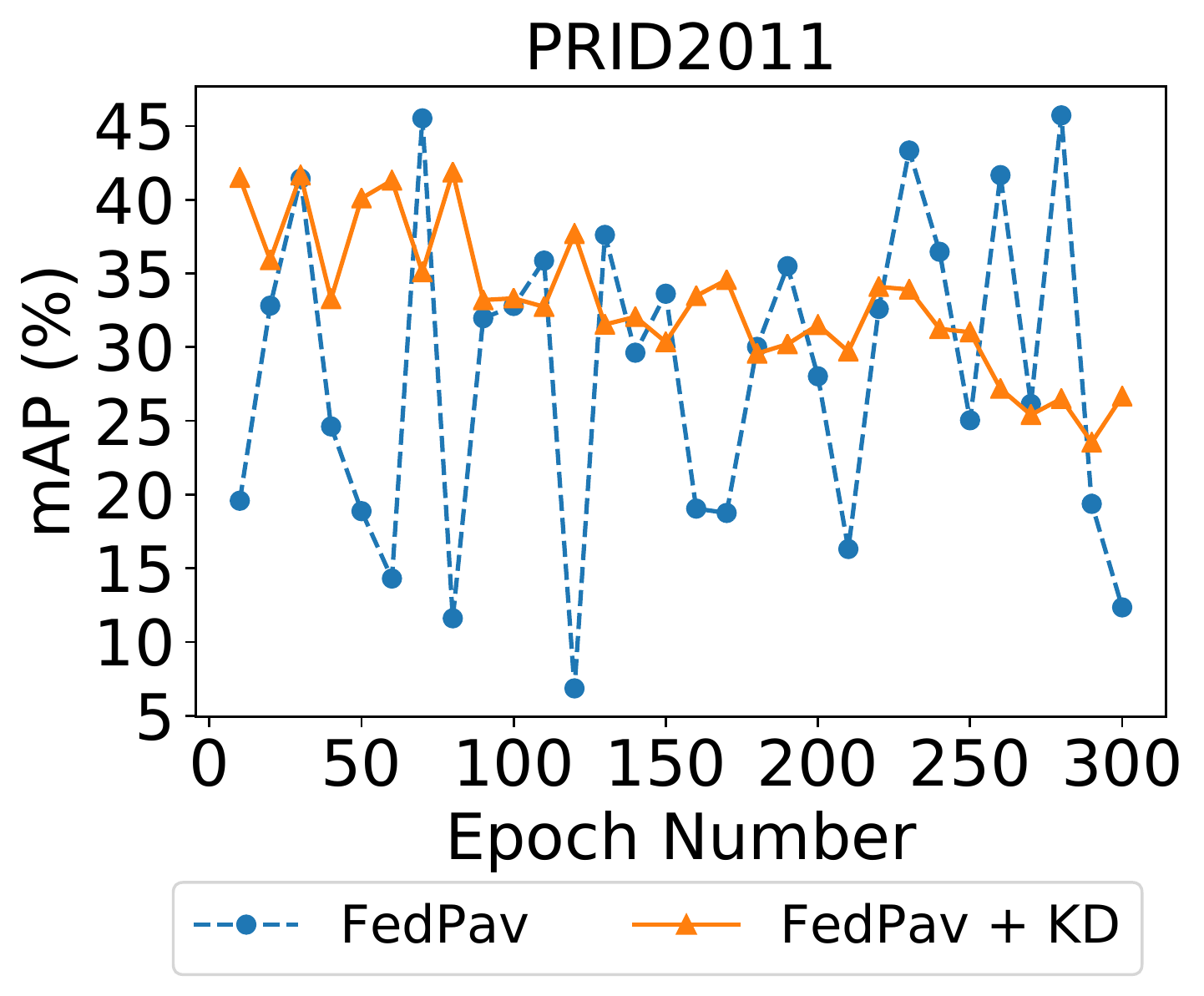}
    \caption{}
  \end{subfigure}
  \hspace{1em}%
  \begin{subfigure}[b]{0.45\linewidth}
    \includegraphics[width=\linewidth]{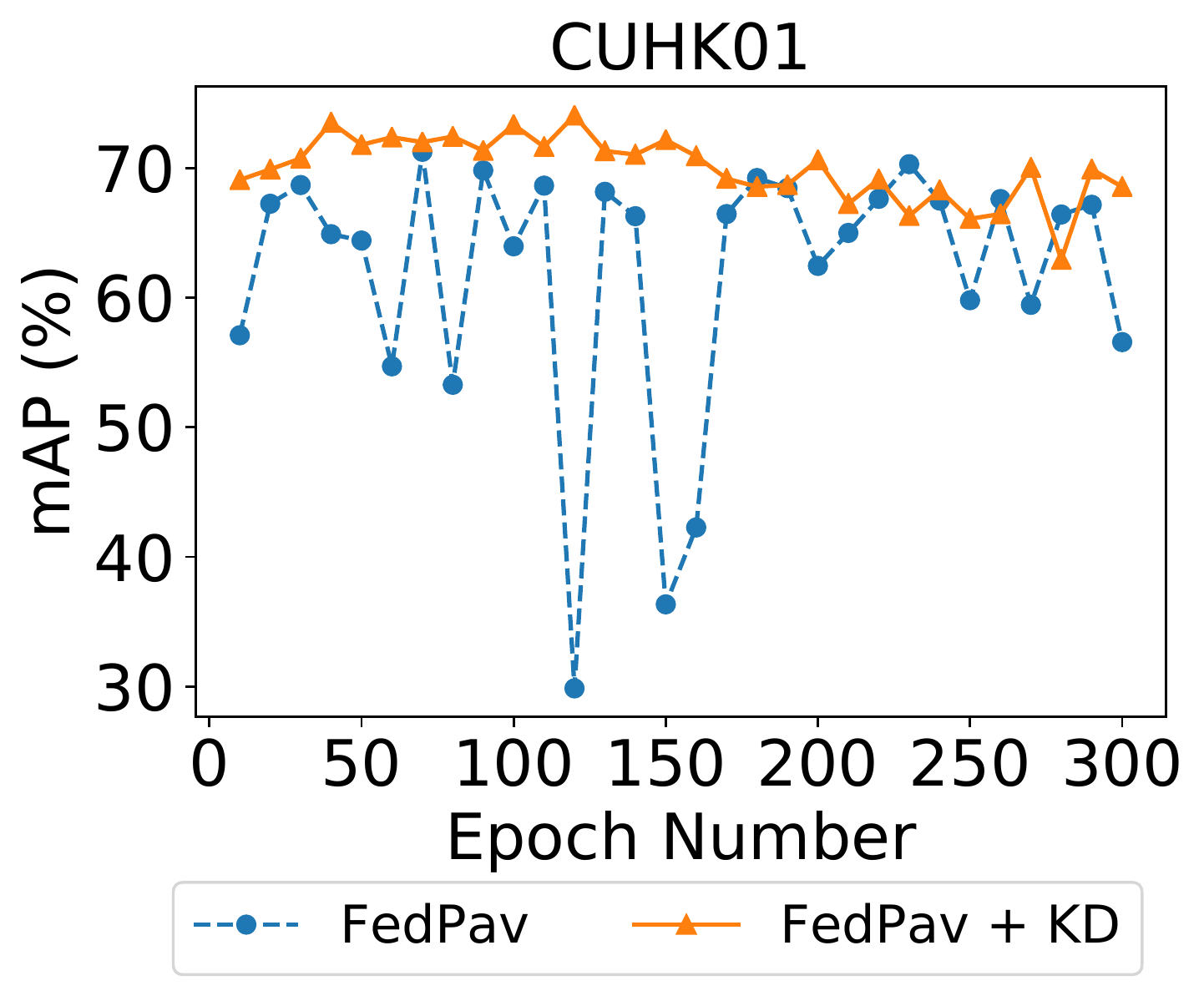}
    \caption{}
  \end{subfigure}
  
   \centering
  \begin{subfigure}[b]{0.45\linewidth}
    \includegraphics[width=\linewidth]{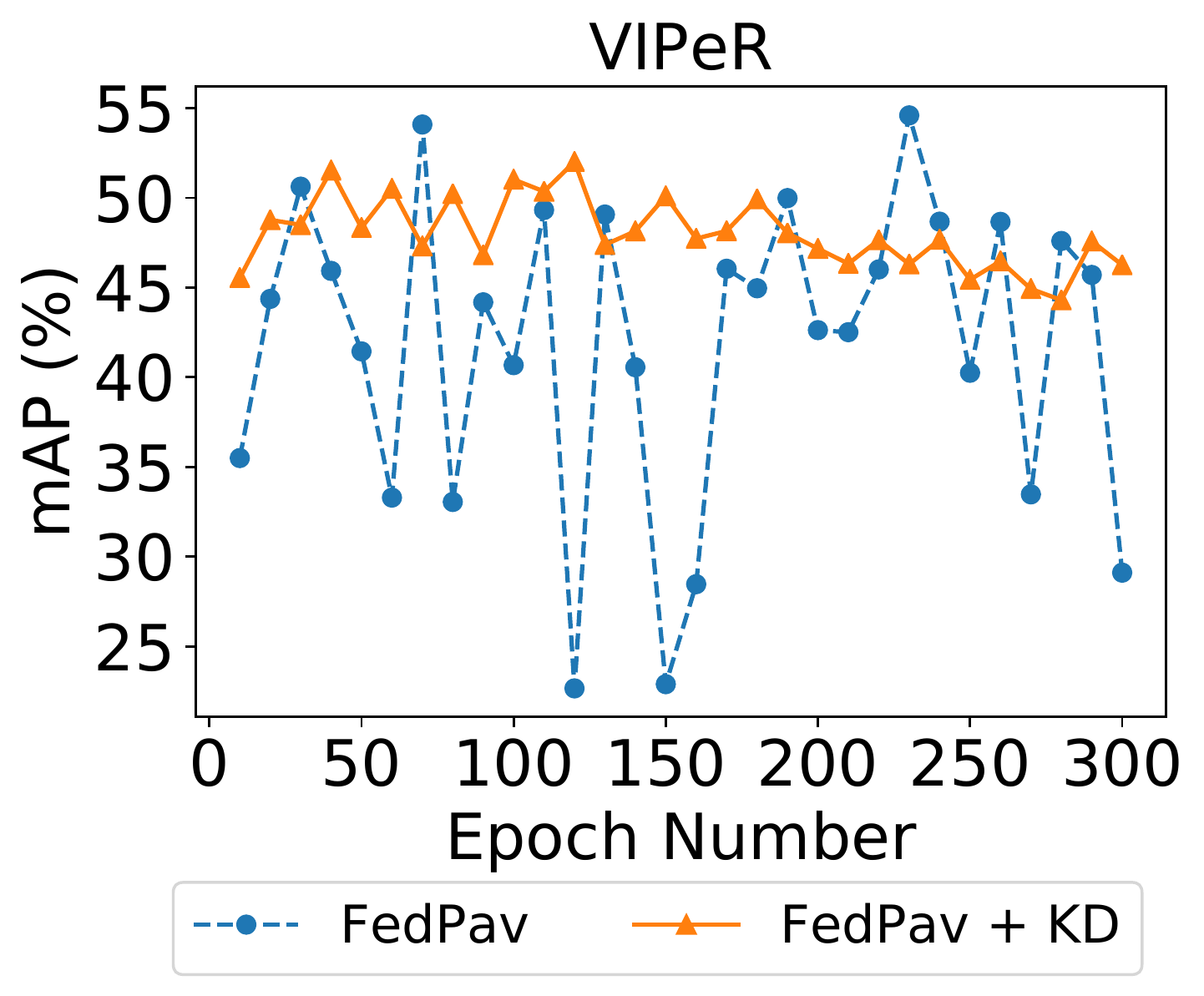}
    \caption{}
  \end{subfigure}
  \hspace{1em}%
  \begin{subfigure}[b]{0.45\linewidth}
    \includegraphics[width=\linewidth]{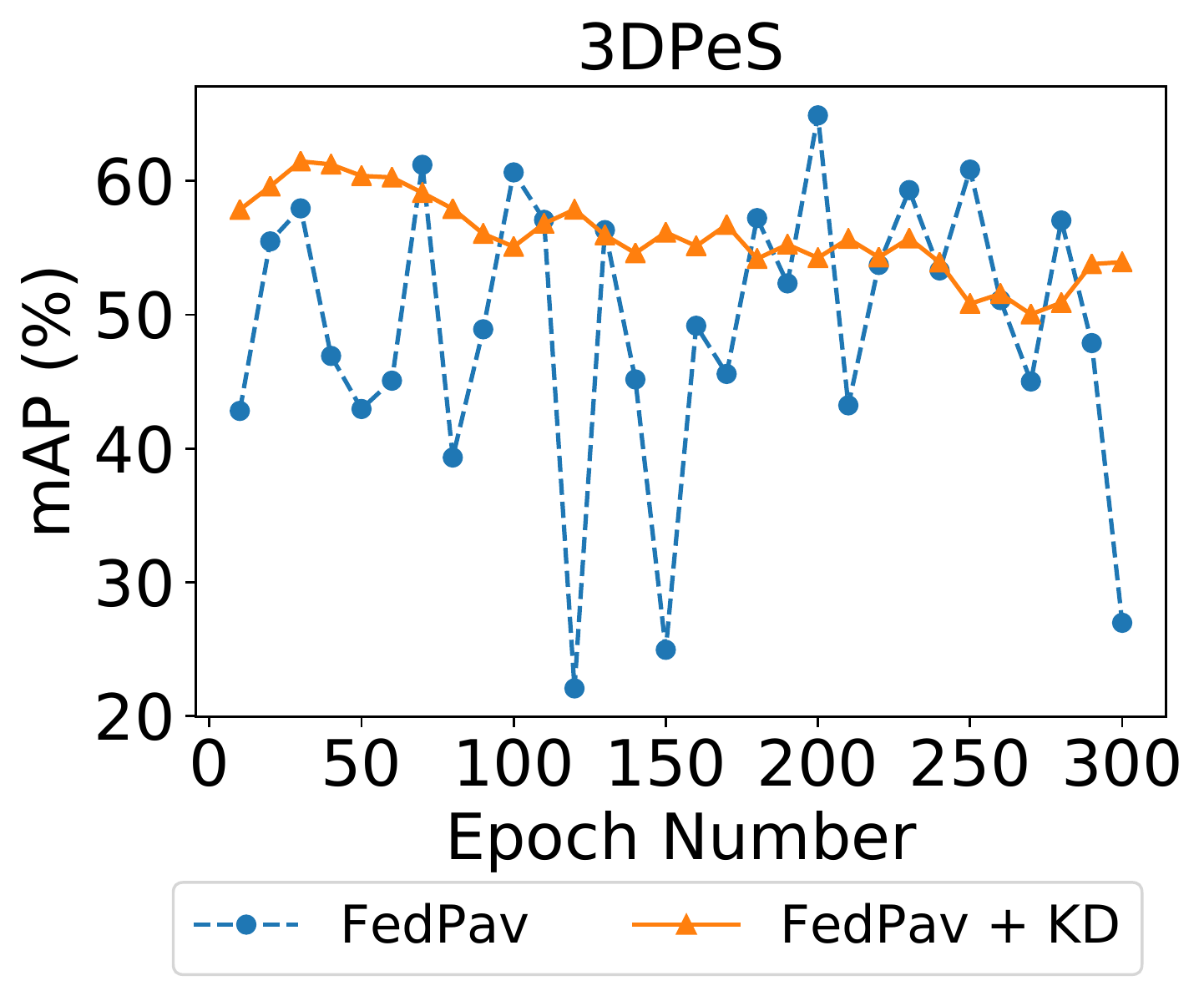}
    \caption{}
  \end{subfigure}
  
  \centering
  \begin{subfigure}[b]{0.45\linewidth}
    \includegraphics[width=\linewidth]{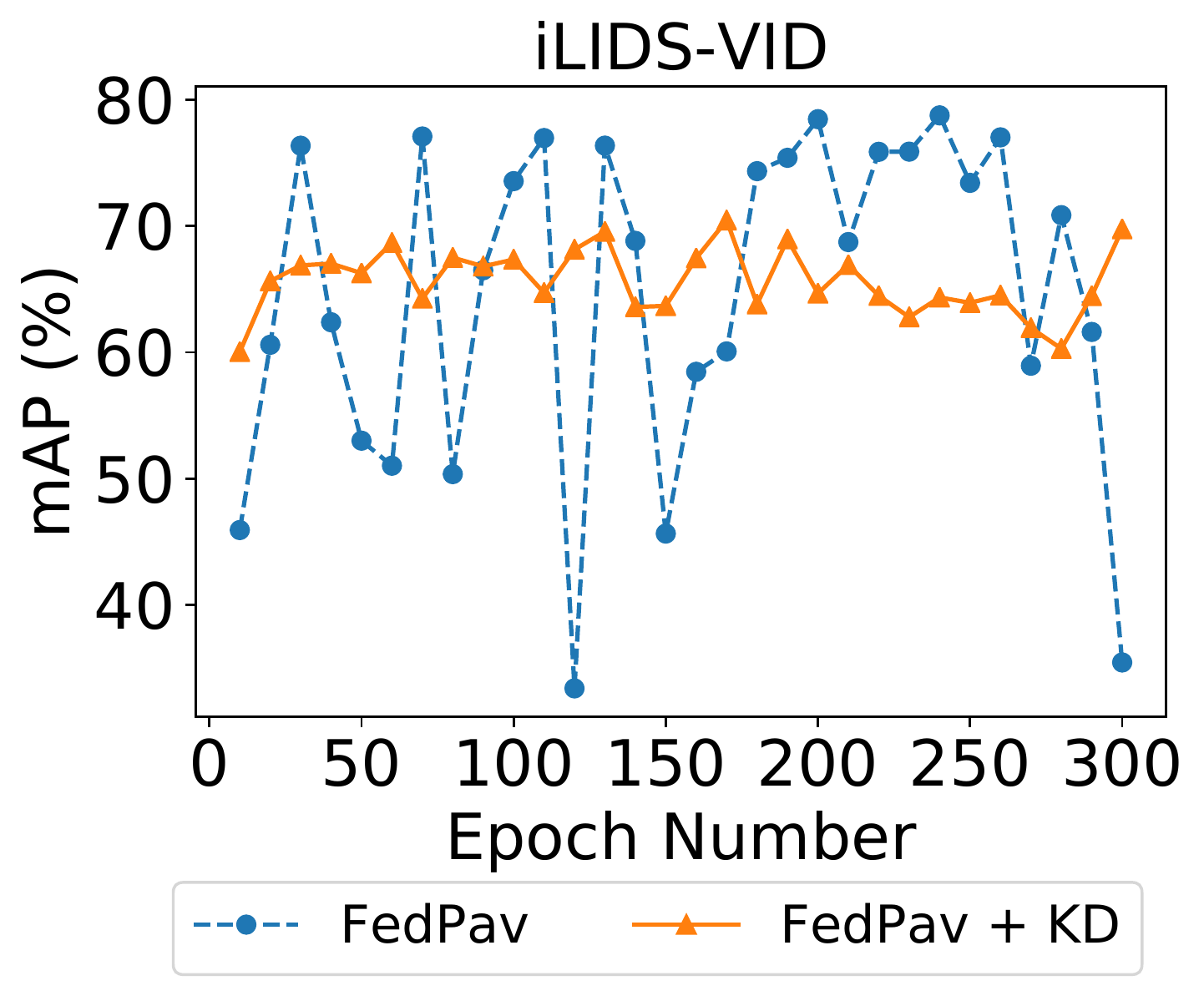}
    \caption{}
  \end{subfigure}
  \caption{Performance and convergence comparison of FedPav and FedPav with knowledge distillation (KD) in all datasets, measured by mAP accuracy.}
  \label{fig:fedpav-kd-map}
 
\end{figure}

Figure \ref{fig:fedpav-kd-rank1} and \ref{fig:fedpav-kd-map} show the performance and convergence comparison of FedPav and FedPav with knowledge distillation (KD) measured by rank-1 accuracy and mAP accuracy respectively. The rank-1 accuracy and mAP have similar patterns. As reported in the paper, although it does not guarantee performance improvement, it effectively facilitates the convergence of FedReID training.

\subsection{FedPav with KD and CDW}

\begin{figure}[h!]
  \centering
  \begin{subfigure}[b]{0.45\linewidth}
    \includegraphics[width=\linewidth]{MSMT17_CDW_Rank1.pdf}
    \caption{}
  \end{subfigure}
  \hspace{1em}%
  \begin{subfigure}[b]{0.45\linewidth}
    \includegraphics[width=\linewidth]{DukeMTMC-reID_CDW_Rank1.pdf}
    \caption{}
  \end{subfigure}
 
  \centering
  \begin{subfigure}[b]{0.45\linewidth}
    \includegraphics[width=\linewidth]{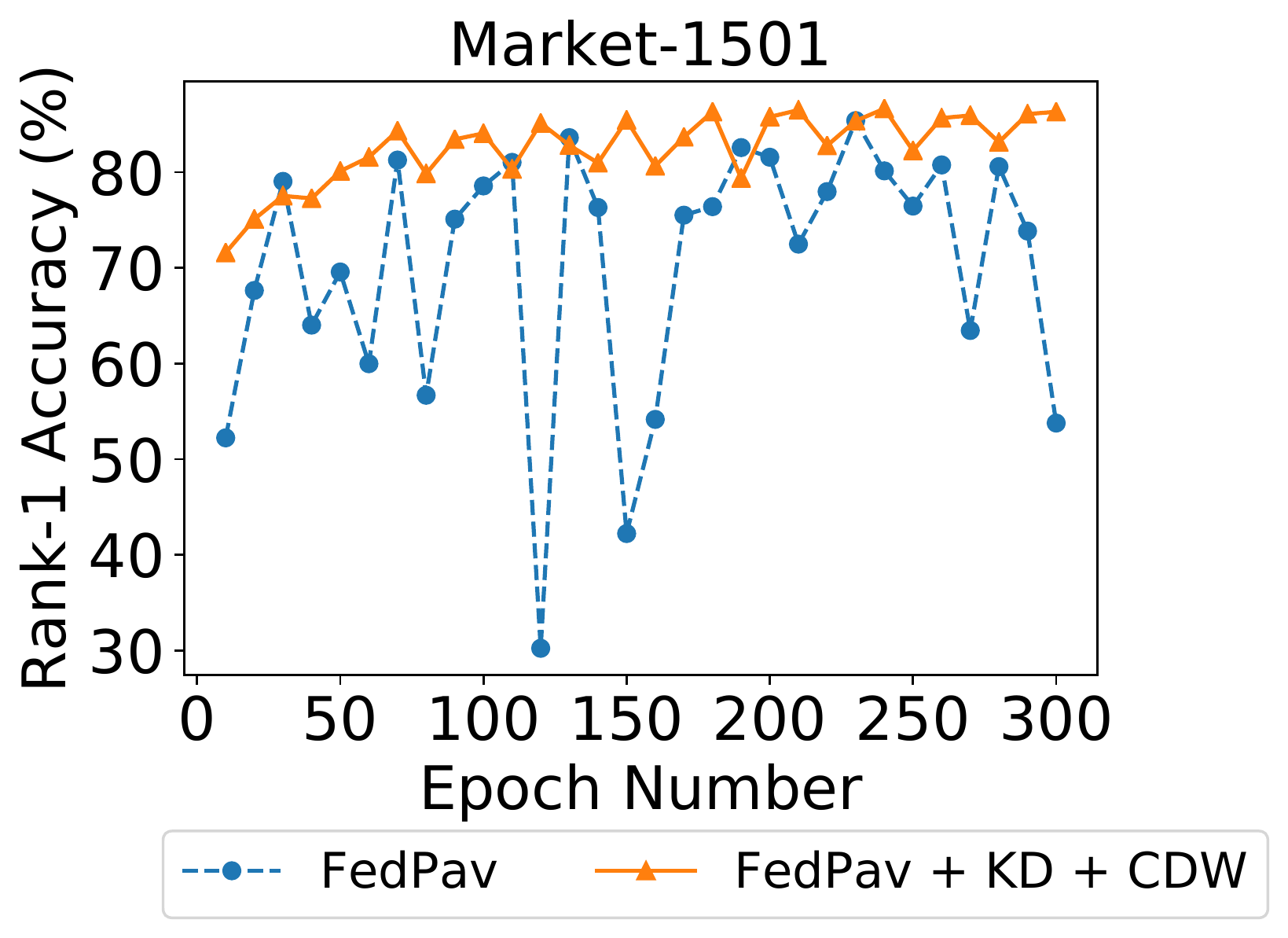}
    \caption{}
  \end{subfigure}
  \hspace{1em}%
  \begin{subfigure}[b]{0.45\linewidth}
    \includegraphics[width=\linewidth]{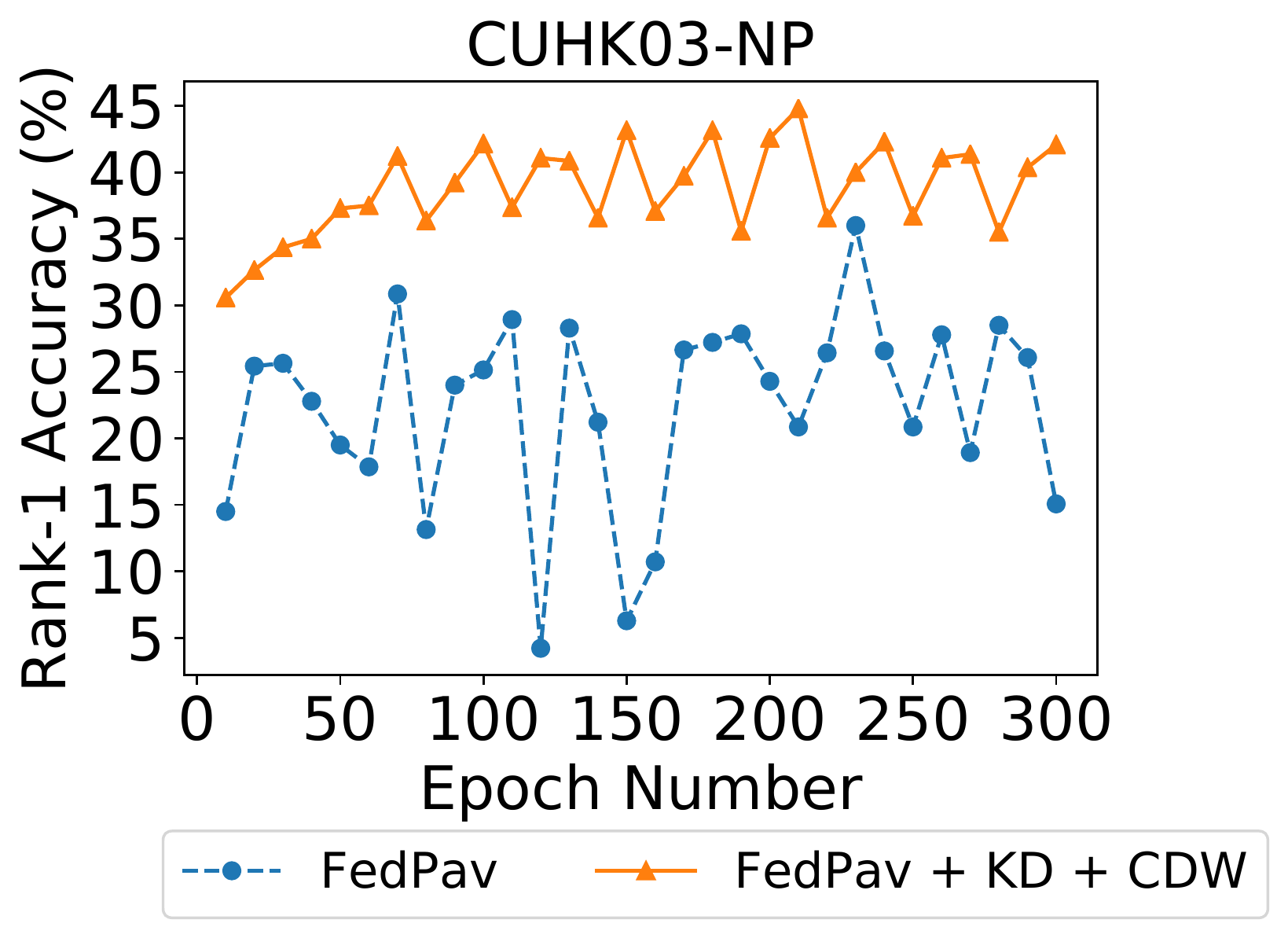}
    \caption{}
  \end{subfigure}
  
  \centering
  \begin{subfigure}[b]{0.45\linewidth}
    \includegraphics[width=\linewidth]{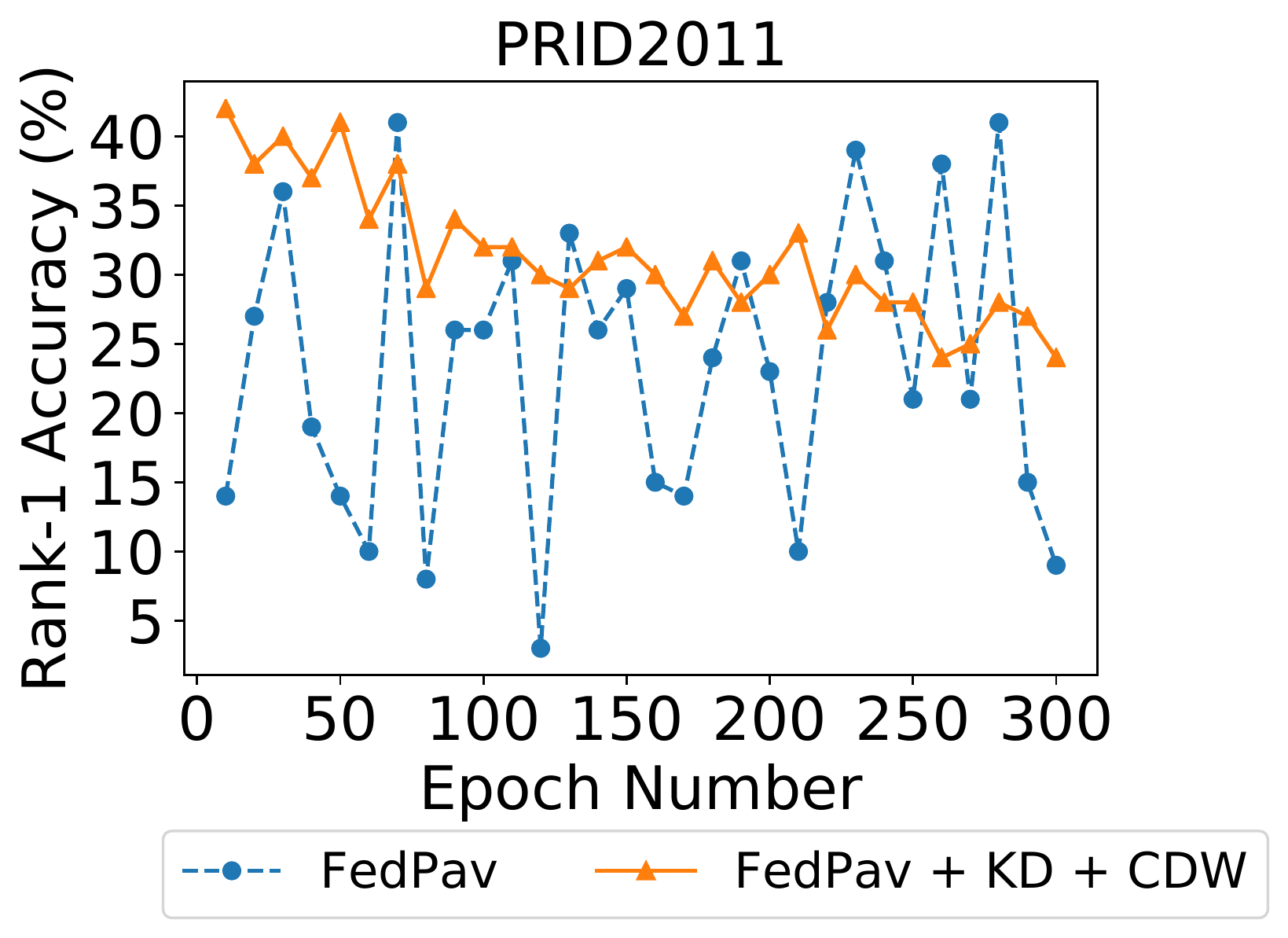}
    \caption{}
  \end{subfigure}
  \hspace{1em}%
  \begin{subfigure}[b]{0.45\linewidth}
    \includegraphics[width=\linewidth]{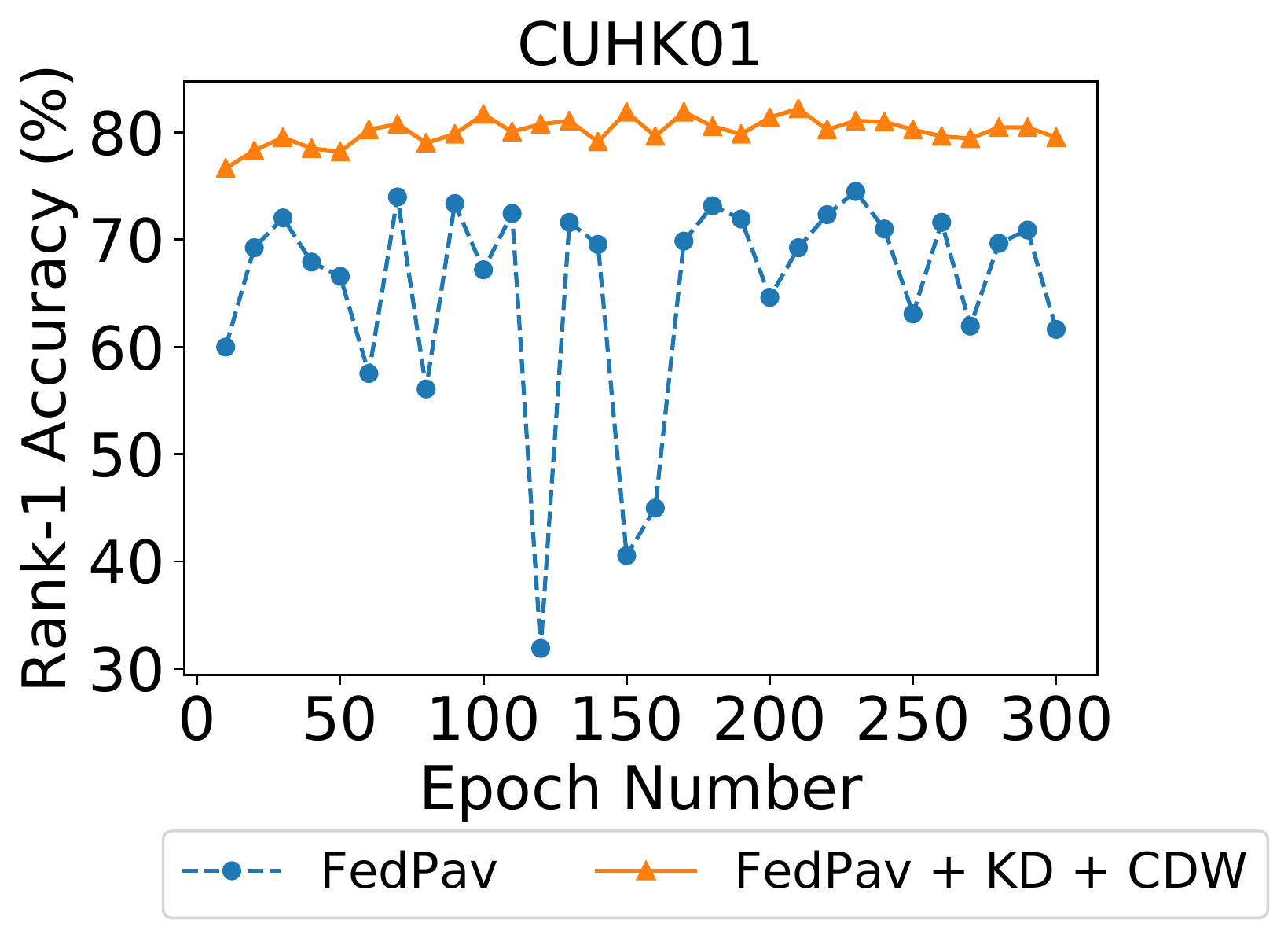}
    \caption{}
  \end{subfigure}
  
   \centering
  \begin{subfigure}[b]{0.45\linewidth}
    \includegraphics[width=\linewidth]{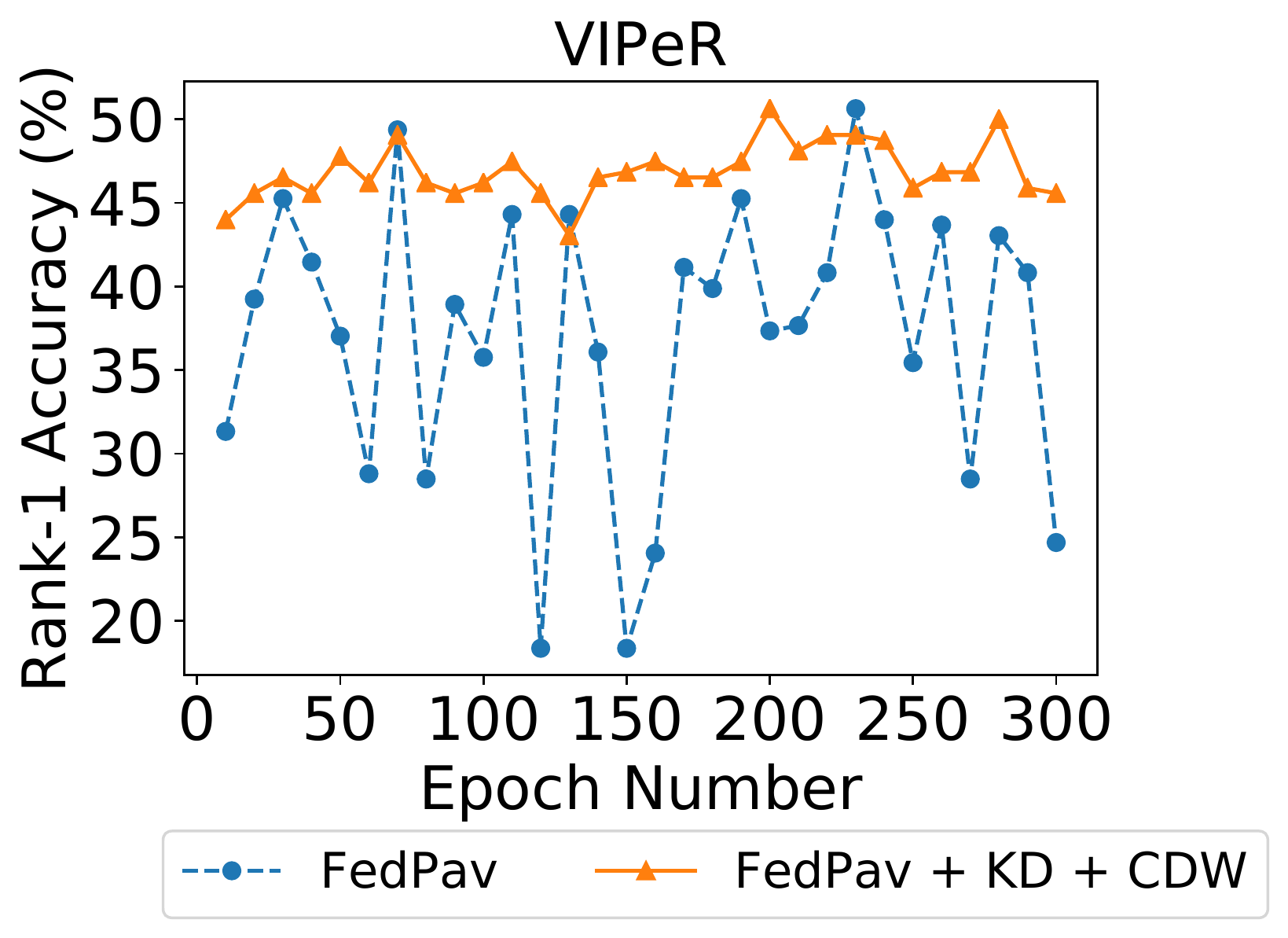}
    \caption{}
  \end{subfigure}
  \hspace{1em}%
  \begin{subfigure}[b]{0.45\linewidth}
    \includegraphics[width=\linewidth]{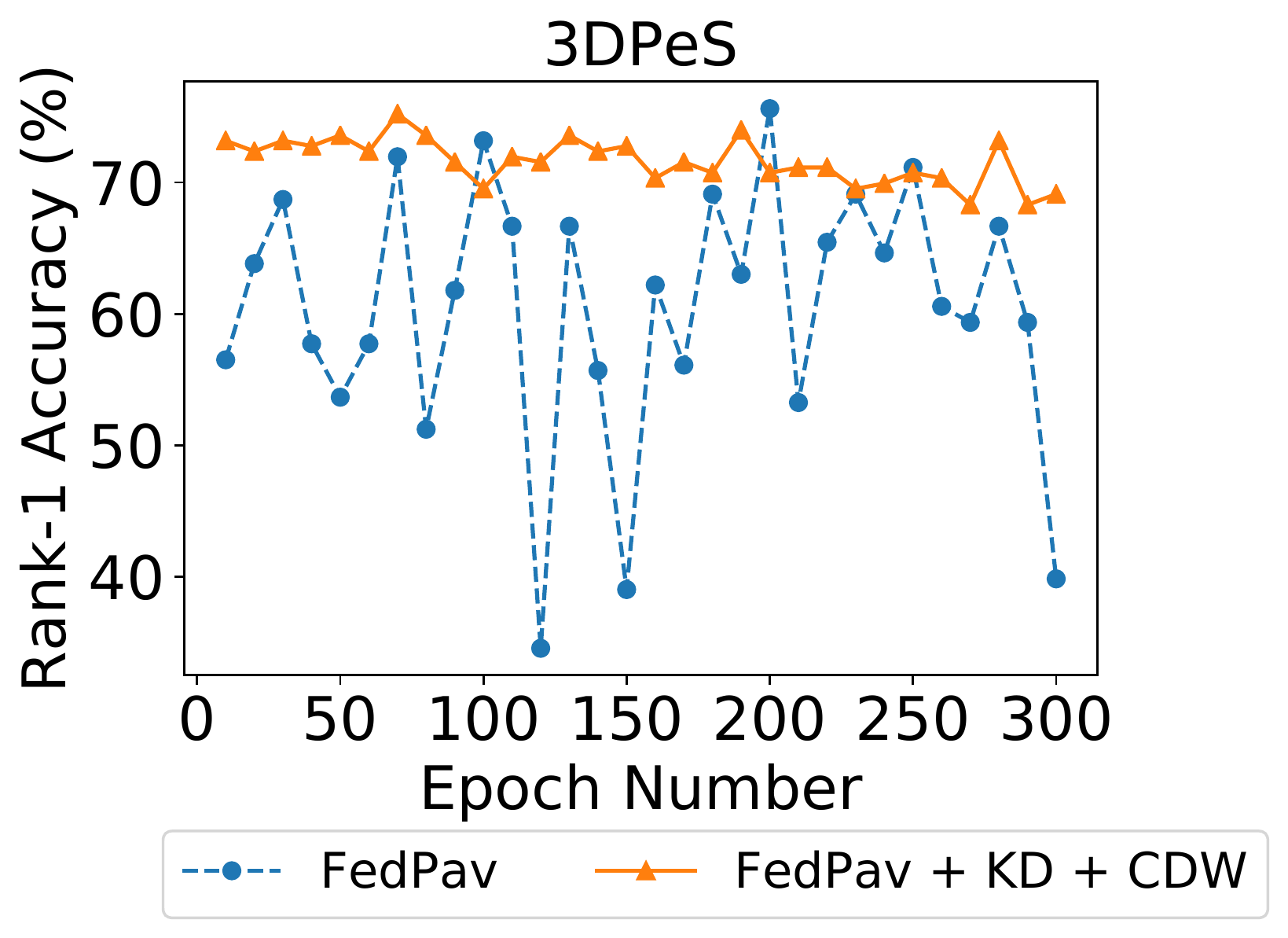}
    \caption{}
  \end{subfigure}
  
  \centering
  \begin{subfigure}[b]{0.45\linewidth}
    \includegraphics[width=\linewidth]{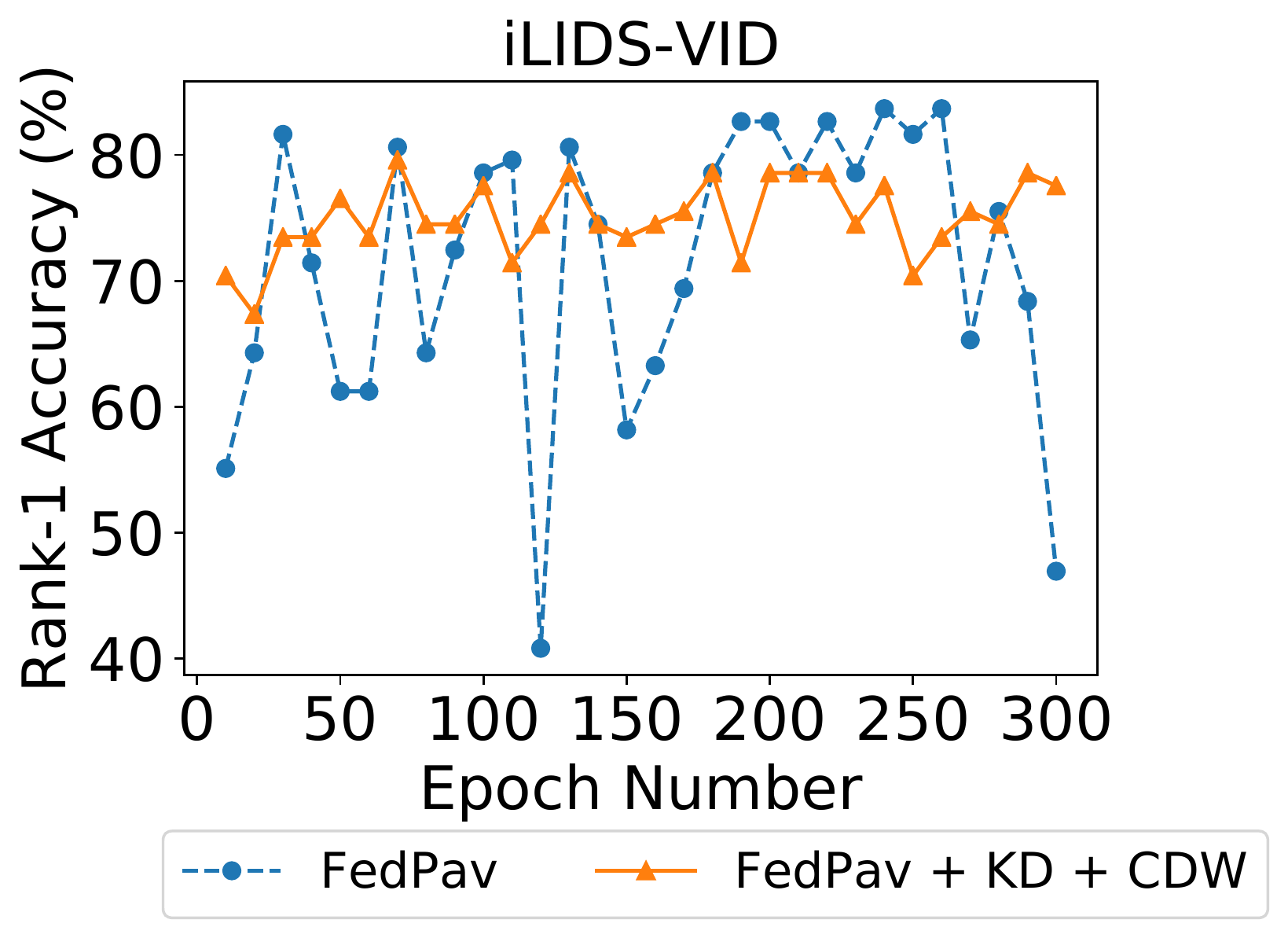}
    \caption{}
  \end{subfigure}
  \caption{Performance and convergence comparison of FedPav and FedPav with knowledge distillation (KD) and cosine distance weight (CDW) in all datasets, measured by rank-1 accuracy.}
  \label{fig:fedpav-cdw-kd-rank1}
 
\end{figure}

\begin{figure}[h!]
  \centering
  \begin{subfigure}[b]{0.45\linewidth}
    \includegraphics[width=\linewidth]{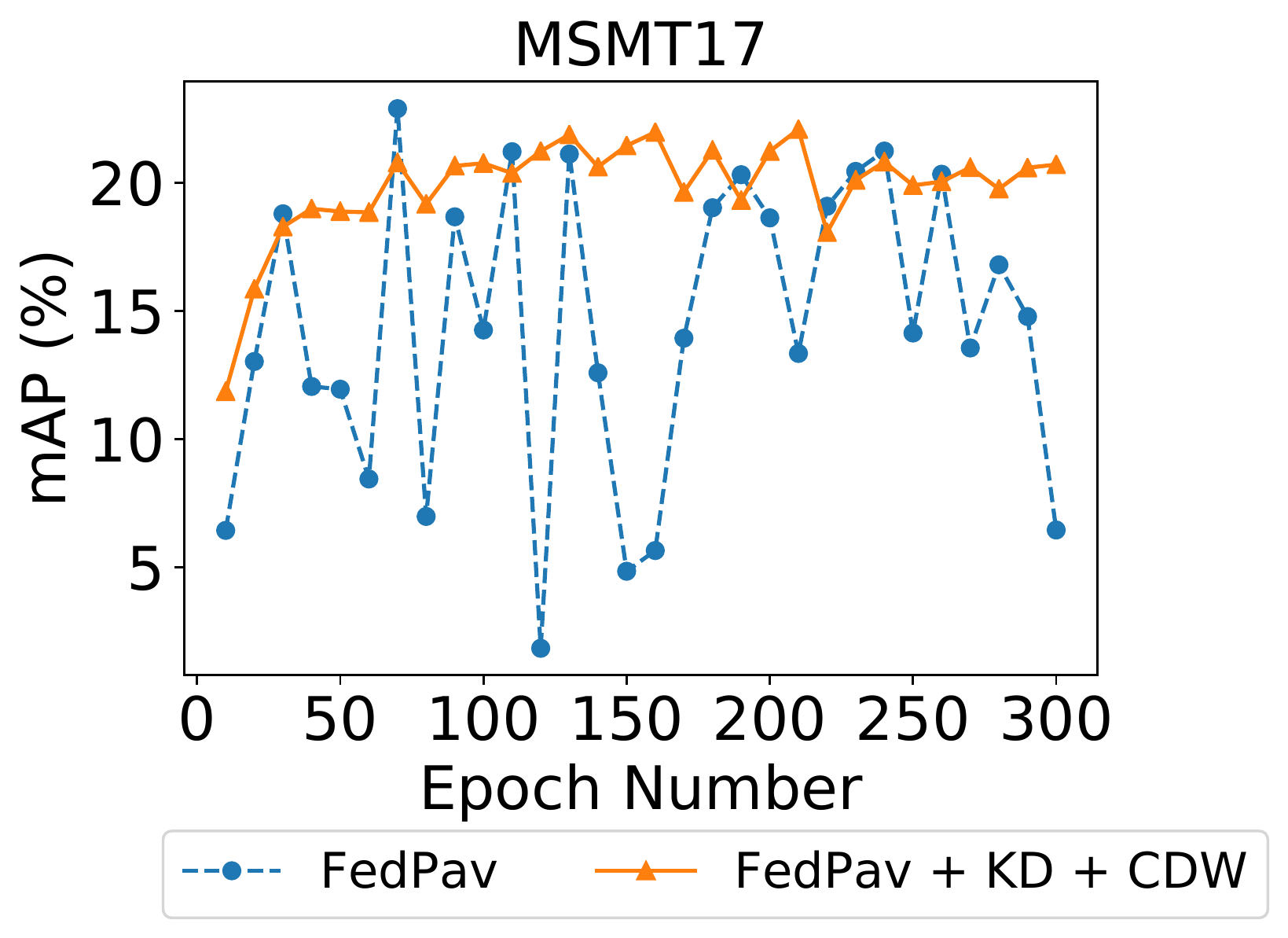}
    \caption{}
  \end{subfigure}
  \hspace{1em}%
  \begin{subfigure}[b]{0.45\linewidth}
    \includegraphics[width=\linewidth]{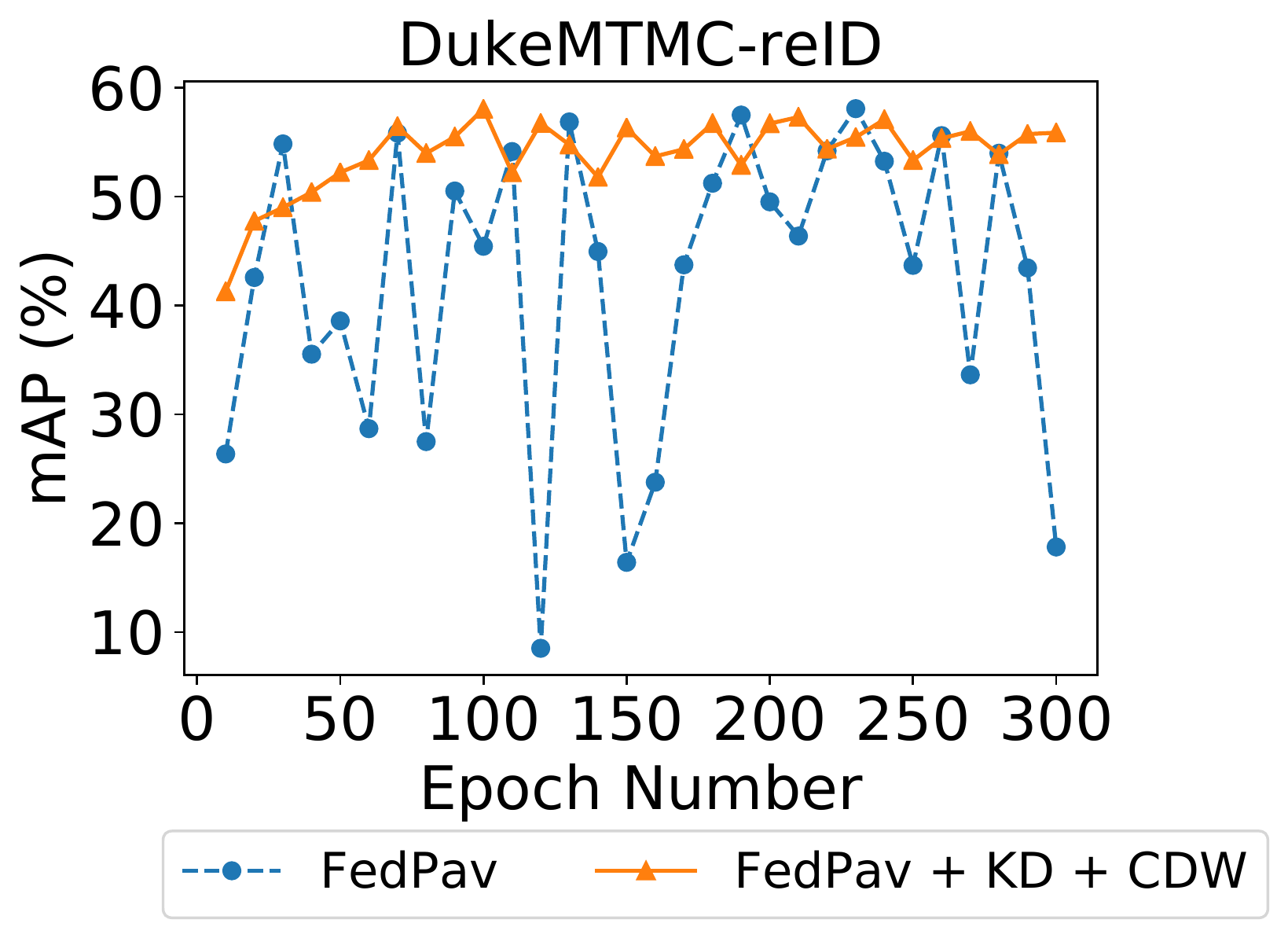}
    \caption{}
  \end{subfigure}
 
  \centering
  \begin{subfigure}[b]{0.45\linewidth}
    \includegraphics[width=\linewidth]{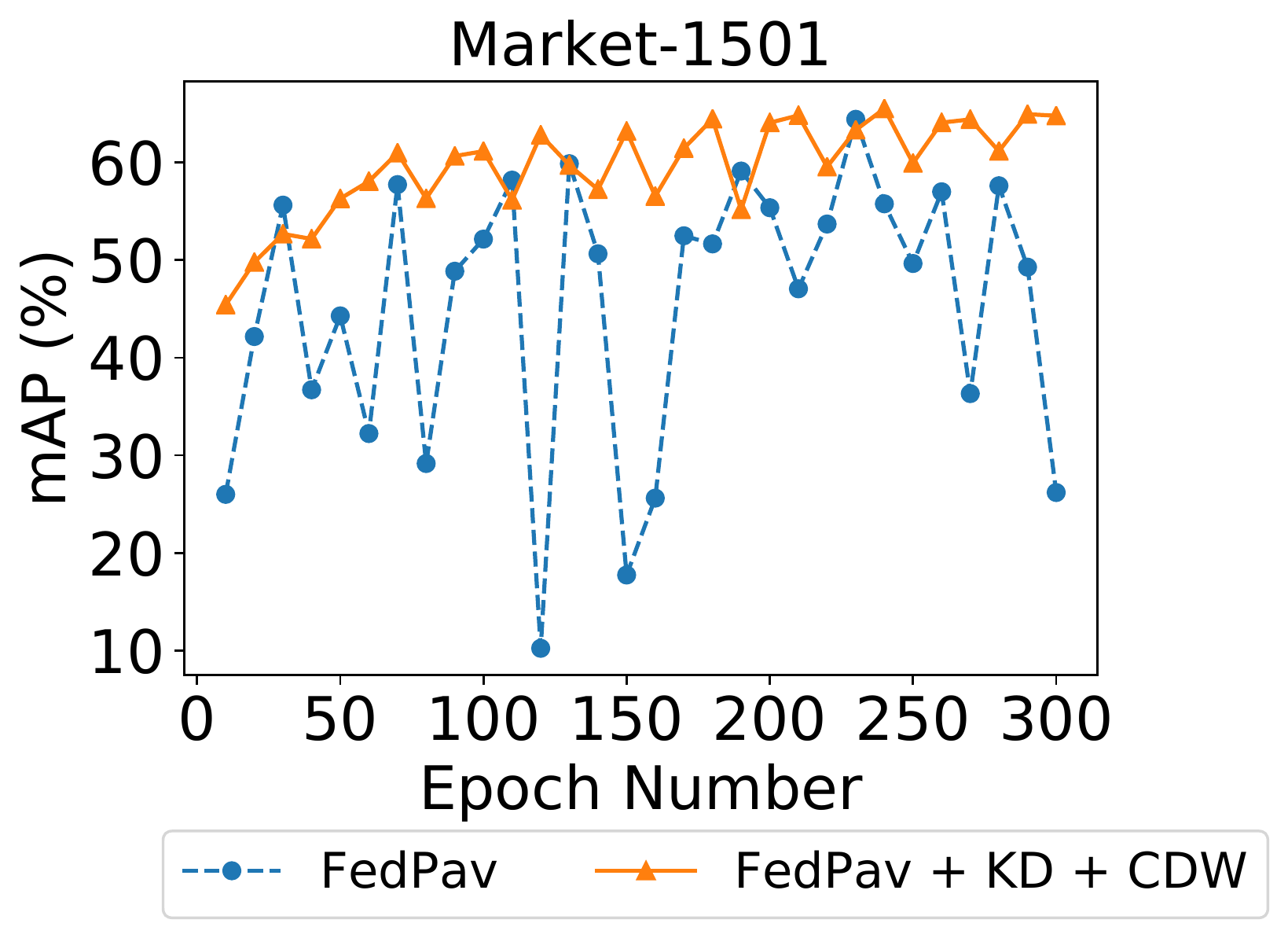}
    \caption{}
  \end{subfigure}
  \hspace{1em}%
  \begin{subfigure}[b]{0.45\linewidth}
    \includegraphics[width=\linewidth]{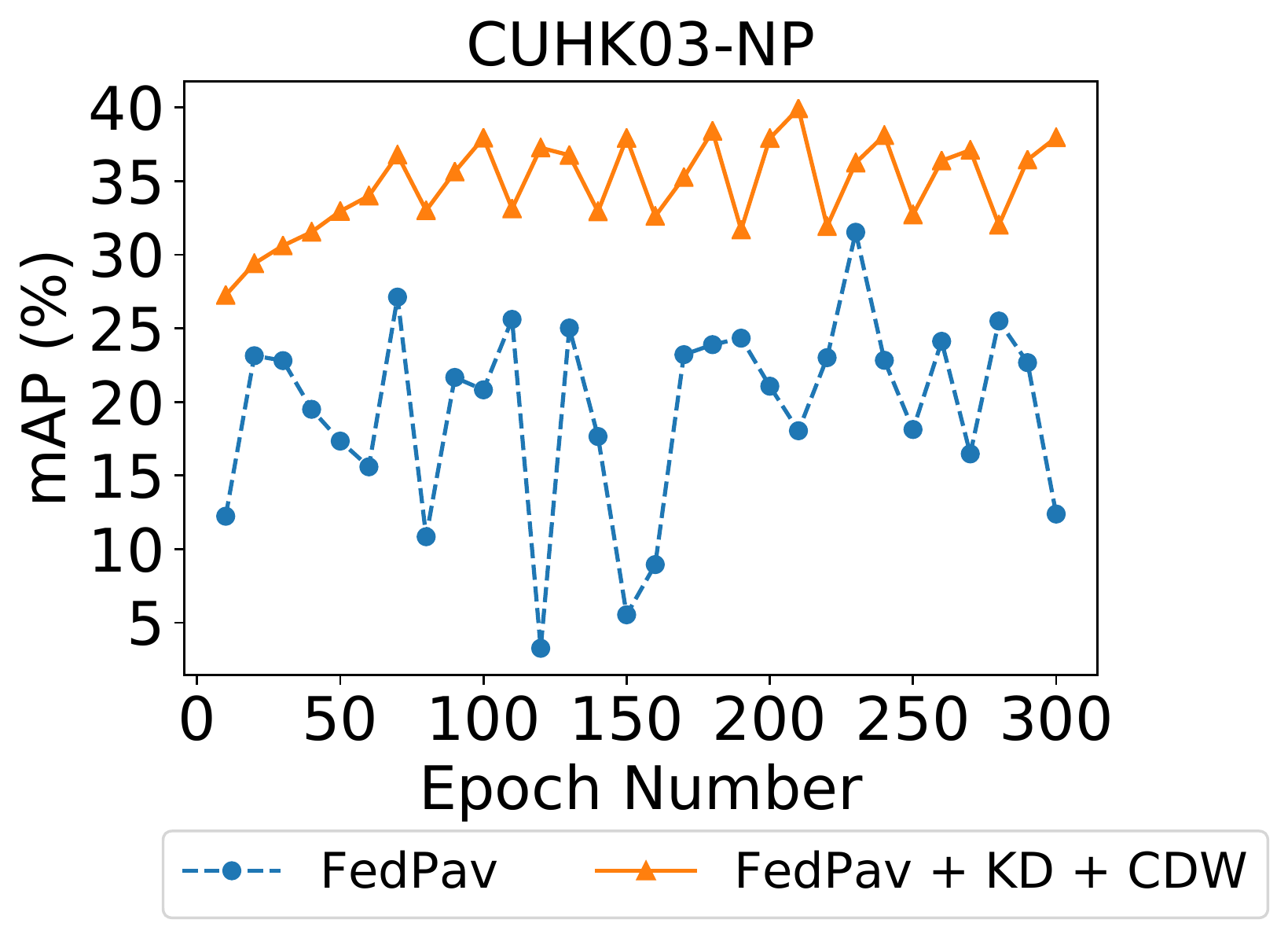}
    \caption{}
  \end{subfigure}
  
  \centering
  \begin{subfigure}[b]{0.45\linewidth}
    \includegraphics[width=\linewidth]{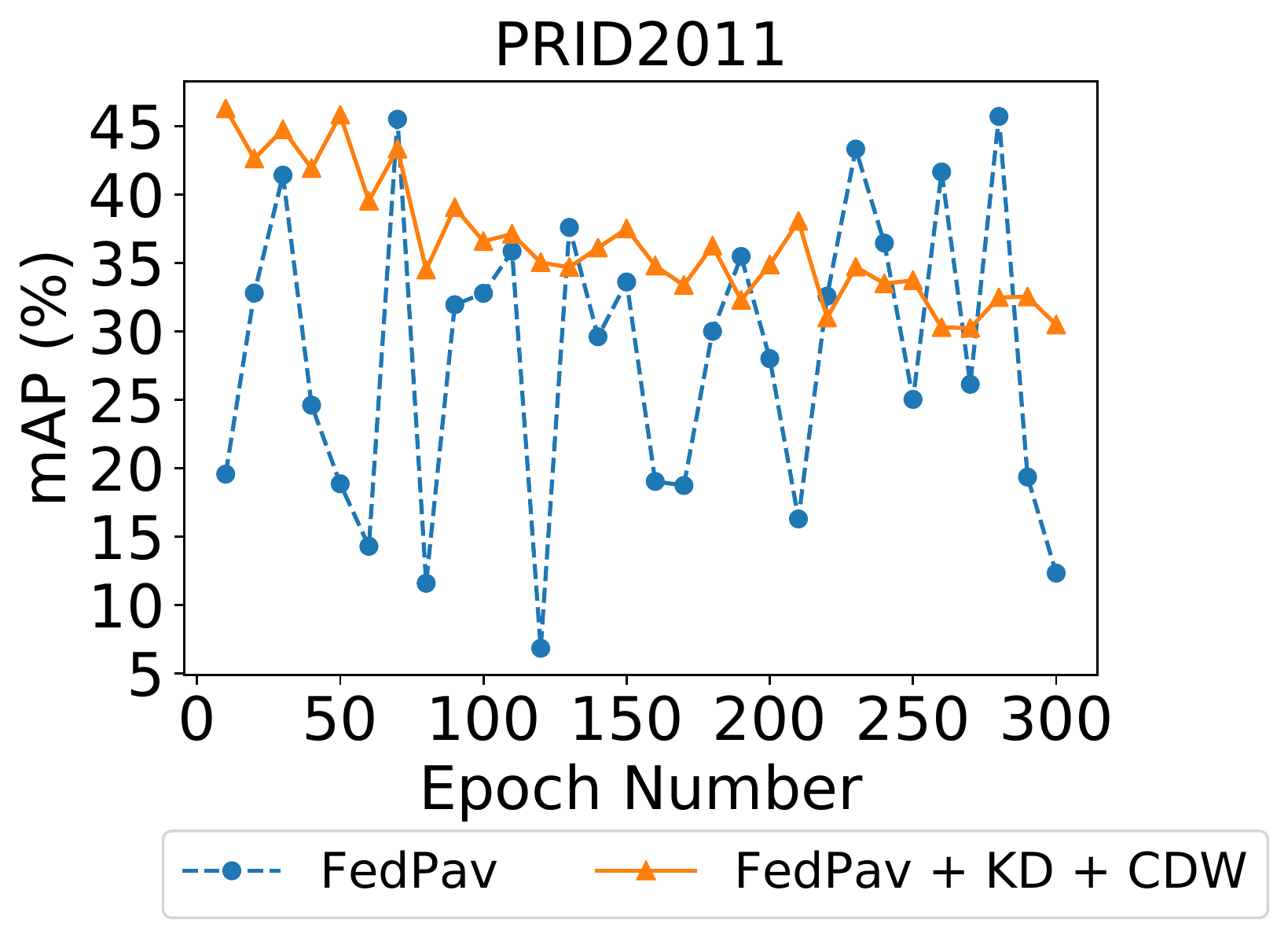}
    \caption{}
  \end{subfigure}
  \hspace{1em}%
  \begin{subfigure}[b]{0.45\linewidth}
    \includegraphics[width=\linewidth]{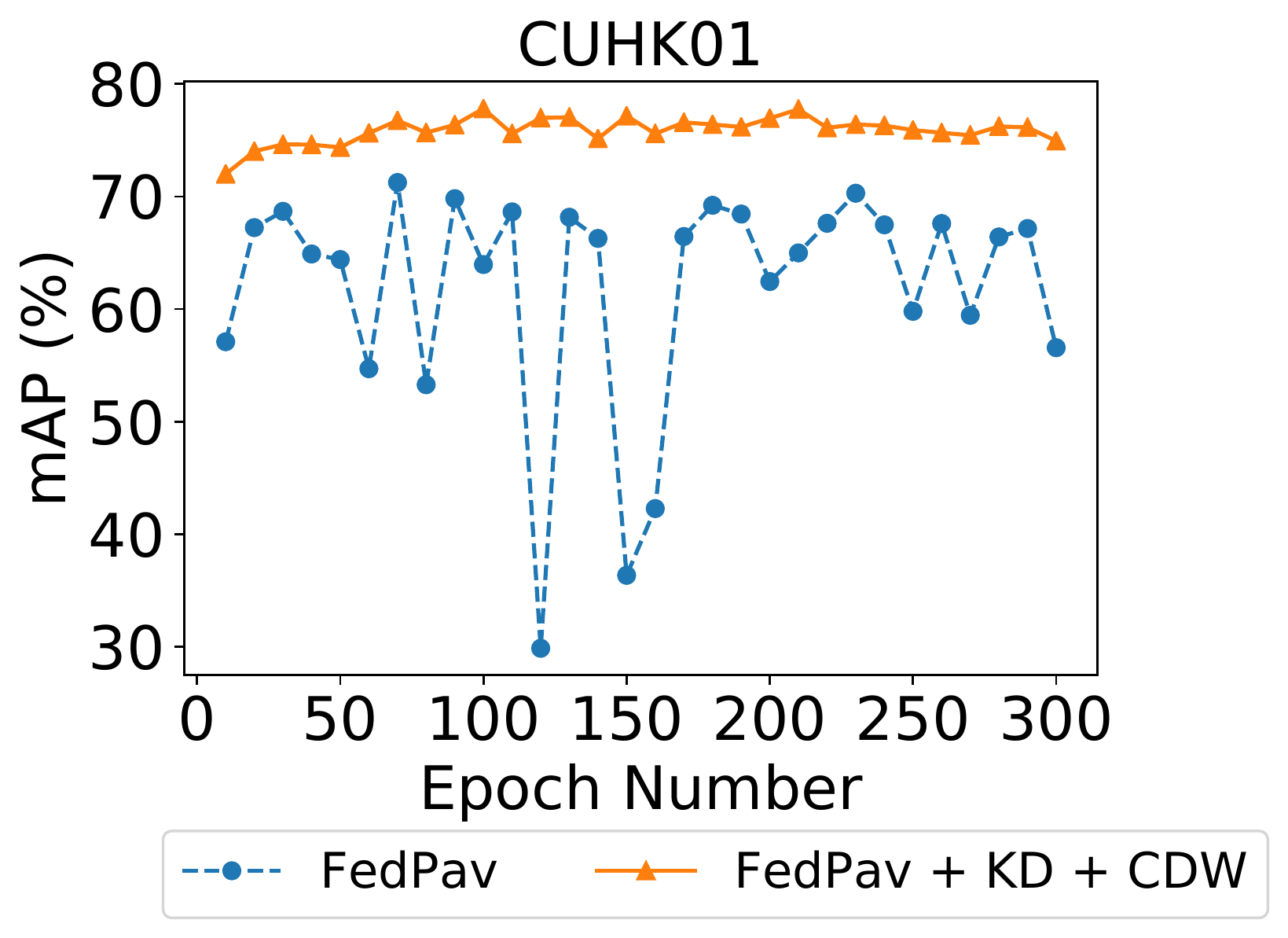}
    \caption{}
  \end{subfigure}
  
   \centering
  \begin{subfigure}[b]{0.45\linewidth}
    \includegraphics[width=\linewidth]{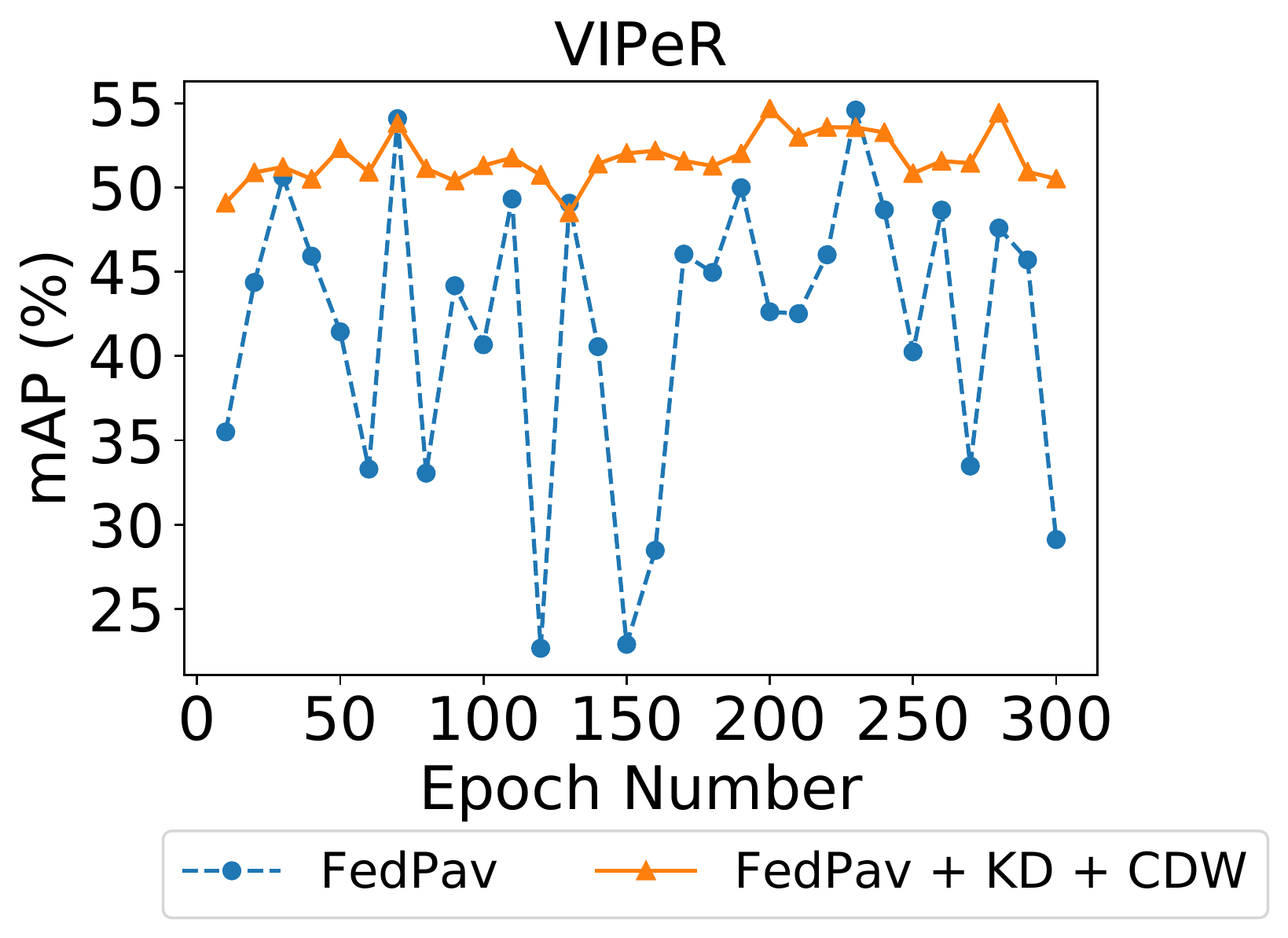}
    \caption{}
  \end{subfigure}
  \hspace{1em}%
  \begin{subfigure}[b]{0.45\linewidth}
    \includegraphics[width=\linewidth]{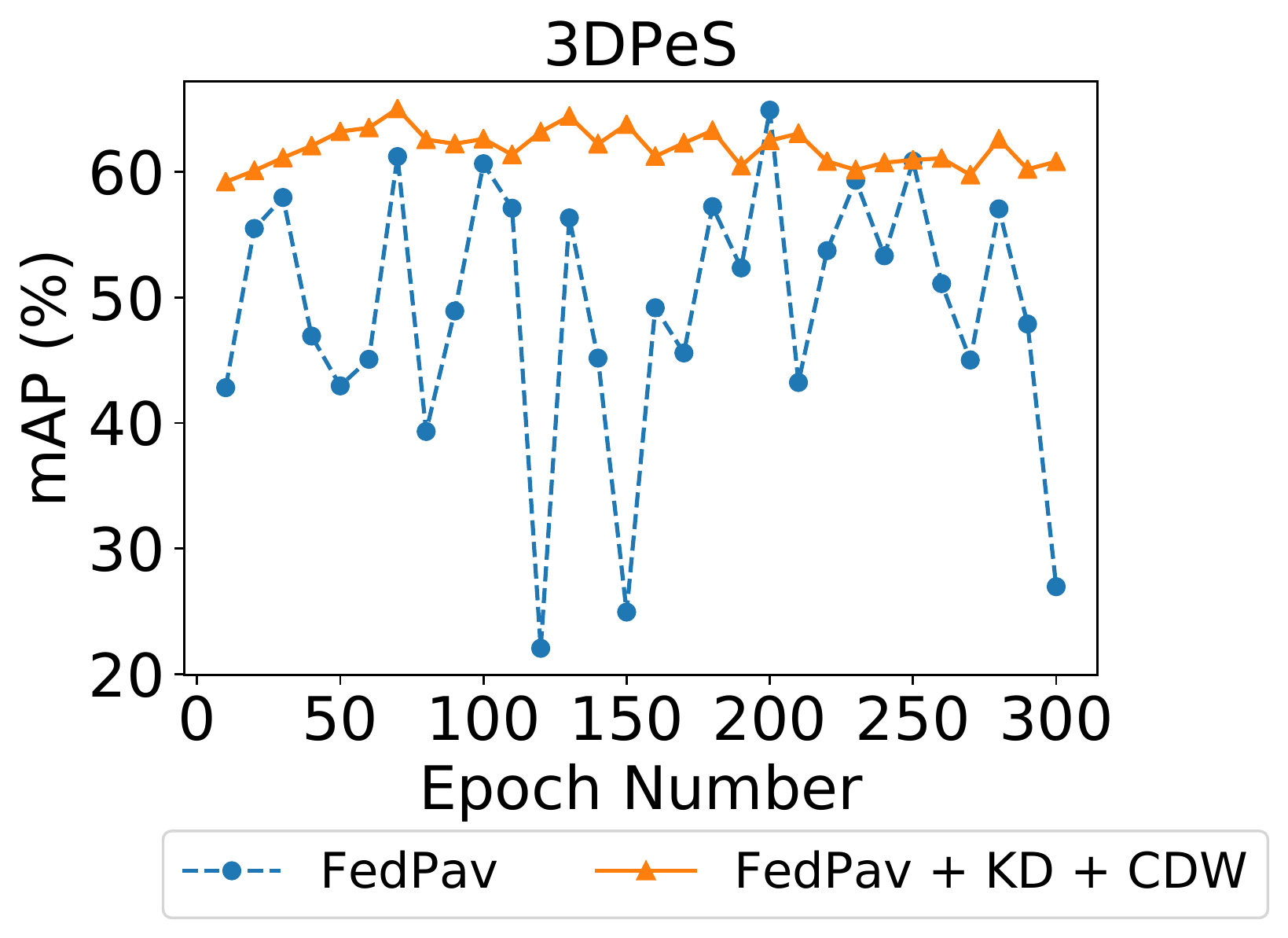}
    \caption{}
  \end{subfigure}
  
  \centering
  \begin{subfigure}[b]{0.45\linewidth}
    \includegraphics[width=\linewidth]{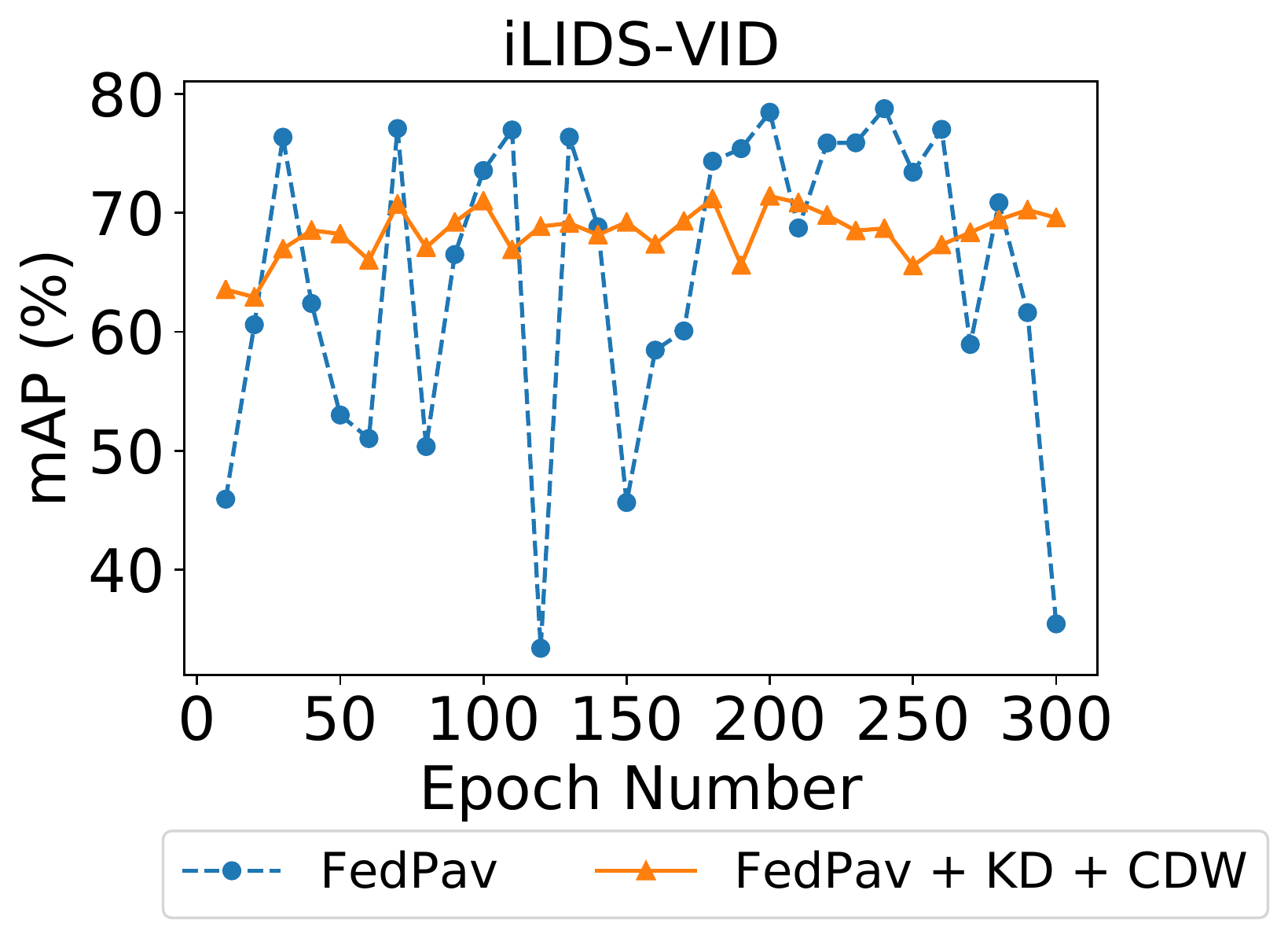}
    \caption{}
  \end{subfigure}
  \caption{Performance and convergence comparison of FedPav and FedPav with knowledge distillation (KD) and cosine distance weight (CDW) in all datasets, measured by mAP accuracy.}
  \label{fig:fedpav-cdw-kd-map}
 
\end{figure}

Figure \ref{fig:fedpav-cdw-kd-rank1} and \ref{fig:fedpav-cdw-kd-map} show the performance and convergence comparison of FedPav and FedPav with KD and CDW measured by rank-1 accuracy and mAP accuracy respectively. The rank-1 accuracy and mAP have similar patterns. The implementation of both KD and CDW achieve better convergence and superior performance on most datasets. Although the performance of FedPav with KD and CDW is lower than the fluctuated the best performance of FedPav on PRID2011 \cite{prid2011} dataset, it is higher than FedPav most of the time and has better convergence. The domain gap between the public dataset (CUHK02 \cite{cuhk02}) and the PRID2011 dataset could be the reason for this slight performance decay. We could achieve even better results if we select the public dataset for KD carefully.

\begin{table}[t]
\caption{Performance of the federated model on 9 datasets with batch size $B = 32$.}
\begin{tabular}{cccccc}
\hline
\multicolumn{2}{l}{Datasets}       & Rank-1            & Rank-5            & Rank-10  &mAP         \\ \hline \hline
\multicolumn{2}{l}{MSMT17 \cite{Wei2017Msmt}} & 41.01 & 53.79 & 59.02 & 21.49 \\
\multicolumn{2}{l}{DukeMTMC-reID \cite{zheng2017dukemtmc-reid}} & 74.30 & 85.58 & 89.36 & 56.92 \\
\multicolumn{2}{l}{Market-1501 \cite{Zheng2015Market1501}} & 83.42 & 93.29 & 95.86 & 60.68 \\
\multicolumn{2}{l}{CUHK03-NP \cite{Li2014CUHK03}} & 31.71 & 49.48 & 59.86 & 27.89 \\
\multicolumn{2}{l}{PRID2011 \cite{prid2011}} & 37.67 & 55.33 & 65.00 & 42.15 \\
\multicolumn{2}{l}{CUHK01 \cite{li2012cuhk01}} & 73.35 & 88.07 & 92.04 & 69.90 \\
\multicolumn{2}{l}{VIPeR \cite{Gray2008ViewpointIP}} & 48.10 & 66.24 & 76.16 & 52.58 \\
\multicolumn{2}{l}{3DPeS \cite{3dpes}} & 69.24 & 85.23 & 91.06 & 58.95 \\
\multicolumn{2}{l}{iLIDS-VID \cite{iLIDS-VID}} & 79.93 & 94.90 & 97.96 & 76.44 \\
\hline
\end{tabular}
\label{tab:B=32}
\end{table}

\begin{table}[t]
\caption{Performance of the federated model on 9 datasets with batch size $B = 64$.}
\begin{tabular}{cccccc}
\hline
\multicolumn{2}{l}{Datasets}       & Rank-1            & Rank-5            & Rank-10  &mAP         \\ \hline \hline
\multicolumn{2}{l}{MSMT17 \cite{Wei2017Msmt}} & 40.63 & 53.08 & 58.40 & 21.34 \\
\multicolumn{2}{l}{DukeMTMC-reID \cite{zheng2017dukemtmc-reid}} & 75.78 & 86.49 & 89.59 & 56.93 \\
\multicolumn{2}{l}{Market-1501 \cite{Zheng2015Market1501}} & 80.38 & 91.26 & 94.12 & 56.01 \\
\multicolumn{2}{l}{CUHK03-NP \cite{Li2014CUHK03}} & 26.64 & 44.55 & 54.55 & 23.96 \\
\multicolumn{2}{l}{PRID2011 \cite{prid2011}} & 33.00 & 56.67 & 64.33 & 38.45 \\
\multicolumn{2}{l}{CUHK01 \cite{li2012cuhk01}} & 70.16 & 86.93 & 91.29 & 66.76 \\
\multicolumn{2}{l}{VIPeR \cite{Gray2008ViewpointIP}} & 45.68 & 65.61 & 72.78 & 50.30 \\
\multicolumn{2}{l}{3DPeS \cite{3dpes}} & 67.07 & 83.74 & 88.08 & 56.83 \\
\multicolumn{2}{l}{iLIDS-VID \cite{iLIDS-VID}} & 78.23 & 92.18 & 97.28 & 73.45 \\
\hline
\end{tabular}
\label{tab:B=64}
\end{table}

\begin{table}[t]
\caption{Performance of the federated model on 9 datasets with batch size $B = 128$.}
\begin{tabular}{cccccc}
\hline
\multicolumn{2}{l}{Datasets}       & Rank-1            & Rank-5            & Rank-10  &mAP         \\ \hline \hline
\multicolumn{2}{l}{MSMT17 \cite{Wei2017Msmt}} & 36.58 & 49.12 & 54.46 & 17.97 \\
\multicolumn{2}{l}{DukeMTMC-reID \cite{zheng2017dukemtmc-reid}} & 74.27 & 85.83 & 89.32 & 55.14 \\
\multicolumn{2}{l}{Market-1501 \cite{Zheng2015Market1501}} & 75.50 & 88.32 & 91.95 & 49.66 \\
\multicolumn{2}{l}{CUHK03-NP \cite{Li2014CUHK03}} & 23.81 & 40.67 & 50.52 & 21.34 \\
\multicolumn{2}{l}{PRID2011 \cite{prid2011}} & 33.00 & 59.33 & 68.33 & 39.03 \\
\multicolumn{2}{l}{CUHK01 \cite{li2012cuhk01}} & 65.09 & 84.57 & 90.05 & 62.16 \\
\multicolumn{2}{l}{VIPeR \cite{Gray2008ViewpointIP}} & 41.24 & 61.18 & 68.99 & 45.97 \\
\multicolumn{2}{l}{3DPeS \cite{3dpes}} & 72.09 & 87.94 & 91.60 & 61.19 \\
\multicolumn{2}{l}{iLIDS-VID \cite{iLIDS-VID}} & 77.21 & 90.48 & 94.22 & 71.32 \\
\hline
\end{tabular}
\label{tab:B=128}
\end{table}

\begin{table}[h]
\caption{Performance of models trained on each dataset (\textit{local training}).}
\begin{tabular}{cccccc}
\hline
\multicolumn{2}{l}{Datasets}       & Rank-1            & Rank-5            & Rank-10  &mAP         \\ \hline \hline
\multicolumn{2}{l}{MSMT17 \cite{Wei2017Msmt}} & 49.56 & 63.06 & 67.85 & 28.66 \\
\multicolumn{2}{l}{DukeMTMC-reID \cite{zheng2017dukemtmc-reid}} & 80.07 & 89.45 & 92.19 & 62.41 \\
\multicolumn{2}{l}{Market-1501 \cite{Zheng2015Market1501}} & 88.93 & 95.34 & 96.88 & 72.62 \\
\multicolumn{2}{l}{CUHK03-NP \cite{Li2014CUHK03}} & 49.29 & 68.86 & 76.57 & 44.52 \\
\multicolumn{2}{l}{PRID2011 \cite{prid2011}} & 55.00 & 75.00 & 84.00 & 59.35 \\
\multicolumn{2}{l}{CUHK01 \cite{li2012cuhk01}} & 69.03 & 87.04 & 91.87 & 64.73 \\
\multicolumn{2}{l}{VIPeR \cite{Gray2008ViewpointIP}} & 27.53 & 51.27 & 62.97 & 33.27 \\
\multicolumn{2}{l}{3DPeS \cite{3dpes}} & 65.45 & 83.33 & 87.80 & 51.83 \\
\multicolumn{2}{l}{iLIDS-VID \cite{iLIDS-VID}} & 52.04 & 68.37 & 85.71 & 45.13 \\
\hline
\end{tabular}
\label{tab:local-training}
\end{table}

\begin{table}[h]
\caption{Performance of the federated model on 9 datasets using FedPav with cosine distance weight.}
\begin{tabular}{cccccc}
\hline
\multicolumn{2}{l}{Datasets}       & Rank-1            & Rank-5            & Rank-10  &mAP         \\ \hline \hline
\multicolumn{2}{l}{MSMT17 \cite{Wei2017Msmt}} & 43.35 & 55.85 & 61.26 & 22.86 \\
\multicolumn{2}{l}{DukeMTMC-reID \cite{zheng2017dukemtmc-reid}} & 77.17 & 87.30 & 90.47 & 60.53 \\
\multicolumn{2}{l}{Market-1501 \cite{Zheng2015Market1501}} & 85.14 & 94.29 & 96.51 & 62.97 \\
\multicolumn{2}{l}{CUHK03-NP \cite{Li2014CUHK03}} & 34.24 & 54.17 & 63.14 & 30.48 \\
\multicolumn{2}{l}{PRID2011 \cite{prid2011}} & 41.33 & 62.33 & 72.67 & 46.59 \\
\multicolumn{2}{l}{CUHK01 \cite{li2012cuhk01}} & 78.33 & 91.53 & 95.03 & 74.25 \\
\multicolumn{2}{l}{VIPeR \cite{Gray2008ViewpointIP}} & 47.89 & 66.46 & 75 & 52.33 \\
\multicolumn{2}{l}{3DPeS \cite{3dpes}} & 72.76 & 88.35 & 91.06 & 63.79 \\
\multicolumn{2}{l}{iLIDS-VID \cite{iLIDS-VID}} & 80.61 &93.88 & 96.60 & 75.80 \\
\hline
\end{tabular}
\label{tab:fedpav-cdw}
\end{table}

 \begin{figure}[h]
    \centering
    \includegraphics[width=0.45\textwidth]{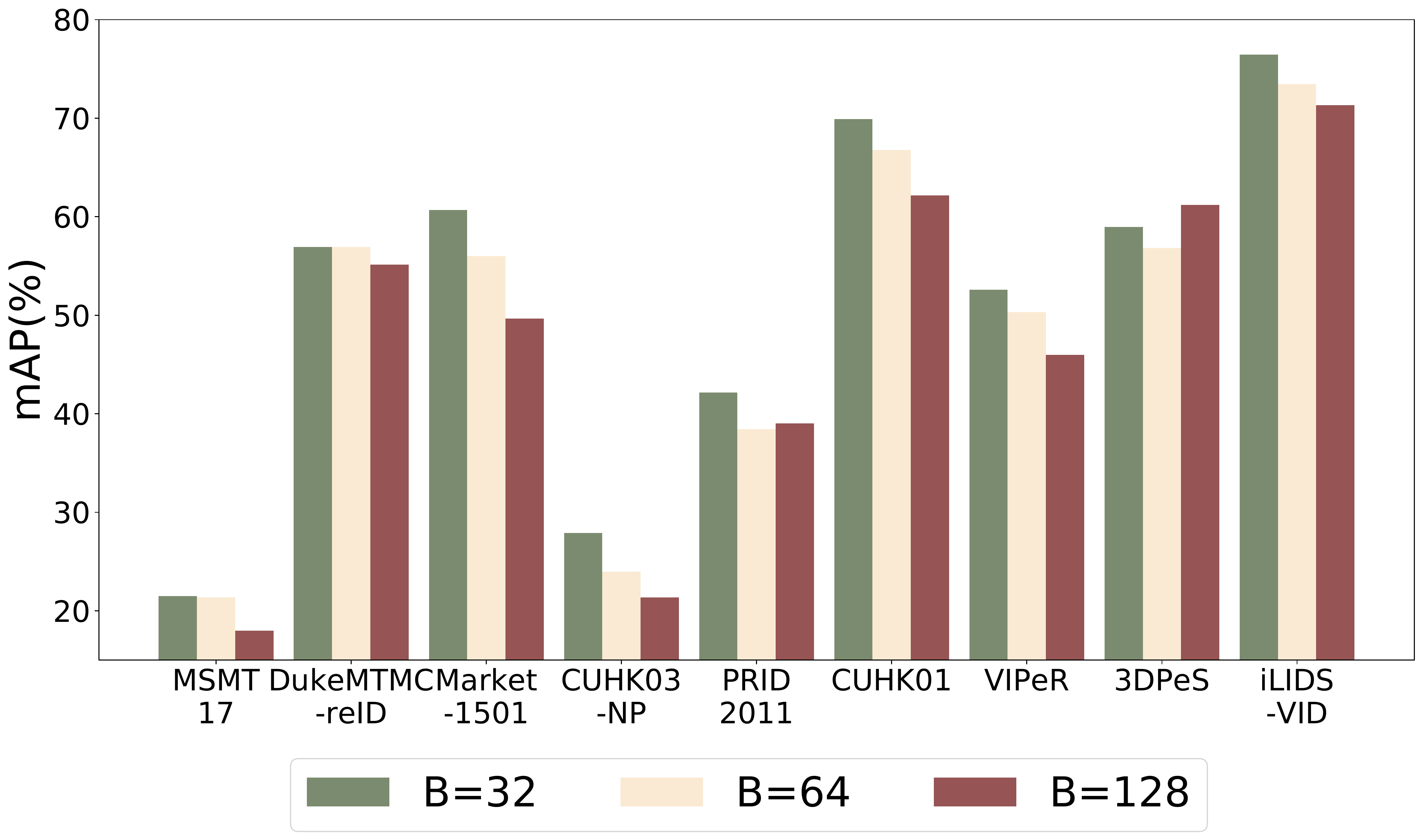}
    \caption{Performance (mAP) comparison of different batch size.}
    \label{fig:batch-size}
\end{figure}

\begin{figure}[h]
\centering
\includegraphics[width=0.45\textwidth]{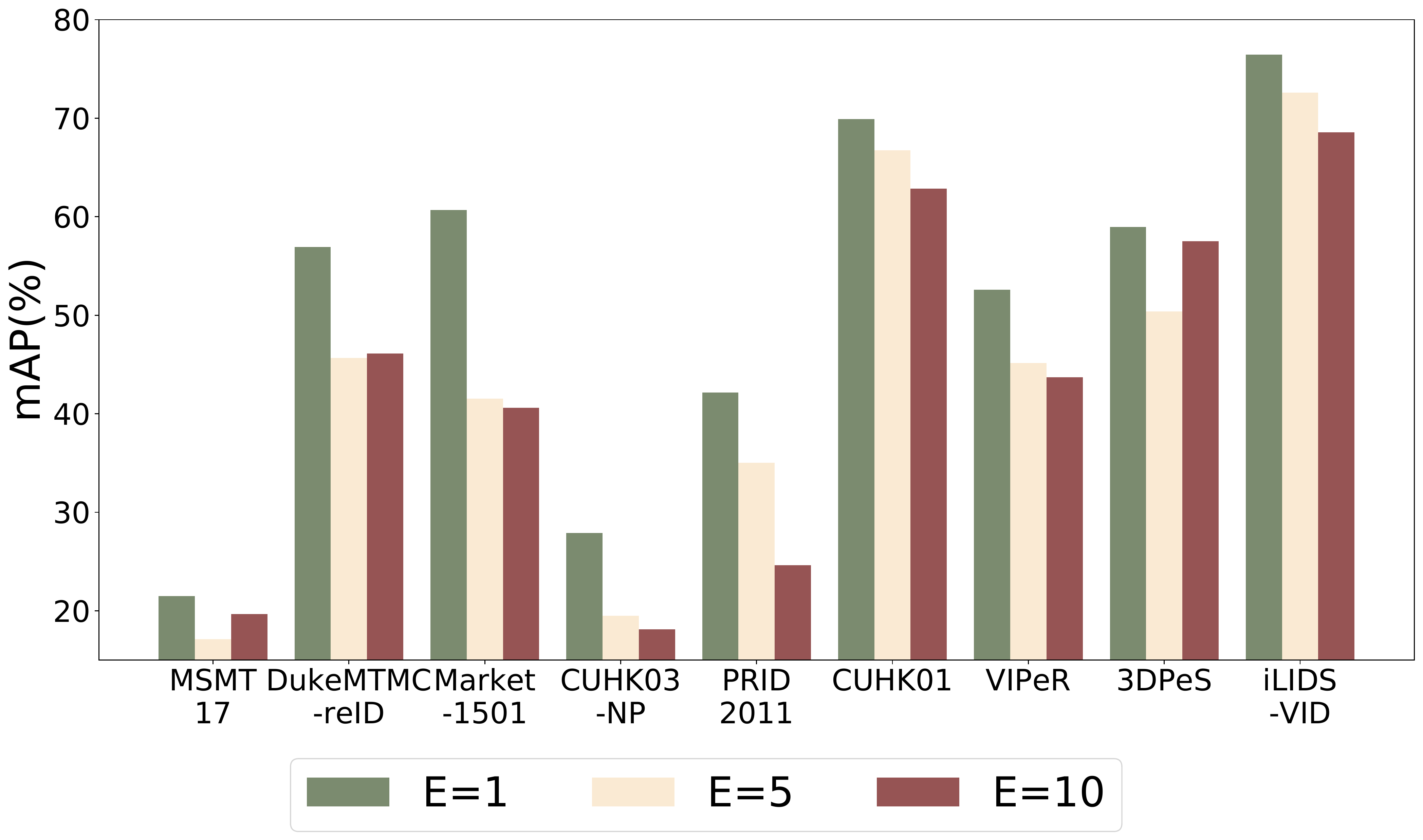}
\caption{Performance (mAP) comparison of different number of local epochs.}
\label{fig:local-epoch}
\end{figure}

\clearpage

\clearpage

\SetKwInput{KwInput}{Input}                
\SetKwInput{KwOutput}{Output}              
\SetKwFunction{FnClient}{}
\SetKwFunction{FnServer}{}

\section{Training Algorithm}

In Algorithm \ref{algo:fedavg}, we present the details of the FedAvg algorithm proposed by McMahan et al. \cite{McMahanMRHA17}. In Algorithm \ref{algo:fedpav-kd-weight}, we present the algorithm that applies both knowledge distillation and cosine distance weight to FedPav.

\begin{algorithm}[h]
    \caption{Federated Averaging (FedAvg)}
    \label{algo:fedavg}
    \SetAlgoLined
    \KwInput{$E, B, K, \eta, T, N, n_k, n$}
    \KwOutput{$w^T$}
    
    \SetKwProg{Fn}{Server}{:}{}
    \Fn{\FnServer}{
        initialize $w^0$\;
        \For{each round t = 0 to T-1}{
            $C_t \leftarrow$ (randomly select K out of N clients)\;
            \For{each client $k \in C_t$ concurrently}{
                $w^{t+1}_k \leftarrow $ \textbf{ClientExecution}($w^t$, $k$)\;
            }
            // $n$: total size of dataset; $n_k$: size of client $k$'s dataset\;
            $w^{t+1} \leftarrow \sum_{k \in C_t} \frac{n_k}{n} w^{t+1}_k$\;
        }
        \KwRet $w^{T}$\;
    }
    
    \SetKwProg{Fn}{ClientExecution}{:}{\KwRet}
    \Fn{\FnClient{w, k}}{
        $\mathcal{B} \leftarrow $ (divide local data into batches of size $B$)\;
        \For{each local epoch e = 0 to E-1}{
            \For{$b \in \mathcal{B}$}{
                $w \leftarrow w - \eta \triangledown \mathcal{L}(w; b)$\;
            }
        }
        \KwRet $w$\;
    }

\end{algorithm}

\begin{algorithm}[b]
    \caption{FedPav with Knowledge Distillation and Cosine Distance Weight}
    \label{algo:fedpav-kd-weight}
    \SetAlgoLined
    \KwInput{$E, B, K, \eta, T, N, \mathcal{D}_{shared}$}
    \KwOutput{$w^T, w^T_k$}
    
    \SetKwProg{Fn}{Server}{:}{}
    \Fn{\FnServer}{
        initialize $w^0$\;
        \For{each round t = 0 to T-1}{
            $C_t \leftarrow$ (randomly select K out of N clients)\;
            \For{each client $k \in C_t$ concurrently}{
                $w^{t+1}_k, \ell^{t+1}_k , m_k \leftarrow $ \textbf{ClientExecution}($w^t$, $k$, $t$)\;
            }
            $m \leftarrow \sum_{k \in C_t} \frac{1}{m_k}$\;
            $w^{t+1} \leftarrow \sum_{k \in C_t} \frac{m_k}{m} w^{t+1}_k$\;
            $\ell^{t+1} \leftarrow \frac{1}{K} \sum_{k \in C_t} \ell^{t+1}_k$\;
            $w^{t+1} \leftarrow $ (fine-tune $w^{t+1}$ with $\ell^{t+1}$ and $\mathcal{D}_{shared}$)
        }
        \KwRet $w^{T}$\;
    }
    
    \SetKwProg{Fn}{ClientExecution}{:}{}
    \Fn{\FnClient{$w$, k, t}}{
        $v \leftarrow$ (retrieve additional layers $v$ if $t > 0$ else initialize)\;
        $\mathcal{B} \leftarrow $ (divide local data into batches of size $B$)\;
        // $(w, v)$ concatenation of two vectors\;
        $(w^{t}, v^{t}) \leftarrow (w, v)$\;
        \For{each local epoch e = 0 to E-1}{
            \For{$b \in \mathcal{B}$}{
                $(w^{t}, v^{t}) \leftarrow (w^{t}, v^{t}) - \eta \triangledown \mathcal{L}((w^{t}, v^{t}); b)$\;
            }
        }
        
        $\mathcal{D}_{batch} \leftarrow$ one batch of $\mathcal{B}$\;
        \For{each data $d \in \mathcal{D}_{batch}$}{
            $f \leftarrow$ (generate logits with data and $(w, v)$)\;
            $f^{t} \leftarrow$ (generate logits with data and $(w^{t}, v^{t})$)\;
            $m_d \leftarrow 1 - cosine\_similarity(f, f^{t})$\;
        }
        $m^{t} \leftarrow \frac{1}{\lvert \mathcal{D}_{batch} \rvert} \sum_{d \in \mathcal{D}_{batch}} m_d$\;
        $\ell^t \leftarrow$ (predict soft labels with $w^t$, $\mathcal{D}_{shared}$)\;
        store $v^{t}$\;
        \KwRet $w^{t}, \ell^t, m^{t}$\;
    }
\end{algorithm}

\end{document}